\documentclass{article}

\usepackage[final, nonatbib]{neurips_2019}

\usepackage[utf8]{inputenc} 
\usepackage[T1]{fontenc}    
\usepackage{hyperref}       
\usepackage{url}            
\usepackage{booktabs}       
\usepackage{amsfonts}       
\usepackage{nicefrac}       
\usepackage{microtype}      

\usepackage{listings}
\usepackage{multirow}
\usepackage[pdftex]{graphicx}
\usepackage{float}
\usepackage{amsmath, amsthm, amssymb}
\usepackage{caption}
\usepackage{tikz}
\usepackage{xcolor}
\usepackage{comment}
\usepackage[disable]{todonotes}
\usepackage{placeins}
\usepackage{rotating}

\definecolor{darkgreen}{rgb}{0, 0.5, 0}
\usepackage[numbers]{natbib}
\usepackage{cleveref}
\Crefname{equation}{Eq.}{Eqs.}
\Crefname{figure}{Fig.}{Figs.}
\Crefname{tabular}{Tab.}{Tabs.}
\Crefname{section}{Sec.}{Secs.}
\Crefname{appendix}{App.}{Apps.}
\usepackage{wrapfig}
\newcommand*{\defeq}{\mathrel{\vcenter{\baselineskip0.5ex \lineskiplimit0pt
                     \hbox{\scriptsize.}\hbox{\scriptsize.}}}%
                     =}
\DeclareMathOperator*{\argmin}{argmin}

\makeatletter
\renewcommand{\paragraph}{%
  \@startsection{paragraph}{4}%
  {\z@}{0.15ex plus 0.5ex minus .1ex}{-1em}{\normalsize\bf}}   

\let\originalparagraph\paragraph
\renewcommand{\paragraph}[2][.]{\originalparagraph{#2#1}}
\makeatother

\usepackage[utf8]{inputenc}
\usepackage{intcalc}
\usepackage{ifthen}
\usetikzlibrary{patterns,arrows,calc,shapes}
\usetikzlibrary{decorations.markings,decorations.pathreplacing}

\title{Grid Saliency for \\Context Explanations of Semantic Segmentation}

\author{%
  Lukas Hoyer\\
  Bosch Center for Artificial Intelligence\\
  \texttt{lukas.hoyer@outlook.com} \\
  \And
  Mauricio Munoz \\
  Bosch Center for Artificial Intelligence\\
  \texttt{AndresMauricio.MunozDelgado@bosch.com} \\
  \AND
  Prateek Katiyar \\
  Bosch Center for Artificial Intelligence\\
  \texttt{Prateek.Katiyar@bosch.com} \\
  \And
  Anna Khoreva \\
  Bosch Center for Artificial Intelligence\\
  \texttt{Anna.Khoreva@bosch.com} \\
  \And
  Volker Fischer \\
  Bosch Center for Artificial Intelligence\\
  \texttt{Volker.Fischer@bosch.com} \\
}

\begin{document}

\maketitle

\begin{abstract}
	
Recently, there has been a growing interest in developing saliency methods that provide visual explanations of network predictions. 
Still, the usability of existing methods is limited to image classification models. 
To overcome this limitation, we extend the existing approaches to generate \emph{grid saliencies}, 
which provide spatially coherent visual explanations for (pixel-level) dense prediction networks. 
As the proposed grid saliency allows to spatially disentangle the object and its context, 
we specifically explore its potential to produce context explanations for semantic segmentation networks, 
discovering which context most influences the class predictions inside a target object area. 
We investigate the effectiveness of grid saliency on a synthetic dataset with an artificially induced bias between objects and their context as well as on the real-world Cityscapes dataset using state-of-the-art segmentation networks. 
Our results show that grid saliency can be successfully used to provide easily interpretable context explanations and, 
moreover, can be employed for detecting and localizing contextual biases present in the data.

\end{abstract}

\section{Introduction}
\label{sec:introduction}

\begin{figure}
	\begin{centering}
		\setlength{\tabcolsep}{0em}
		\renewcommand{\arraystretch}{0}
		\par\end{centering}
	\begin{centering}
		\vspace{-1em}
		\hfill{}%
		\begin{tabular}{@{}c@{\hskip 0.05in}c@{}c@{}c@{}}
			
			\includegraphics[width=0.24\linewidth, height=0.075\textheight]{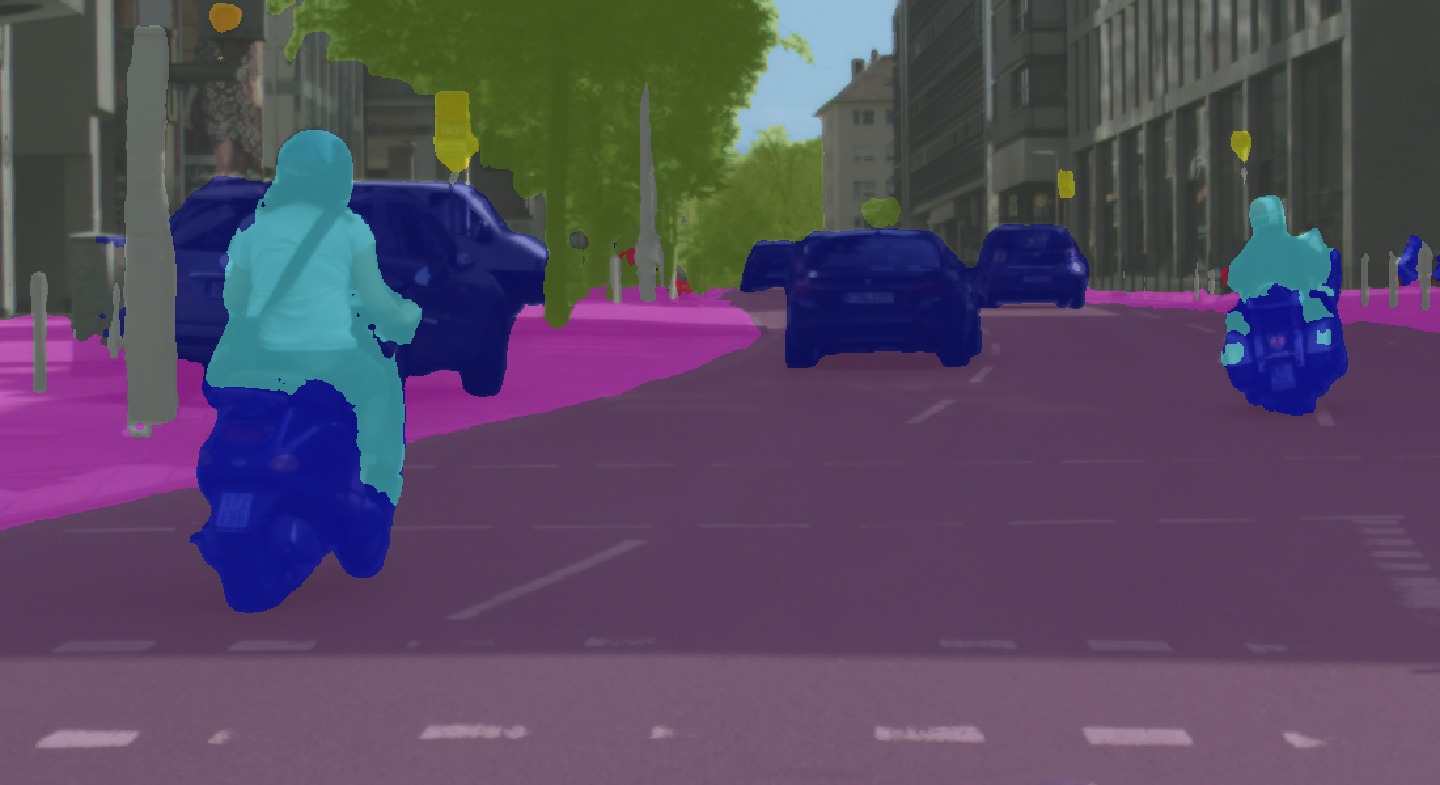} & {\footnotesize{}}
			\includegraphics[width=0.24\linewidth, height=0.075\textheight]{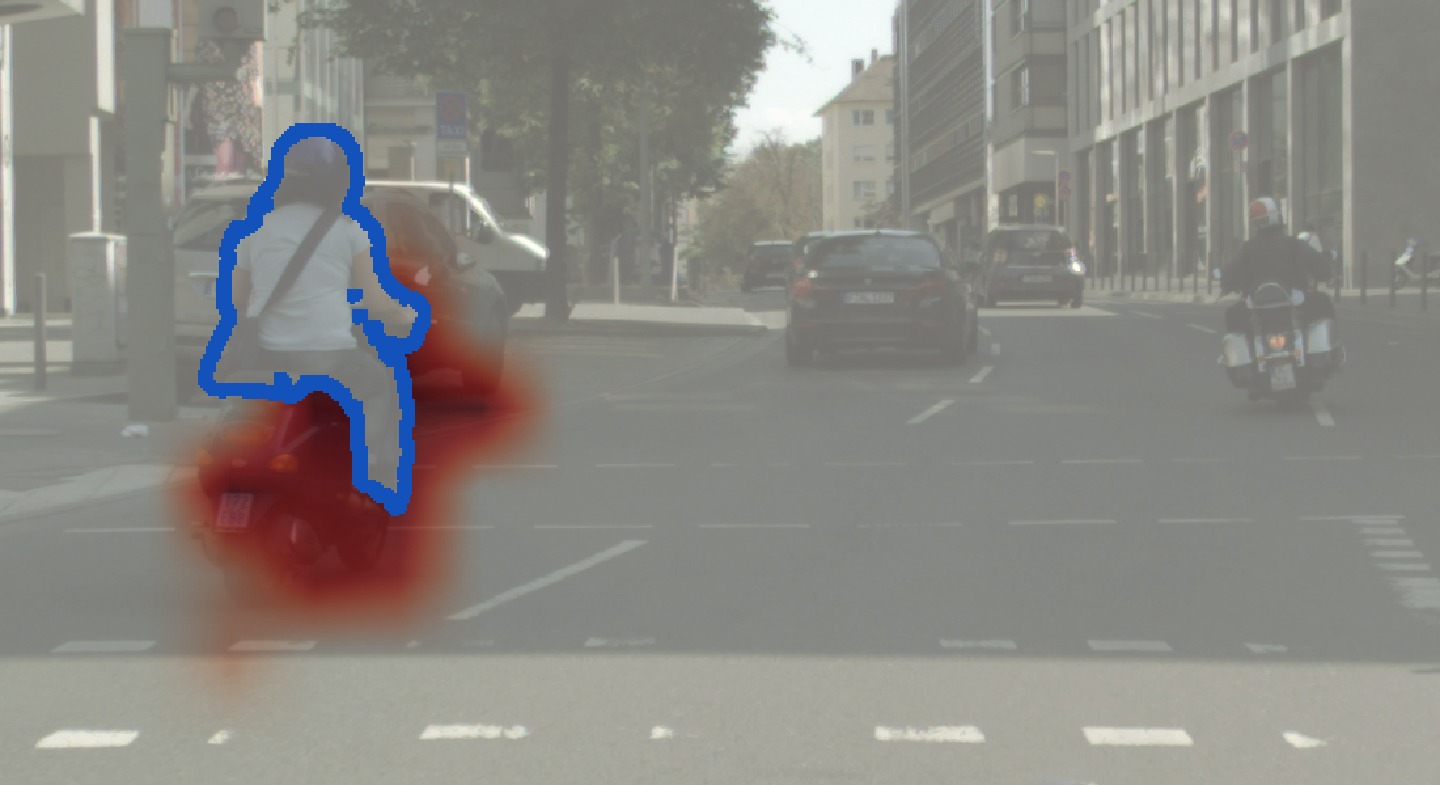} & {\footnotesize{}}
			\includegraphics[width=0.24\linewidth, height=0.075\textheight]{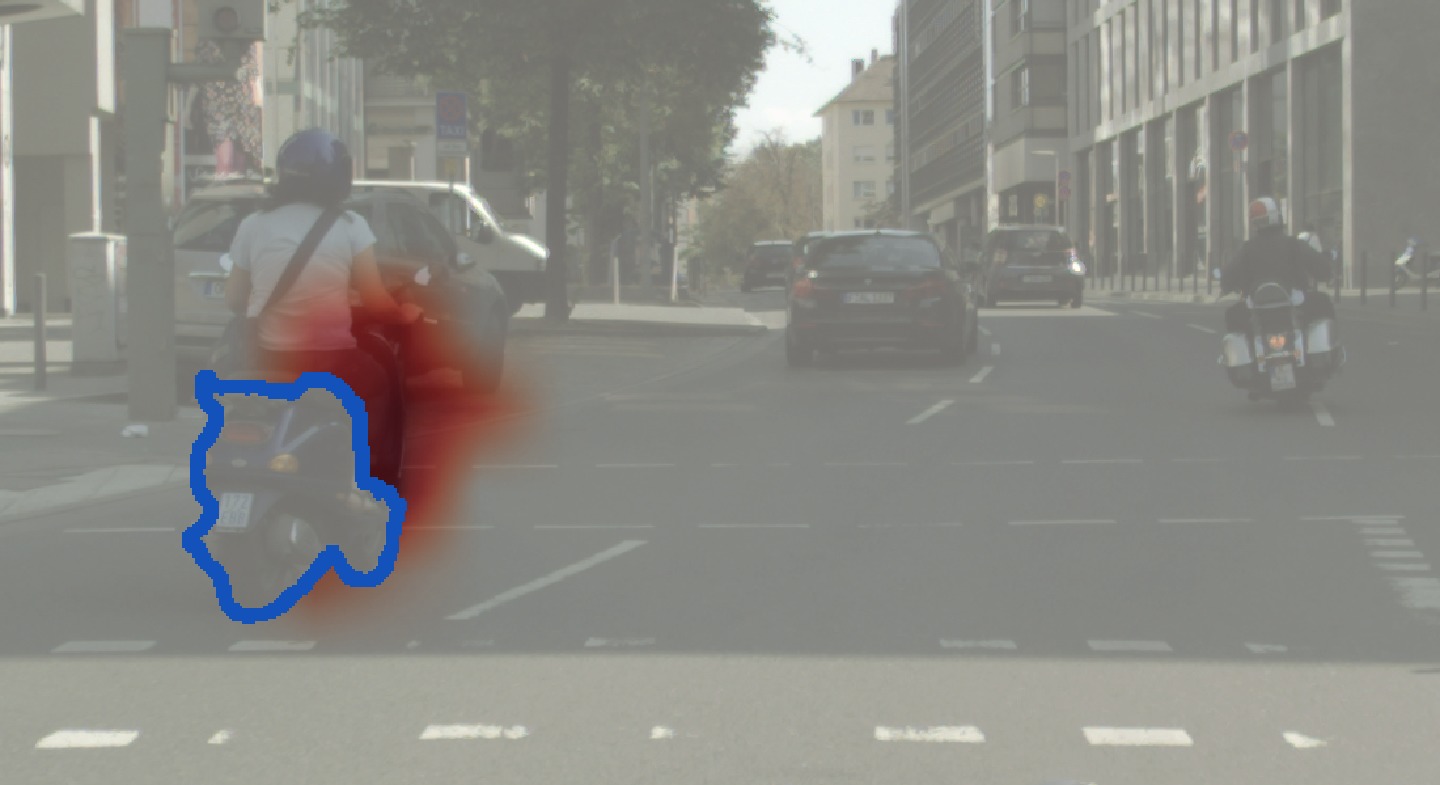} &  {\footnotesize{}}
			\includegraphics[width=0.24\linewidth, height=0.075\textheight]{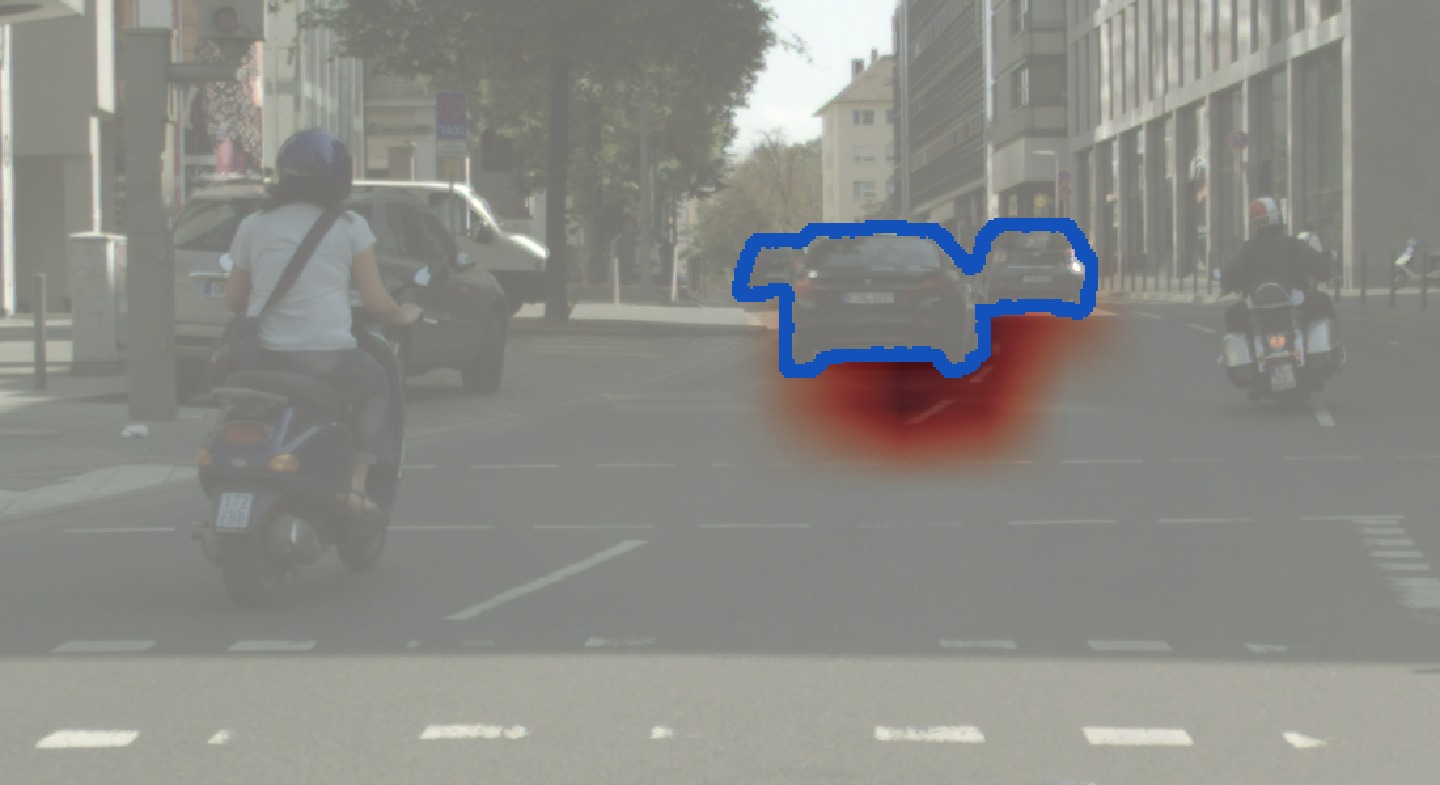} \tabularnewline

 			\includegraphics[width=0.24\textwidth, height=0.075\textheight]{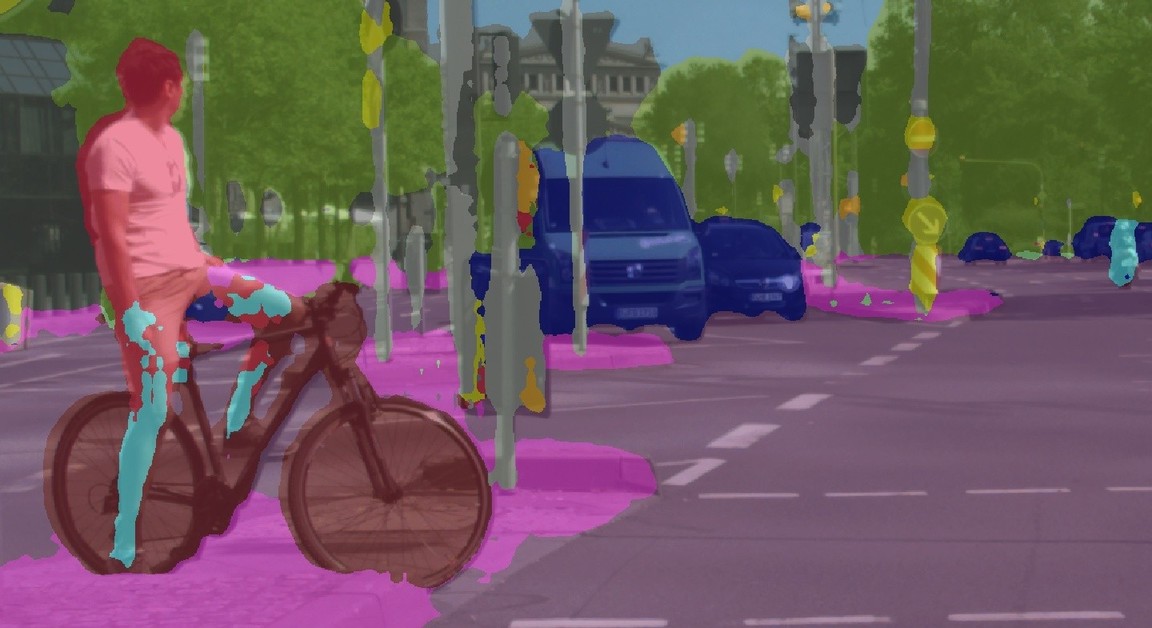} & {\footnotesize{}}
 			\includegraphics[width=0.24\textwidth, height=0.075\textheight]{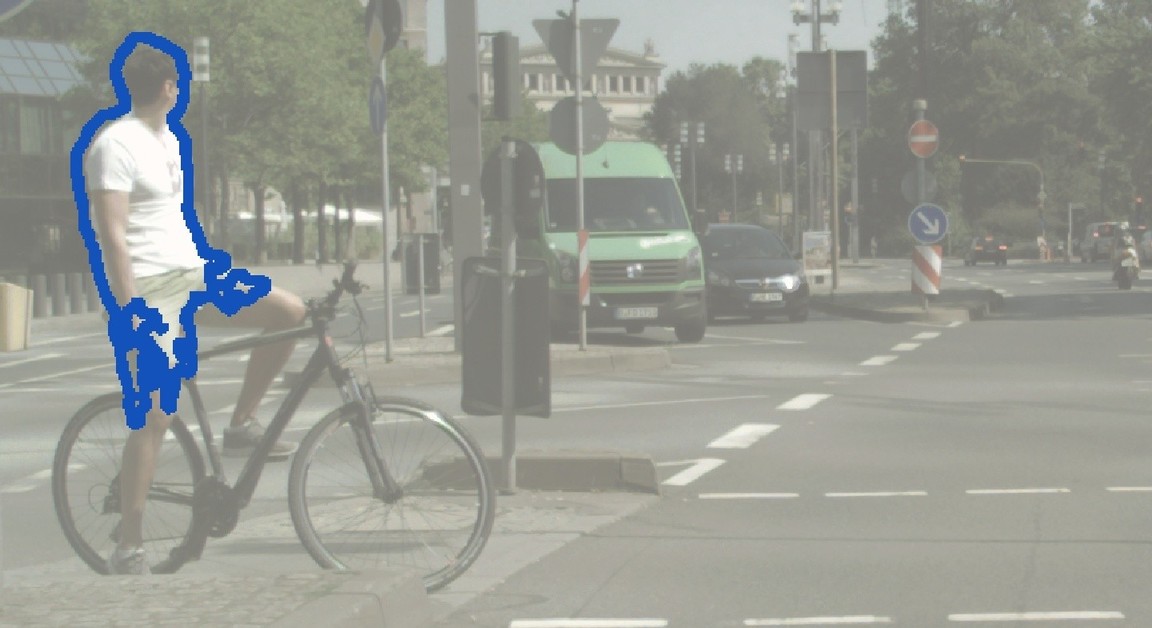} & {\footnotesize{}}
 			\includegraphics[width=0.24\textwidth, height=0.075\textheight]{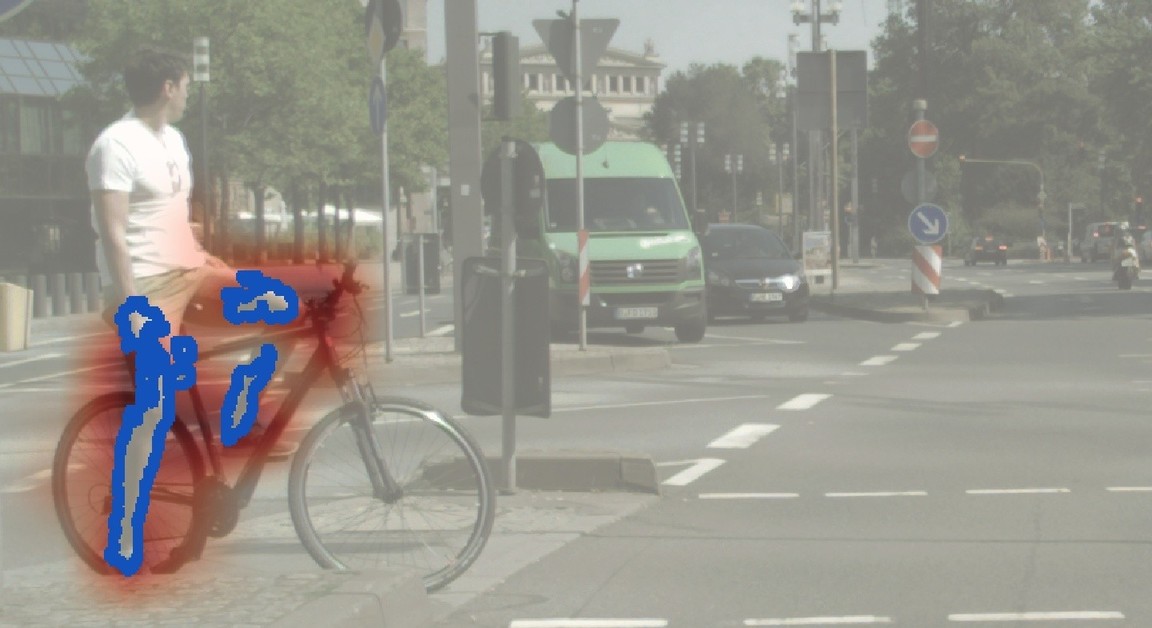} & {\footnotesize{}}
 			\includegraphics[width=0.24\textwidth, height=0.075\textheight]{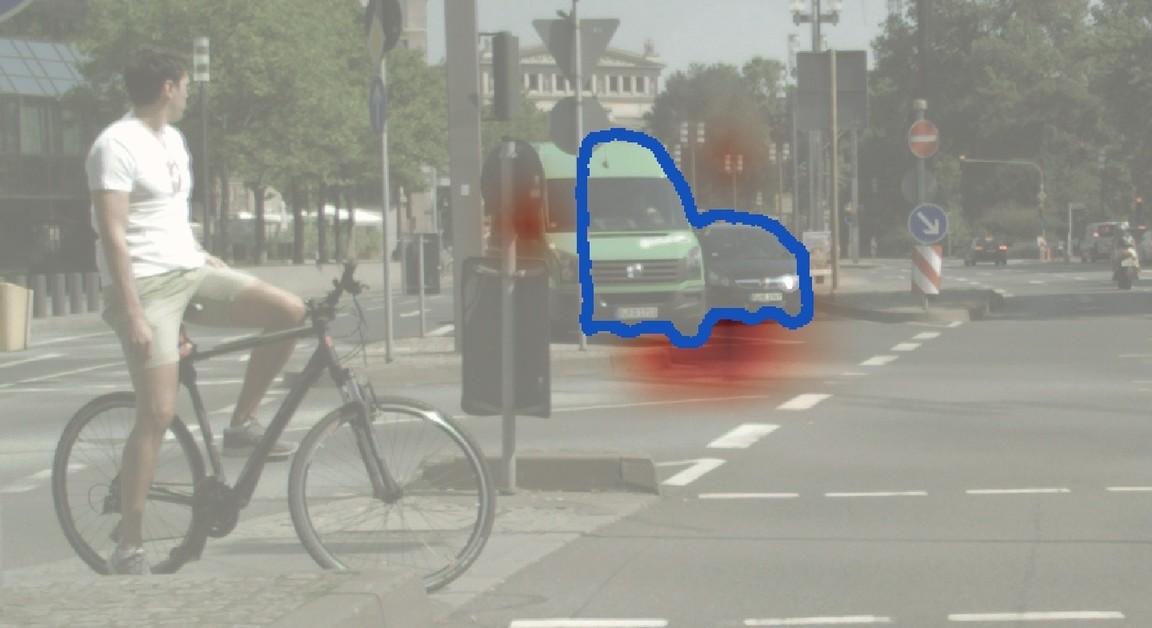} 
 			\tabularnewline
 			(a) Semantic segment. & \multicolumn{3}{c}{(b) Context explanations (red) for different segments (outlined in blue)}  \tabularnewline
		\end{tabular}\hfill{}
		\par\end{centering}
	\caption{Context explanations by grid saliency for semantic segmentation \cite{Sandler2018MobileNetV2IR,Chen2018EncoderDecoderWA} on Cityscapes~\cite{Cordts2016Cityscapes}.  Grid saliency not only can contextually explain correct predictions: in the first row the network looks at the motorbike to correctly predict the class rider (light blue); but can also explain erroneous predictions: in the second row the upper body of the rider is incorrectly predicted as person, but for this prediction the bicycle is not salient in contrast to the correctly predicted legs of the rider.}
	\vspace{-1em}
	\label{fig:concept}
\end{figure}

\begin{figure}
	\vspace{-0.5em}
	\begin{centering}
		\setlength{\tabcolsep}{0em}
		\renewcommand{\arraystretch}{0}
		\par\end{centering}
	\begin{centering}
		\hfill{}%
		\begin{tabular}{@{}c@{}c@{\hskip 0.05in}c@{}c@{}}
		
 			\includegraphics[width=0.24\textwidth, height=0.09\textheight]{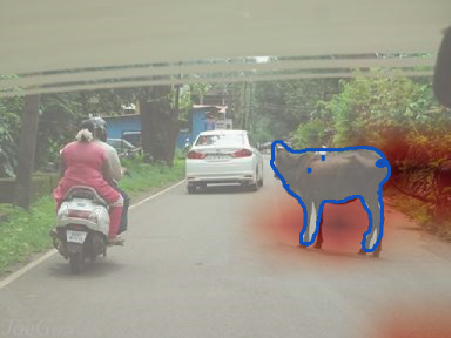} & {\footnotesize{}}
			\includegraphics[width=0.24\textwidth, height=0.09\textheight]{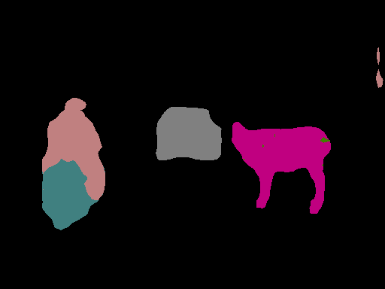}& {\footnotesize{}}
			\includegraphics[width=0.24\textwidth, height=0.09\textheight]{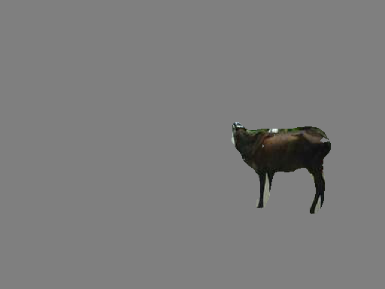} & {\footnotesize{}}
			\includegraphics[width=0.24\textwidth, height=0.09\textheight]{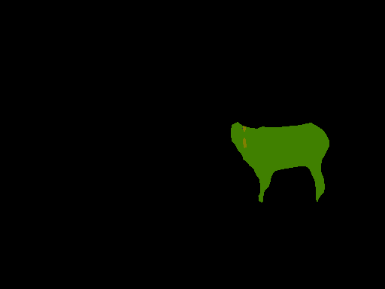} 
			\tabularnewline

			(a) Context explanation & (b) Semantic segm. & (c) Image w/o context & (d) Segm. w/o context
			
		\end{tabular}\hfill{}
		\par\end{centering}
		\caption{Effect of context on semantic segmentation and the context explanation provided by grid saliency for an erroneous prediction, the image is taken from MS COCO~\cite{Lin2014MicrosoftCC}. The grid saliency (a) shows the responsible context for misclassifying the cow (green) as horse (purple) in the semantic segmentation (b). It shows the training bias that horses are more likely on road than cows. Removing the context (c) yields a correctly classified cow (d).}
	\vspace{-1.8em}
	\label{fig:context_cow}
\end{figure}

In many real-world scenarios, the presence of an object, its location and appearance are highly correlated with the contextual information surrounding this object, such as the presence of other nearby objects or more global scene semantics.
For example, in the case of an urban street scene, a cyclist is more likely to co-occur on a bicycle and a car to appear on the road below sky and buildings (cf. objects and their context explanations in Fig. \ref{fig:concept}). 
These semantic correlations are inherently present in real-world data. A data-driven model, such as a deep neural network, is prone to exploit these statistical biases during training in order to improve its prediction performance.
These biases picked up by the network during training can lead to erroneous predictions and  impair network generalization (cf. the effect of context on misclassification in Fig. \ref{fig:context_cow}).
An effective and safe utilization of deep learning models for real-world applications, e.g. autonomous driving, requires a good understanding of these contextual biases inherent in the data and the extent to which a learned model incorporated them into its decision making process.

Saliency methods~\cite{Simonyan2013DeepIC,Selvaraju2016GradCAMWD,Fong2017InterpretableEO,Adebayo2018SanityCF,Chang2018ExplainingIC,Zintgraf2017VisualizingDN} have become a popular tool to explain predictions of a trained model by highlighting parts of the input that presumably have a high relevance for its predictions.
However, to the best of our knowledge the existing saliency methods are mostly focused on image classification networks and thus are not able to spatially differentiate between prediction explanations.
In this work, we propose a way to extend existing saliency methods designed for image classification towards (pixel-level) dense prediction tasks, which allows to generate spatially coherent explanations by exploiting spatial information in dense predictions. We call our approach \emph{grid saliency}, which is a perturbation-based
saliency method,
formulated as an optimization problem of identifying the minimum unperturbed area of the image needed to retain the network predictions inside a target object region. 
As our grid saliency allows to differentiate between objects and their associated context areas in the saliency map, we specifically explore its potential to produce \emph{context explanations} for semantic segmentation networks. The contextual information is known to be one of the essential recognition cues
~\cite{Mottaghi2014TheRO, uijlings2012visual, azaza2018context}, thus we aim to investigate which local and global context is the most relevant for the network class predictions inside a target object area (see Fig. \ref{fig:concept} and Fig. \ref{fig:context_cow} for examples).

In real-world scenarios, context biases are inherently present in the data.
To evaluate whether the proposed grid saliency is sensitive to context biases and has the ability to detect them, we introduce a synthetic toy dataset for semantic segmentation, generated by combining MNIST digits~\cite{MNIST} with different fore- and background textures, for which we artificially induce a context bias between the digit and the background texture. Besides detecting the mere presence of the context bias, we also analyze the ability of our method to localize it in the image (see Sec. \ref{sec:experiments_toydata_simple}). We employ this dataset to compare our approach with different baselines, i.e., introduced extensions of gradient-based saliency methods of \cite{Simonyan2013DeepIC,Sundararajan2017AxiomaticAF,Smilkov2017SmoothGradRN} to produce context explanations. We show that the proposed dataset can serve as a valid benchmark for assessing the quality of saliency methods to detect and localize context biases. We find that gradient-based techniques, in contrast to our grid saliency, are ill-suited for the context bias detection. By design they tend to produce noisy saliency maps which are not faithful to the context bias present in the data, whereas our method has higher sensitivity for context bias and thus can also precisely localize it in the image. We further evaluate grid saliency performance to produce context explanations for semantic segmentation on the real-world Cityscapes dataset \cite{Cordts2016Cityscapes} and experimentally show that the produced context explanations faithfully reflect spatial and semantic correlations present in the data, which were thus picked up by the segmentation network \cite{Sandler2018MobileNetV2IR, Chen2018EncoderDecoderWA}.

To the best of our knowledge, we are the first to extend saliency towards dense-prediction tasks and use it to produce context explanations for semantic segmentation.
In summary, our contributions are the following: (1) We propose an extension of saliency methods designed for classification towards dense prediction tasks. (2) We exploit the proposed grid saliency to produce context explanations for semantic segmentation and show its ability to detect and localize context biases. (3) We create a synthetic dataset to benchmark the quality of produced explanations as well as their effectiveness for context bias detection/localization. (4) We investigate the faithfulness of context explanations for semantic segmentation produced by the grid saliency on real-world data.

 \vspace{-0.5em}
 
 \section{Related Work}
 \label{sec:related_work}
 
 \vspace{-0.5em}
 
\paragraph{Explanations}
Many methods attempt to interpret the network decision making process by producing explanations via bounding boxes \cite{Karpathy2015DeepVA,JosOramas2016ModelingVC} or attributes \cite{Gulshad2019InterpretingAE}, providing textual justifications \cite{Park2018MultimodalEJ,Hendricks2016GeneratingVE,Dong2017ImprovingIO} or generating low-level visual explanations \cite{Simonyan2013DeepIC,Zhou2016LearningDF, Selvaraju2016GradCAMWD, Fong2017InterpretableEO}. 
Our work builds on top of the latter approaches, also known as \emph{saliency methods}, which try to identify the image pixels that contribute the most to the network prediction.
These methods mostly focus on the task of image classification
and can be divided into two categories: gradient-based and perturbation-based methods.

Gradient-based methods \cite{Zeiler2014VisualizingAU,Simonyan2013DeepIC, Springenberg2014StrivingFS} compute a saliency
map that visualizes the sensitivity of each image pixel to a specific class prediction, which is obtained by backpropagating the gradient
for the prediction with respect to the image and estimating how moving along the
gradient influences the class output. 
To circumvent noise and visual diffusion in saliency maps, 
\cite{Sundararajan2017AxiomaticAF} proposes to sum up the gradients over the intensity-scaled input versions, while 
\cite{Smilkov2017SmoothGradRN} averages over many noisy samples of the input.
Other methods \cite{Selvaraju2016GradCAMWD,Bach2015OnPE,Zhang2017TopDownNA} explore integrating network activations into their saliency maps.
Gradient-based methods mostly rely on heuristics for backpropagation and as has been shown by \cite{Adebayo2018SanityCF} may provide explanations which are not faithful to the model or data.

Perturbation-based methods \cite{Fong2017InterpretableEO,Dabkowski2017RealTI,Zintgraf2017VisualizingDN,Chang2018ExplainingIC} 
evaluate the class prediction change with respect to a perturbed image, e.g. in which specific regions of the image are either replaced with the mean image values or removed by applying blur or Gaussian noise.
The approach in \cite{Fong2017InterpretableEO} formulates this output change as an optimization problem of the original and perturbed image, while \cite{Dabkowski2017RealTI} estimates the perturbation mask by training an auxiliary network.
To account for image discontinuities, \cite{Zintgraf2017VisualizingDN,Chang2018ExplainingIC} compute the saliency of the masked region by marginalizing it out either over neighboring
image regions or by conditioning the trained generative model on the non-masked region and then estimating the classification change. 
Perturbation approaches might be vulnerable to network artifacts resulting in arbitrary saliency regions 
\cite{Chang2018ExplainingIC}. To overcome this, \cite{Fong2017InterpretableEO, Dabkowski2017RealTI} resort to perturbe large image regions.

The above methods are limited to explanations of image classification. In this work, we propose a way to extend them to produce grid saliency maps for dense prediction networks as well (see Sec. \ref{sec:theory}). 
We showcase the usability of grid saliency for context explanations of semantic segmentation (see Sec. \ref{sec:experiments_toydata_simple} and \ref{sec:experiments_cityscapes}). The closest related work to produce context explanations is \cite{AlShedivat2018ContextualEN}, which proposes a network that jointly learns to predict and contextually explain its decisions. In contrast to \cite{AlShedivat2018ContextualEN}, we focus on the post hoc model explanations and are not limited to the image classification task.

\paragraph{Semantic segmentation}
CNNs have become a default technique for semantic segmentation \cite{Zhao2017PyramidSP,Pohlen2017FullResolutionRN,Lin2017RefineNetMR}.
\cite{Shelhamer2015FullyCN} first showcased the use of CNNs for segmentation. 
Since then, multiple techniques have been proposed, from utilizing dilated convolutions \cite{Chen2015SemanticIS,Yu2016MultiScaleCA} and post-processing smoothing operations \cite{Lin2016EfficientPT,Zheng2015ConditionalRF,Chandra2016FastEA,Jampani2016LearningSH} to employing spatial pyramid pooling \cite{Chen2018DeepLabSI,Zhao2017PyramidSP,Ghiasi2016LaplacianPR,Chen2018EncoderDecoderWA} and encoder/decoder architectures \cite{Hong2015DecoupledDN,Noh2015LearningDN,Ronneberger2015UNetCN,Badrinarayanan2016SegNetAD}. 
As the accuracy of these methods comes at a high computational cost, there has been an increasing interest in developing real-time semantic segmentation networks with low memory needs \cite{Poudel2019FastSCNNFS,Sandler2018MobileNetV2IR,Paszke2017ENetAD,Mazzini2018GuidedUN,Zhao2018ICNetFR}.  
In this work, we aim to produce context explanations for semantic segmentation. To investigate the effectiveness of our approach and show its generalization across architectures, we 
employ DeepLabv3+ \cite{Chen2018EncoderDecoderWA} and U-Net \cite{Ronneberger2015UNetCN} with different backbones \cite{Sandler2018MobileNetV2IR,He16, simonyan2014very}.

\vspace{-0.5em}

\section{Method}
\label{sec:theory}

\vspace{-0.5em}

In Sec.~\ref{subsec:theory_perturbation} we introduce the grid saliency method, which allows to produce spatially coherent explanations for dense predictions, 
and present a way to use it for context explanations of semantic segmentation. 
Next, in Sec. \ref{subsec:theory_gradient} we extend the popular gradient-based saliency methods \cite{Simonyan2013DeepIC,Sundararajan2017AxiomaticAF,Smilkov2017SmoothGradRN} to produce spatial explanations as well, which we later compare with grid saliency in Sec. \ref{sec:experiments_toydata_simple}. 

\subsection{Grid Saliency via Perturbation}
\label{subsec:theory_perturbation}

Let \(f : I \rightarrow O\) denote the prediction function, e.g. a deep neural network, 
which maps a grid input space $I = \mathbb{R}^{H_I \times W_I \times C_I}$ to a grid output space 
$O = \mathbb{R}^{H_O \times W_O \times C_O}$, 
where $W$ and $H$ are the respective width and height of the input and output, 
and $C_I$ and $C_O$ are the number of input channels (e.g. $3$ or $1$ for images) and output prediction channels (e.g. number of classes for semantic segmentation). To keep the discussion concrete,
we consider only images as input, $x \in I$, and per-pixel dense predictions of the network $f(x) \in O$ as output.
The goal is to find the smallest saliency map $M\in [0,1]^{H_I \times W_I}$ that must retain in the image $x$ in order to preserve the network prediction in the request mask area $R \in \{0,1\}^{H_O \times W_O}$ for class (channel) $c \in \{1, ..., C_O\}$. Further on, for simplicity we assume that the input and output spatial dimensions are the same. 

Our method builds on top of perturbation saliency methods \cite{Fong2017InterpretableEO,Dabkowski2017RealTI,Zintgraf2017VisualizingDN,Chang2018ExplainingIC} designed for image classification. They aim to find salient image regions most responsible for a classifier decision by replacing parts of the image with uninformative pixel values, i.e. perturbing the image, and evaluating the corresponding class prediction change. 
We follow the same image perturbation strategy as \cite{Fong2017InterpretableEO}.
Let $p$ denote a perturbation function that removes information from an image $x$ outside of the saliency $M$. For example, such perturbation function can be the interpolation between $x$ and $a\in I$, where $a$ can be a constant color image, gaussian blur, or random noise. In this case, $p(x, M) = x \circ M + a \circ (1 - M)$.
Note, that in practice $M$ operates on a lower resolution to avoid adversarial artifacts \cite{Fong2017InterpretableEO, Dabkowski2017RealTI} and 
is later upsampled to the original image resolution. In addition, the pixel values of the perturbed image $p(x, M)$ are clipped to preserve the range of the original image space.

With this notation in hand, we can formulate the problem of finding the saliency map $M$ for the prediction of class $c$ as the following optimization problem:
\begin{align}
M^{*}(x, c) = \argmin_{M} \lambda \cdot \|M\|_1 + \|\max(f_c(x) - f_c(p(x,M)), 0)\|_1 , \label{eq1}
\end{align}
where $\| \cdot \|_1$ denotes the $l_1$ norm and $f_c(x)$ is the network prediction for class $c$ .
The first term can be considered as a mask loss that minimizes the salient image area and perturbs the original image as much as possible. The second term serves as a preservation loss which ensures that the network prediction $f_c(p(x,M))$ for class $c$ on the perturbed image $p(x,M)$ reaches at least the confidence of the network prediction $f_c(x)$ on the original unperturbed image. Thus, the second loss term can be considered as a penalty for not meeting the constraint $f_c(p(x,M)) \geqslant f_c(x)$, hence the use of $\text{max}(\cdot,0)$ in Eq.~\ref{eq1}. The parameter $\lambda$ controls the sparsity of $M$. 

We then can spatially disentangle explanations given in the saliency map $M$ for the network predictions in the requested area of interest $R$ from the explanations for the other predictions, by restricting the preservation loss to the request mask $R$ in Eq. (\ref{eq1}): 
\begin{align}
M^{*}_{\text{grid}}(x, R, c) = \argmin_{M} \lambda \cdot \|M\|_1 + \frac{\|R \circ\max( f_c(x) - f_c(p(x,M)),0)\|_1}{\|R\|_1} . \label{eq2}
\end{align}
Further on, we will refer to $M^{*}_{\text{grid}}$  in Eq. (\ref{eq2}) as a grid saliency map.

\paragraph{Context explanations for semantic segmentation}

We now adapt the grid saliency formulation from Eq. (\ref{eq2}) to specifically provide \emph{context explanations} for the requested area of interest $R$. Context explanations are of particular interest for semantic segmentation, as 
context often serves as one of the main cues for semantic segmentation networks (see Fig. \ref{fig:context_cow}). Thus, here we focus on context explanations for semantic labelling predictions and assume that $R$ is the area covering the object of interest in the image $x$. To optimize for salient parts of the object context, we integrate the object request mask $R$ in the perturbation function. 
For the request mask $R$, the perturbed image $p(x, R) \in I$ contains only the object information inside $R$ and all the context information outside $R$ is removed (with a constant color image $a$). For optimization, we will now use this new perturbed image $p(x, R)$ instead of the maximally perturbed image $ p(x, M=0) = a$ and denote the context perturbation function as 
$p_{\text{context}}(x,R,M) = x \circ M + p(x, R) \circ (1 - M)$.

The context saliency map for class $c$ and request object $R$ can be obtained via optimization of
\begin{align}
M^{*}_{\text{context}}(x, R, c) = \argmin_{M} \lambda \cdot \|M\|_1 + \frac{ \|R \circ \max(f_c(x) -f_c(p_{\text{context}}(x,R,M)), 0)\|_1}{\|R\|_1}, \label{eq3}
\end{align}
where the saliency map is optimized to select the minimal context necessary to at least yield the original prediction for class $c$ inside the request mask $R$. Note that, $M^{*}_{\text{context}}$ can be an empty mask if no context information is needed to recover the original prediction inside $R$.


\subsection{Gradient-Based Variants}
\label{subsec:theory_gradient}
Another way to produce spatially coherent saliency maps is to make use of the popular gradient-based saliency methods. Thus, we additionally consider the Vanilla Gradient (VG) \cite{Zeiler2014VisualizingAU}, Integrated Gradient (IG) \cite{Sundararajan2017AxiomaticAF}, and SmoothGrad (SG) \cite{Smilkov2017SmoothGradRN} saliency methods. 

Let \(G(x,c) = \partial g_c(x) / \partial x \in \mathbb{R}^{H_I \times W_I \times C_I}\) denote the gradient of the classification network prediction \(g_c(x) \in \mathbb{R}\) for class $c$ with respect to the input image \(x \in I\).
For the classification task, the saliency maps of VG, IG and SG are computed as:
\begin{align}
\begin{split}
M^{\text{VG}}(x,c) = \sum_{c \in C_I}\big\rvert G(x,c)\big\lvert, \ \ \ \ \ \ 
M^{\text{SG}}(x,c) = \sum_{c \in C_I}\bigg\rvert\frac{1}{n} \sum_{k=1}^{n} G\left(x + \mathcal{N}(0, \sigma^2),c\right)\bigg\lvert,\\ 
M^{\text{IG}}(x,c) = \sum_{c \in C_I}\bigg\rvert\frac{1}{n} \sum_{k=1}^{n} G\left(\frac{k}{n}x,c\right)\bigg\lvert,
\label{eq:standard_saliency}
\end{split}
\end{align}
where $n$ is the number of approximation steps for IG or the number of samples for SG, and $\mathcal{N}(0, \sigma^2)$ represents Gaussian noise with standard deviation $\sigma$.

Following Sec. \ref{subsec:theory_perturbation}, we next extend the above approaches to produce the saliency $M$ for dense predictions $f_c(x) \in O$ and to spatially disentangle explanations given in the saliency $M$ for the network predictions in the request area $R$ from other predictions. 
For a given input \(x\) and a binary request mask \(R\), we denote the normalized network prediction score for class $c$ in the request area $R$ as $S(x, R, c) = \|R \circ f_c(x)\|_1 / \|R\|_1$, \(S(x, R, c) \in \mathbb{R}\).
Similarly to \(G(x, c)\), we define \(G_{\text{grid}}(x, R, c) \defeq \partial S(x, R, c) / \partial x \in \mathbb{R}^{H_I \times W_I \times C_I}\) which directly yields \(M^{\text{VG/SG/IG}}_{\text{grid}}(x, R, c)\) by replacing \(G(x, c)\) in Eq. \ref{eq:standard_saliency} with \(G_{\text{grid}}(x, R, c)\).
For the gradient-based context saliency, as in Sec. \ref{subsec:theory_perturbation} only salient pixels outside of the object area are considered, i.e. 
\begin{align}
M_{\text{context}}^{\text{VG/IG/SG}}(x, R, c) \defeq (1-R) \circ M_{\text{grid}}^{\text{VG/IG/SG}}(x, R, c).\label{eq:grad_based_context_sal}
\end{align}
Gradient-based saliency maps are prone to be noisy. Thus, to circumvent this and also make them more comparable to the lower resolution perturbation-based grid saliency, in our experiments we apply the spatial mean filter on top of the saliency map with a \((W_{I} / W_{S}) \times (H_{I} / H_{S})\) kernel and stride, where \(W_{S} \times H_{S}\) is the resolution of the perturbation-based saliency map.

\vspace{-0.5em}

\section{Context Bias Detection on Synthetic Data}
\label{sec:experiments_toydata_simple}

\vspace{-0.5em}

Although context biases are inherently present in real-world data, in practice it is hard to measure and alter these correlations and thus to employ this data for benchmarking context bias detection.
In order to evaluate whether the grid saliency is sensitive to context biases and has the ability to detect them, in Sec.~\ref{subsec:toydata_benchmark} we introduce a synthetic dataset for semantic segmentation with artificially induced context biases.\footnote{The code for generating the proposed synthetic dataset with induced context biases can be found here:  \url{https://github.com/boschresearch/GridSaliency-ToyDatasetGen}.} We use this dataset to compare the grid saliency methods proposed in Sec.~\ref{sec:theory} and show that this dataset can serve as a valid benchmark for assessing the quality of saliency methods for context bias detection and localization, see Sec.~\ref{subsec:toydata_results}.

\subsection{Benchmark for Context Bias Detection}
\label{subsec:toydata_benchmark}

\paragraph{Dataset}

\begin{figure}[t!]
	\begin{centering}
		\hfill \input{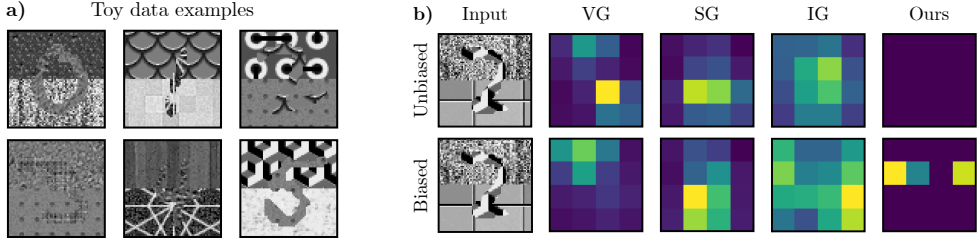} 	\hfill
		\end{centering}
	\caption{\label{fig:toy_dataset} \textbf{a)} Synthetic dataset images, see Sec.~\ref{subsec:toydata_benchmark} for details. 
		\textbf{b)} Context saliency maps of different methods.  For the image on the left, the different context saliency maps $M_{\text{context}}$ are shown for the networks \cite{Sandler2018MobileNetV2IR} trained on the unbiased (top row) and biased (bottom row) dataset versions. For the biased version, the digit \emph{2} is biased with the background texture in the top half of the image. Our perturbation-based grid saliency is able to precisely detect this bias, in contrast to other methods. 
	}
	\vspace{-1.3em}
\end{figure}

The proposed synthetic toy dataset consists of gray scale images of size \(64 \times 64\) pixels, generated by combining upscaled digits from MNIST ~\cite{MNIST} with foreground and background textures from \cite{BackgTextGen,BackgTextPatterns}, as can be seen in Fig. \ref{fig:toy_dataset} (a). 
In order to introduce different context information for the digit, two background textures are used for the upper and lower half of the image.
For each synthetic image a corresponding segmentation ground truth is generated, where the MNIST digit defines the mask and the semantic class (including a background class). 

To evaluate the ability of saliency methods to explain context biases, we propose to generate biased and unbiased versions of the dataset.
For the unbiased version (DS$^\mathtt{{no{\text -}bias}}$), all fore- and background textures appear with equal probability for all digits. For the biased version, a single digit class is coupled with a specific background texture, located randomly either in the upper or lower half of the background. We consider two variants of it, with a weakly DS$^\mathtt{w{\text -}bias}$ and strongly induced bias DS$^\mathtt{s{\text -}bias}$.
For the dataset with a strongly induced bias DS$^\mathtt{s{\text -}bias}$, a specific texture appears if and only if a certain digit class is present.
For a weakly induced bias DS$^\mathtt{w{\text -}bias}$, a specific texture always appears along with the biased digit but also uniformly appears with other digits.
From the pool of \(25\) textures, \(5\) textures are chosen randomly to induce context bias for one of $10$ digits. For all \(50\) texture/digit combinations, a weakly and strongly biased dataset variant with train/test splits is generated.

\paragraph{Evaluation metrics} 
To evaluate the extent to which a network is able to pick up a context bias present in the training data, we propose to measure on the unbiased DS$^\mathtt{{no-bias}}$ test set the performance of the network trained on the biased DS$^\mathtt{w/s{\text -}bias}$ training set, using the standard Intersection over Union (IoU, see \cite{Shelhamer2015FullyCN}) metric for semantic segmentation.
If the network has picked up the bias, we expect to see a significant drop in IoU for the biased digit segmentation.

To benchmark how well different saliency methods (perturbation- or gradient-based) can detect the context bias of semantic segmentation networks, we propose to evaluate to which extent a context saliency map $M_{\text{context}}(x, R, c)$ (cf. Eq.~\ref{eq3} and \ref{eq:grad_based_context_sal} in Sec.~\ref{sec:theory}) for the request object mask $R$ and its corresponding class $c$ is concentrated on the ground truth context area \(\mathrm{C} = 1 - R_{GT}\), where $R_{GT}$ is the ground truth mask of the object $R$, by using a context bias detection metric (CBD):
\begin{align}
\text{CBD}(x, R, c) = {\| \mathrm{C} \circ M_{\text{context}}(x, R, c)\|_1} / {\|\mathrm{C}\|_1}.\label{CBD}
\end{align}
To benchmark the ability of different saliency methods to localize a context bias,
we propose to measure how much of the context saliency $M_{\text{context}}(x, R, c)$ falls into the 
ground truth biased context area $\mathrm{C}_{bias}$, the upper or lower half of the image $x$ by the design of DS$^\mathtt{w/s{\text -}bias}$, where $\mathrm{C}_{\text{bias}}$ is a binary mask of the biased context area. We refer to this metric as a context bias localization (CBL): 
\begin{align}
\text{CBL}(x, R, c) = {\|\mathrm{C}_{\text{bias}} \circ M_{\text{context}}(x, R, c) \|_1} / {\|\mathrm{C} \circ M_{\text{context}}(x, R, c)\|_1}. \label{CBL}
\end{align}
In our experiments we report mIoU and the mean CBD and CBL measures (mCBD, mCBL) per biased digit class $c$, averaging the results across all images in the DS$^\mathtt{w/s{\text -}bias}$ test set, $5$ randomly selected bias textures and $5$ different initial random seeds for the training set generation.

\subsection{Experimental Results}
\label{subsec:toydata_results}

\paragraph{Implementation details}
\label{subsec:toydata_experimental_setup}
We use the U-Net \cite{Ronneberger2015UNetCN} architecture with a VGG16 \cite{simonyan2014very}
backbone. 
As request mask, the segmentation prediction of the target digit is used.
The saliency maps with a size of \(4 \times 4\) are optimized using SGD with momentum of \(0.5\) and a learning rate of \(0.2\) for \(100\) steps starting with a \(0.5\) initialized mask. A weighting factor of \(\lambda = 0.05\) is used (see Eq. \ref{eq3}). A constant color image is used for perturbation.
Further implementation details are provided in the supp. material.

\paragraph{Bias in the trained network} 

\begin{figure}[t]
	\centering
	\vspace{-0.5em}
  \input{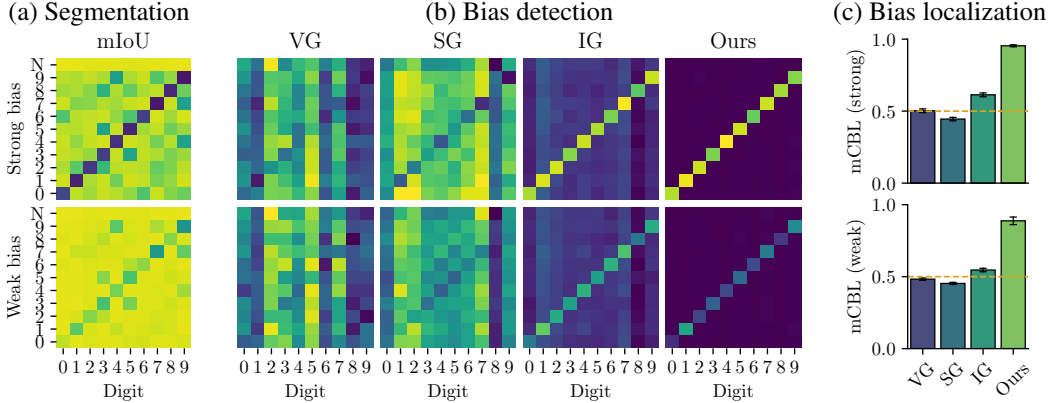}
  \caption{
  Context bias detection/localization of different saliency methods. 
  \textbf{(a)} shows the segmentation mIoU and \textbf{(b)} \& \textbf{(c)} report the context bias detection and localization results, for the strongly and weakly biased datasets, respectively. In \textbf{(a)}-\textbf{(b)}, the \(y\) axis indicates the network bias towards the digit (0-9), with \(N\) denoting the unbiased setting. The \(x\) axis indicates the corresponding digit result.  
  In contrast to VG~\cite{Zeiler2014VisualizingAU}, IG~\cite{Sundararajan2017AxiomaticAF}, and SG~\cite{Smilkov2017SmoothGradRN}, our grid saliency is able to accurately detect (see diagonal elements in \textbf{(b)}) and localize \textbf{(c)} the induced context bias, see Sec.~\ref{subsec:toydata_results} and suppl. material for details. 
  }
	\vspace{-1.8em}
  \label{fig:results_toy_saliencies}
\end{figure}

We first investigate if the networks trained on \(\text{DS}^{\text{w-bias}}\) and \(\text{DS}^{\text{s-bias}}\) have picked up the induced weak and strong context biases in the training data.
For this purpose, we evaluate their performance in terms of mIoU on the unbiased dataset \(\text{DS}^{\text{no-bias}}\) and report the results in Fig. \ref{fig:results_toy_saliencies} (a), which visualizes the digit-wise mIoU with respect to the biased digit.
The first row of the heat map (labeled as N) in Fig. \ref{fig:results_toy_saliencies} (a) shows the performance of the networks trained on \(\text{DS}^{\text{no-bias}}\). We observe a clear drop in performance for biased digits (diagonal elements) in comparison to the first row. As expected, the performance drop is higher for the stronger bias. Moreover, the mIoU of the unbiased digits (non-diagonal elements) is also affected by the introduced bias. For example, inducing a bias for the digit nine leads to a decreased performance for the digit four (see second row in Fig. \ref{fig:results_toy_saliencies} (a)). We observe that this effect mostly occurs for the similar looking digits and, most likely, is caused by the fact that on the unbiased dataset the bias textures also occur with the unbiased digits, resulting in the confusion of similar looking digits for the network.
From the observed mIoU drop for biased digits we can conclude that the networks have picked up the introduced bias. However, in real world it is often impossible to collect fully unbiased data. Thus, we next evaluate the ability of our grid saliency to detect context bias only using the biased data.

\paragraph{Context bias detection with grid saliency}
In the last column of Fig. \ref{fig:results_toy_saliencies} (b), we report the context bias detection results for our perturbation-based grid saliency method, described in Sec.~\ref{subsec:theory_perturbation}, using the CBD metric~(see Eq. \ref{CBD}).
The mCBD values are visualized with respect to networks trained on data biased to different digits \(\text{DS}^{\text{s/w-bias}}\) (y-axis) and for the different digit classes (x-axis) in the biased test set.
The only exception is the first row (labeled as N), where for comparison we show the results with no bias, for the network trained on \(\text{DS}^{\text{no-bias}}\).
We observe that our grid saliency shows substantial evidence of context bias for digits with induced bias (diagonal elements), both strong and weak. 
Even for the weak bias in Fig. \ref{fig:results_toy_saliencies} (b) the grid saliency still clearly differentiates between biased and unbiased digits (diagonal vs. non-diagonal elements). 
Note that the bias detection using the grid saliency does not require an unbiased test set.
In the suppl. material, we also study the influence of the bias texture as well as the choice of hyperparameters.

\paragraph{Comparison across different saliency methods}
In Fig. \ref{fig:results_toy_saliencies} (b) and (c) we compare our perturbation-based grid saliency with the context saliency extensions of gradient-based methods, i.e. VG, SG, and IG (see Eq. \ref{eq:standard_saliency}), using mCBD and mCBL metrics proposed in Sec.~\ref{subsec:toydata_benchmark}. From Fig. \ref{fig:results_toy_saliencies} (b) we see that VG and SG are not able to reliably detect the context bias, while IG achieves a comparable performance to our grid saliency. However, in contrast to the perturbation saliency, IG has also high mCBD values for unbiased digits, complicating its use in practice, as one would need to tune a reliable detection threshold, which is particularly challenging for weakly biased data.

\paragraph{Context bias localization}
In addition to the detection performance, we evaluate how well the saliency methods localize the context bias using the mCBL measure. Fig. \ref{fig:results_toy_saliencies} (c) shows that VG, IG, and SG have low localization performance, comparable to random guessing ($\sim0.5$ mCBL), while our grid saliency is the only method which is able to accurately localize the context bias (mCBL above $0.9$) on both strongly and weakly biased data.
We evaluate the ability of our perturbation grid saliency to detect and localize context bias across different semantic segmentation networks. For U-Net \cite{Ronneberger2015UNetCN} with the VGG16 \cite{simonyan2014very}, ResNet18 \cite{He16} and MobileNetv2 \cite{Sandler2018MobileNetV2IR} backbones, we observe similar bias detection and localization performance (see also supp. material).

\vspace{-0.5em}

\section{Cityscapes Experiments}
\label{sec:experiments_cityscapes}

\vspace{-0.5em}

Motivated by the success of our perturbation-based grid saliency at detecting context biases on the synthetic dataset described in Sec.~\ref{sec:experiments_toydata_simple}, we next apply it to Cityscapes~\cite{Cordts2016Cityscapes} in order to produce and analyze context explanations for semantic segmentation in real-world scenes.

\paragraph{Experimental setup}
\label{subsec:cityscapes_experimental_setup}
We use $500$ finely annotated images of the Cityscapes validation set, considering only a subset of classes for the analysis. For our experiments we use the Deeplabv3+ \cite{Chen2018EncoderDecoderWA} network with a Mobilenetv2 backbone \cite{Sandler2018MobileNetV2IR}. Our optimization setup largely carries over from Sec.~\ref{subsec:toydata_experimental_setup}, with the exception that we optimize a coarse \(16\) by \(32\) pixel mask using SGD with a learning rate of \(1\) for \(80\) steps and use \(\lambda = 0.01\). Additional implementation details are provided in the supp. material.


\paragraph{Analysis of context explanations}
\label{subsec:cityscapes_class_statistics}

\begin{figure}[t!]
 	\begin{centering}
 		\setlength{\tabcolsep}{0.1em}
 		\renewcommand{\arraystretch}{0}
 		\par\end{centering}
 	\begin{centering}
 		\hfill{}%
 		\begin{tabular}{@{}c@{}c@{}c@{}c@{}}
 			
 				\includegraphics[width=0.24\textwidth, height=0.075\textheight]{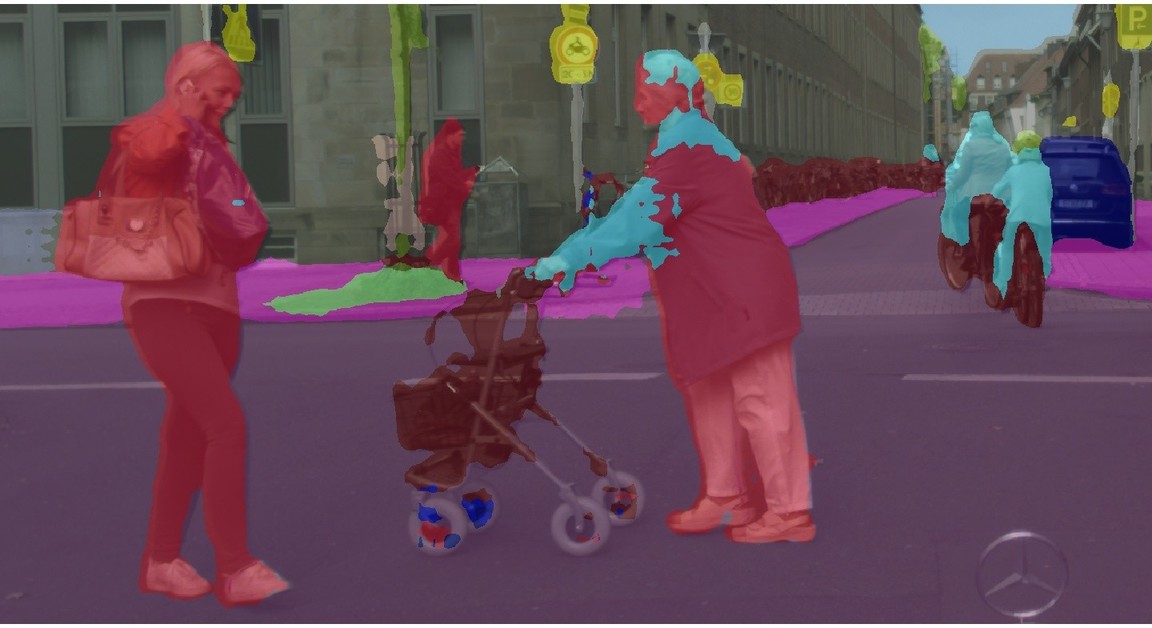} & {\footnotesize{}}
 			\includegraphics[width=0.24\textwidth, height=0.075\textheight]{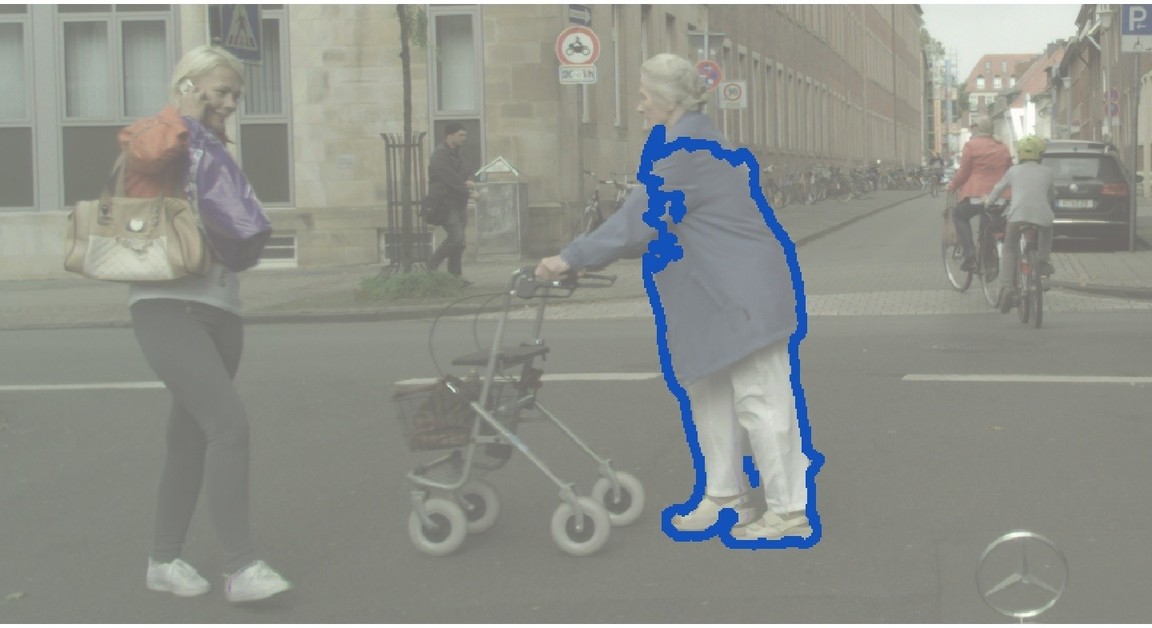} & {\footnotesize{}}
 			\includegraphics[width=0.24\textwidth, height=0.075\textheight]{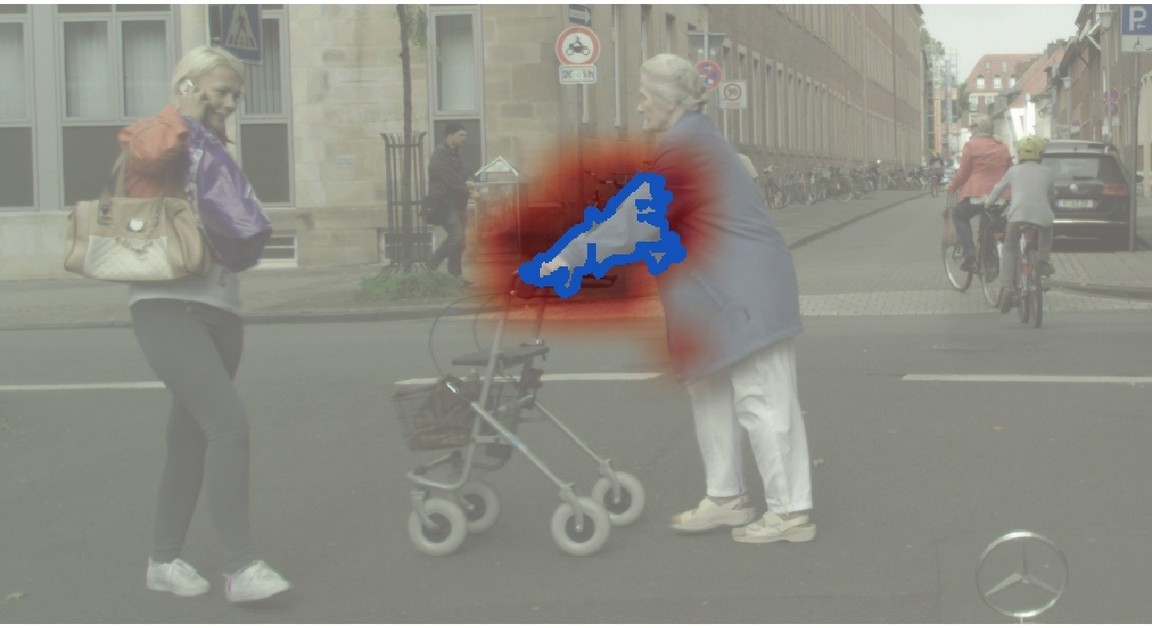}& {\footnotesize{}}
 			\includegraphics[width=0.24\textwidth, height=0.075\textheight]{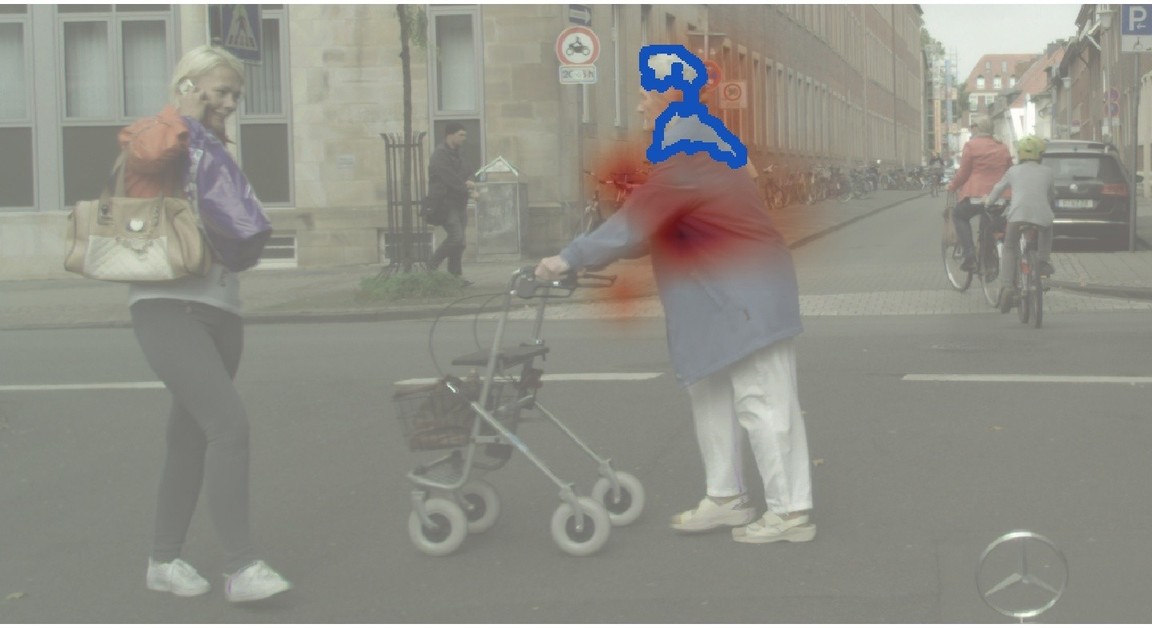}
 			
 			\tabularnewline
 			Semantic segment.~\cite{Chen2018DeepLabSI} & Person (corr) & Rider (err) & Rider (err)  \tabularnewline
 			 			
  			\includegraphics[width=0.24\textwidth, height=0.075\textheight]{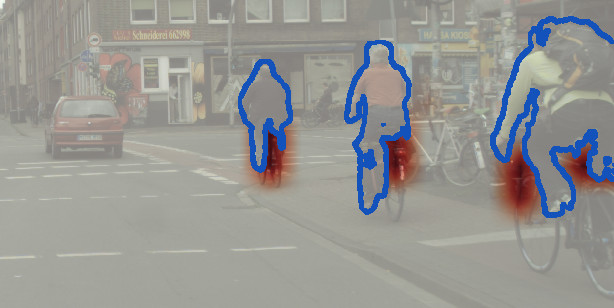} & {\footnotesize{}}
 			\includegraphics[width=0.24\textwidth, height=0.075\textheight]{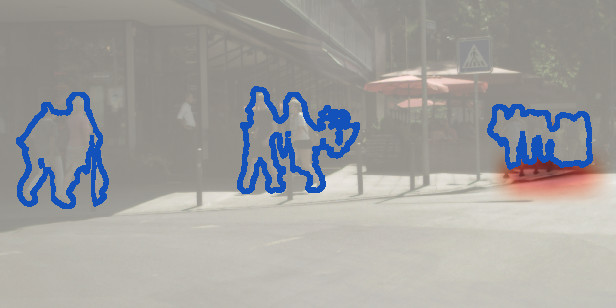}& {\footnotesize{}}
 			\includegraphics[width=0.24\textwidth, height=0.075\textheight]{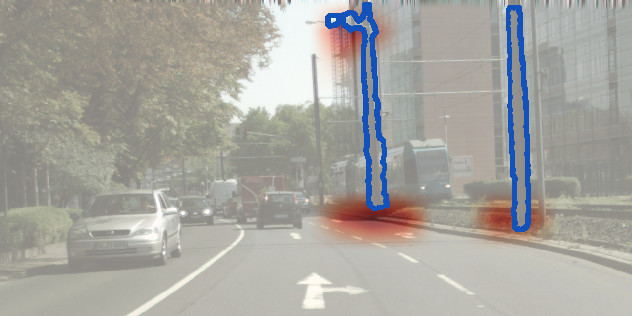}& {\footnotesize{}}
 			\includegraphics[width=0.24\textwidth, height=0.075\textheight]{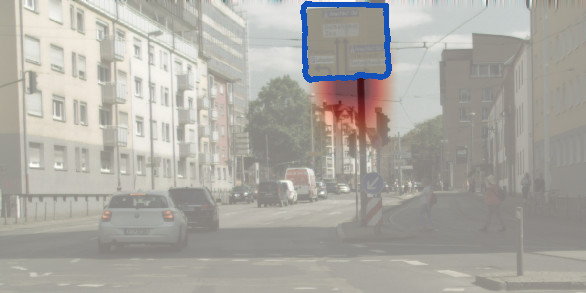} 
 			\tabularnewline
 			Rider (corr) &  Person (corr) &  Pole (corr) &  Traffic sign (corr)\tabularnewline

 		\end{tabular}\hfill{}
 		\par\end{centering}
 	
	\vspace{-0.5em}
	\caption{\label{fig:cityscapes_errorcases}
		Context explanations by grid saliency for erroneous (err) and correct (corr) semantic segmentation predictions on Cityscapes~\cite{Cordts2016Cityscapes}. Salient red regions visualize the most important context $M_{\text{context}}$ for the requested segment prediction $R$ outlined in blue. See Sec.~\ref{sec:experiments_cityscapes} for discussion.
	}
	\vspace{-0.5em}
\end{figure}

\begin{figure}[t!]
	\vspace{-0.5em}
	\centering
		\hfill{}%
		\begin{tabular}{@{}c@{}c@{}c@{}}

	\includegraphics[height=0.13\textheight]{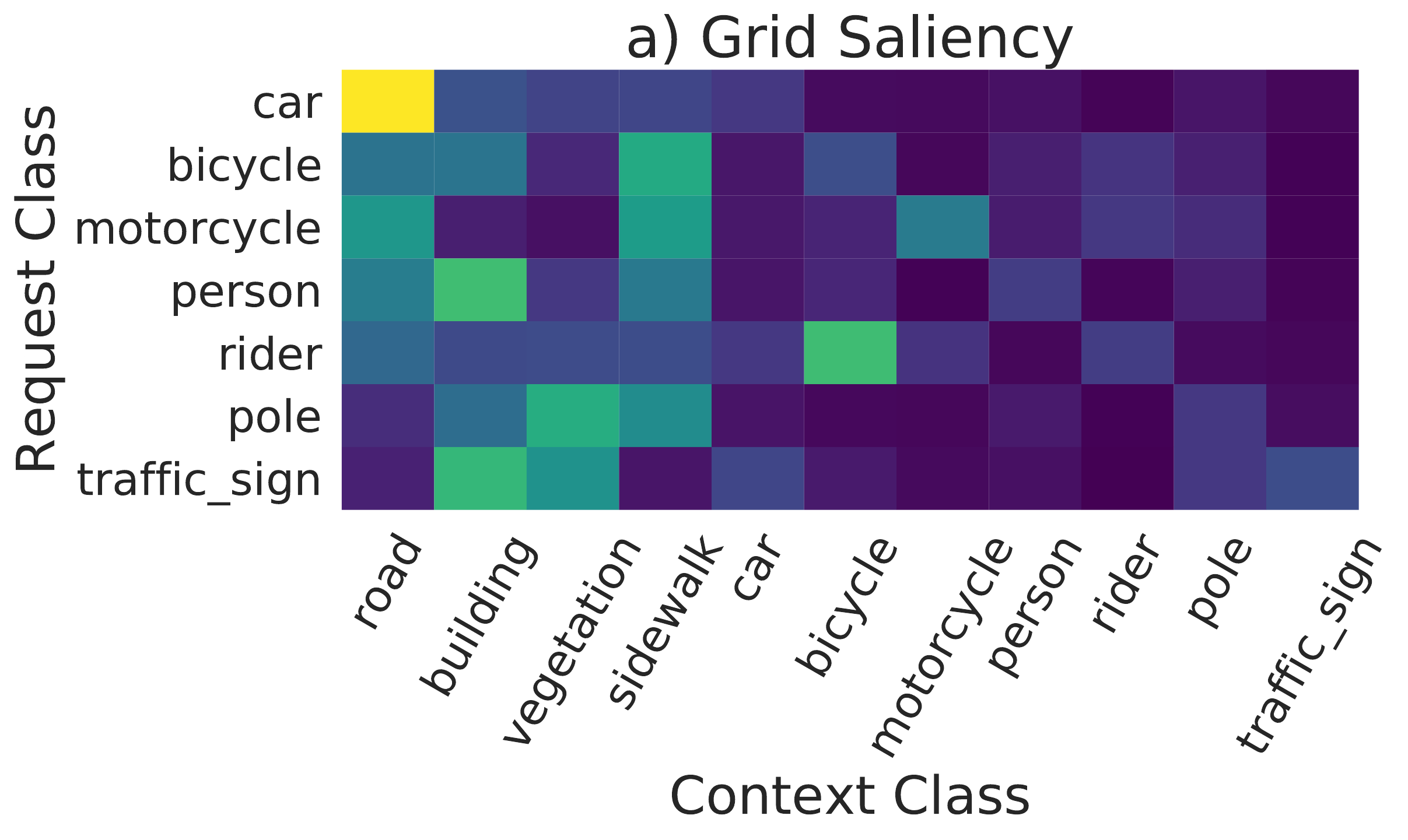} &
		\includegraphics[height=0.13\textheight]{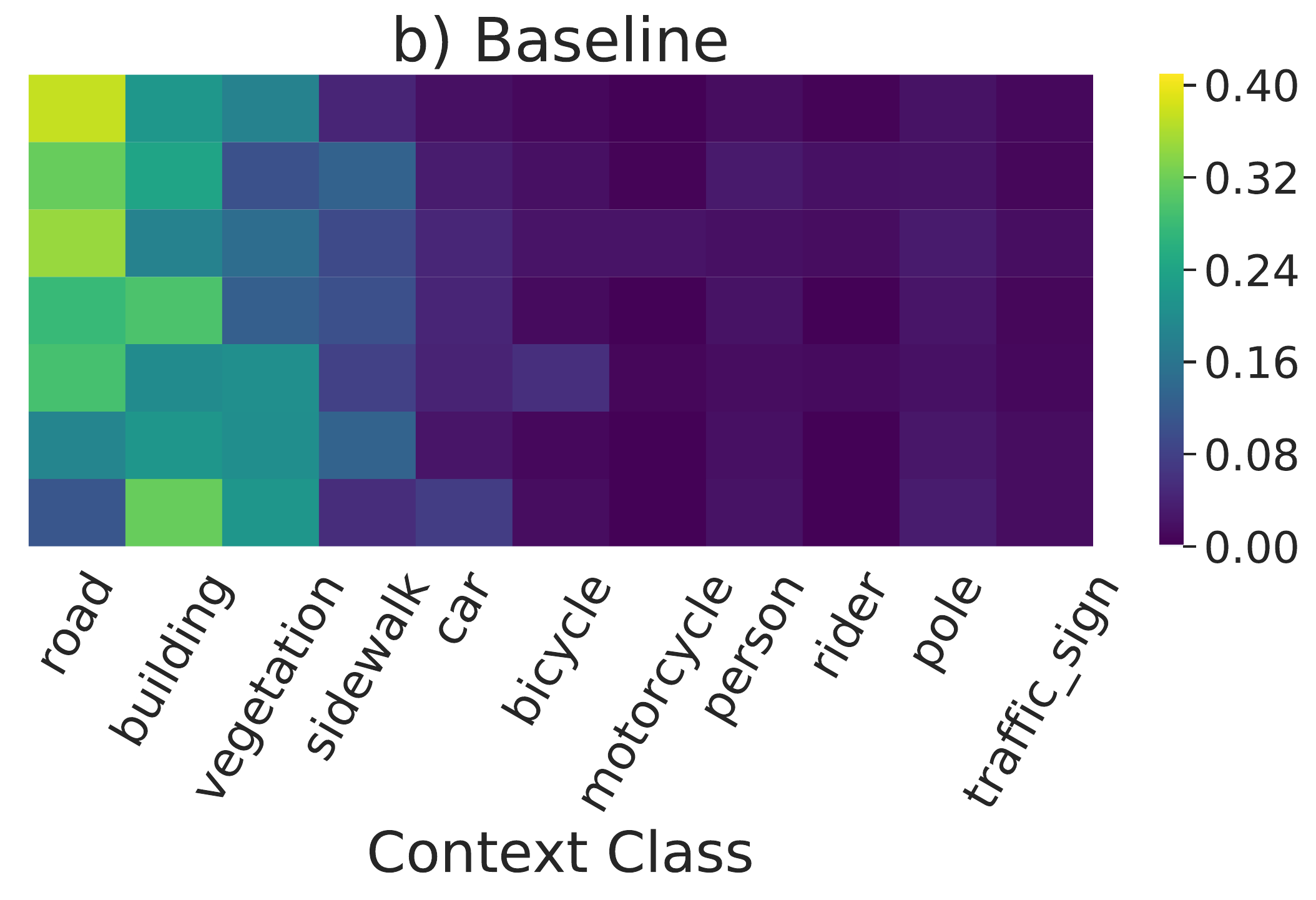} &
		\includegraphics[height=0.13\textheight]{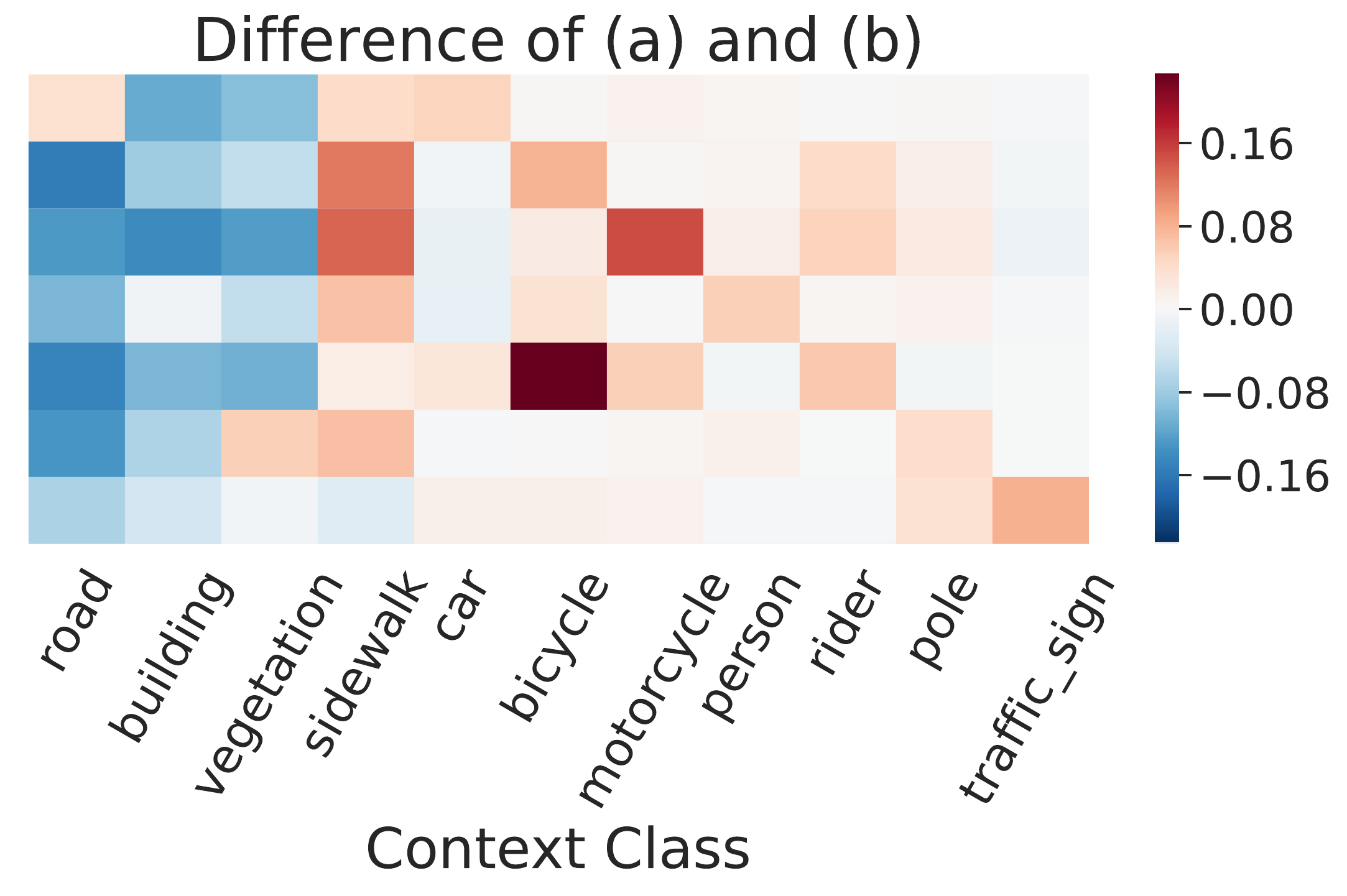}
		\tabularnewline

	\end{tabular}\hfill{}
	\vspace{-0.7em}
	\caption{\label{fig:cityscapes_statistics} Class statistics of context explanations on Cityscapes, see Sec.~\ref{sec:experiments_cityscapes} for discussion.
	}
	\vspace{-1.8em}
\end{figure}

In order to gain a global understanding of the model context bias, we aggregate statistics from the produced grid saliency maps for several requested classes across all validation images. These statistics are summarized in Fig.~\ref{fig:cityscapes_statistics}. For each image and class, the context saliency is computed for all sufficiently large instances (at least $10$k pixels), as can be seen in the second row of Fig. \ref{fig:cityscapes_errorcases}. Next, for each image and semantic class we compute the weighted label distribution of context salient pixels. The weight of each pixel is given by the computed saliency intensity at said pixel and its context class label is taken from the ground truth segmentation. For each requested object class, we group by label across all images and sum the saliency values within each group. This yields an accumulated saliency value for each label. We normalize the results by the sum of all saliency values across all classes and images to yield a probability distribution across labels.

To show that context explanations produced by our grid saliency are meaningful, we additionally compare
its accumulated context class statistics with their baseline class distributions. These baselines are computed in the same manner as above, but instead of relying on the optimized saliency map we use a fixed dilation of the contour of all predicted object segments considered in the image, thereby capturing the immediate context around each object in a uniform set of directions.

Fig. \ref{fig:cityscapes_statistics} compares these accumulated statistics of context explanations for the requested object classes, given on the $y$ axis. We observe that the context explanations are focused across reasonable and somewhat expected class subsets, which vary per class. For instance, in comparison to the baseline, context saliency for the rider class shifts attention from its spatially co-occurring classes such as road, vegetation and building, to mainly the bicycle class. Car context saliency attention is decreased on building and vegetation and mainly focused on road, sidewalk and other cars. Bicycle context saliency mostly attends to sidewalk rather than the road class. Overall, our grid saliency is able to provide sensible and coherent explanations for network decision making, which reflect semantic dependencies present in street scenes.
These quantitative findings are validated by the qualitative results in Fig. \ref{fig:cityscapes_errorcases}. The second row of Fig. \ref{fig:cityscapes_errorcases} gives several representative examples of context saliencies. Please refer to the supp. material for additional examples and a high-resolution version of Fig. \ref{fig:cityscapes_statistics} as well as a qualitative model comparison on MS COCO.

\paragraph{Context explanations of erroneous predictions} One of the motivations for this work is also to explain unexpected model behavior. Specifically, in case of erroneous predictions (i.e., Fig. \ref{fig:concept} and \ref{fig:context_cow}), we wish to understand to which extent context bias contributes to the failure.
Table \ref{tab:citysc} shows class statistics of context explanations for correct and erroneous predictions, where
the intersection of the ground truth mask and the correct or misclassified prediction region are used as the request mask $R$, and similarly to Fig. \ref{fig:cityscapes_statistics} semantic class statistics are computed inside the salient context for all sufficiently large request masks (at least $625$ pixels).
We observe that the saliency severely changes for the error cases in comparison to the correct predictions. E.g. for the correctly classified rider class, context saliency mostly focuses on bicycle ($22\%$), but is much less present ($10\%$) when rider is mistaken as person and in this case the context saliency shifts from bicycle to building ($24\%$). The opposite effect is observed when person is misclassified as rider. Sidewalk saliency is decreased when bicycle is misclassified as motorcycle ($19\%$ vs. $13\%$) and when pole is misclassified as building ($10\%$ vs. $2\%$). In general, grid saliency is able to provide reasonable explanations of erroneous predictions.

\begin{table}[t]
	\vspace{-0.5em}
	\setlength{\tabcolsep}{0.2em}
	\renewcommand{\arraystretch}{0.95}
	\centering
	\caption{Class statistics of context explanations of correct and erroneous predictions on Cityscapes.}
	\label{tab:citysc}
	\begin{tabular}{cc|cccccccccc}
		\multirow{2}{*}{	\footnotesize GT }	& \multirow{2}{*}{\footnotesize Prediction} & \multicolumn{10}{c}{ \footnotesize Context class} \tabularnewline
	&   & \footnotesize road &  \footnotesize bicycle
		&  \footnotesize motorcycle & \footnotesize car &  \footnotesize vegetation & \footnotesize building &  \footnotesize sidewalk &  \footnotesize rider &  \footnotesize person &  \footnotesize pole  \tabularnewline
		\hline 	\hline

				\multirow{2}{*}{\footnotesize rider} & \footnotesize \textcolor{darkgreen}{rider}   & \footnotesize $0.17$ & \footnotesize $\mathbf{0.22}$  & \footnotesize $0.03$ & \footnotesize $0.07$ &\footnotesize $0.09$ & \footnotesize $0.11$ & \footnotesize $0.10$ & \footnotesize $0.07$ & \footnotesize $0.02$ & \footnotesize $0.02$  \tabularnewline

	& \footnotesize \textcolor{red}{person}   & \footnotesize $0.13$ & \footnotesize $0.10$ & \footnotesize $0.01$ & \footnotesize $0.05$ &\footnotesize $0.10$ & \footnotesize $\mathbf{0.24}$ & \footnotesize $0.05$ & \footnotesize $0.08$ & \footnotesize $0.11$ & \footnotesize $0.04$  \tabularnewline

			\hline

				\multirow{3}{*}{\footnotesize person} & \footnotesize \textcolor{darkgreen}{person}  & \footnotesize $0.12$ & \footnotesize $0.03$ & \footnotesize $0.00$ & \footnotesize $0.09$ &\footnotesize $0.15$ & \footnotesize $\mathbf{0.22}$ & \footnotesize $0.13$ & \footnotesize $0.01$ & \footnotesize $0.05$ & \footnotesize $0.05$  \tabularnewline

	 & \footnotesize \textcolor{red}{rider}   & \footnotesize $0.09$ & \footnotesize $\mathbf{0.18}$ & \footnotesize $0.02$ & \footnotesize $0.04$ &\footnotesize $0.10$ & \footnotesize $0.13$ & \footnotesize $0.07$ & \footnotesize $0.08$ & \footnotesize $0.15$
			& \footnotesize $0.03$  \tabularnewline

	 & \footnotesize \textcolor{red}{car}   & \footnotesize $0.19$ & \footnotesize $0.01$ & \footnotesize $0.00$ & \footnotesize $\mathbf{0.32}$ &\footnotesize $0.05$ & \footnotesize $0.11$ & \footnotesize $0.03$ & \footnotesize $0.01$ & \footnotesize $0.19$
	& \footnotesize $0.01$  \tabularnewline

			\hline

				\multirow{2}{*}{\footnotesize bicycle} & \footnotesize \textcolor{darkgreen}{bicycle}  & \footnotesize $0.18$ & \footnotesize $0.06$ & \footnotesize $0.01$ & \footnotesize $0.05$ &\footnotesize $0.09$ & \footnotesize $0.15$ & \footnotesize $\mathbf{0.19}$ & \footnotesize $0.05$ & \footnotesize $0.04$ & \footnotesize $0.04$  \tabularnewline

	 & \footnotesize \textcolor{red}{motorcycle}   & \footnotesize $0.19$ & \footnotesize $\mathbf{0.27}$ & \footnotesize $0.13$ &  \footnotesize $0.01$ & \footnotesize $0.02$ & \footnotesize $0.11$ & \footnotesize $0.13$ & \footnotesize $0.03$ & \footnotesize $0.01$ & \footnotesize $0.04$  \tabularnewline

				\hline

      	\multirow{3}{*}{\footnotesize car} & \footnotesize \textcolor{darkgreen}{car}  & \footnotesize $\mathbf{0.39}$ & \footnotesize $0.02$ & \footnotesize $0.00$ & \footnotesize $0.06$ &\footnotesize $0.11$ & \footnotesize $0.10$ & \footnotesize $0.08$ & \footnotesize $0.01$ & \footnotesize $0.04$ & \footnotesize $0.04$ \tabularnewline

  & \footnotesize \textcolor{red}{motorcycle}   & \footnotesize $0.07$ & \footnotesize $0.06$ & \footnotesize $\mathbf{0.22}$ &  \footnotesize $0.17$ & \footnotesize $0.14$ & \footnotesize $0.07$ & \footnotesize $0.02$ & \footnotesize $0.01$ & \footnotesize $0.00$ & \footnotesize $0.04$  \tabularnewline

 & \footnotesize \textcolor{red}{person}   & \footnotesize $0.14$ & \footnotesize $0.01$ & \footnotesize $0.00$ &  \footnotesize $\mathbf{0.28}$ & \footnotesize $0.03$ & \footnotesize $0.14$ & \footnotesize $0.07$ & \footnotesize $0.01$ & \footnotesize $0.22$ & \footnotesize $0.02$  \tabularnewline

		\hline

	  \multirow{3}{*}{ 	\footnotesize motorcycle} & \footnotesize \textcolor{darkgreen}{motorcycle}  & \footnotesize $\mathbf{0.18}$ & \footnotesize $0.04$ & \footnotesize $0.10$ & \footnotesize $0.04$ &\footnotesize $0.07$ & \footnotesize $0.14$ & \footnotesize $\mathbf{0.18}$ & \footnotesize $0.05$ & \footnotesize $0.03$ & \footnotesize $0.06$ \tabularnewline

  & \footnotesize \textcolor{red}{bicycle}   & \footnotesize $0.13$ & \footnotesize $0.10$ & \footnotesize $0.07$ &  \footnotesize $0.10$ & \footnotesize $0.10$ & \footnotesize $\mathbf{0.14}$ & \footnotesize $0.11$ & \footnotesize $0.01$ & \footnotesize $0.05$ & \footnotesize $0.03$ \tabularnewline

	  & \footnotesize \textcolor{red}{car}   & \footnotesize $0.16$ & \footnotesize $0.01$ & \footnotesize $\mathbf{0.18}$ &  \footnotesize $0.10$ & \footnotesize $0.10$ & \footnotesize $0.10$ & \footnotesize $0.11$ & \footnotesize $0.05$ & \footnotesize $0.04$ & \footnotesize $0.06$ \tabularnewline

			\hline

\multirow{3}{*}{	\footnotesize pole }& \footnotesize \textcolor{darkgreen}{pole}  & \footnotesize $0.04$ & \footnotesize $0.02$ & \footnotesize $0.00$ & \footnotesize $0.05$ &\footnotesize $\mathbf{0.28}$ & \footnotesize $0.23$ & \footnotesize $0.10$ & \footnotesize $0.00$ & \footnotesize $0.03$ & \footnotesize $0.04$ \tabularnewline

 & \footnotesize \textcolor{red}{building}   & \footnotesize $0.04$ & \footnotesize $0.01$ & \footnotesize $0.00$ &  \footnotesize $0.04$ & \footnotesize $0.07$ & \footnotesize $\mathbf{0.57}$ & \footnotesize $0.02$ & \footnotesize $0.00$ & \footnotesize $0.01$ & \footnotesize $0.05$ \tabularnewline

 & \footnotesize \textcolor{red}{vegetation}   & \footnotesize $0.03$ & \footnotesize $0.00$ & \footnotesize $0.00$ &  \footnotesize $0.03$ & \footnotesize $\mathbf{0.63}$ & \footnotesize $0.08$ & \footnotesize $0.03$ & \footnotesize $0.00$ & \footnotesize $0.00$ & \footnotesize $0.07$ \tabularnewline

	\end{tabular}
	\vspace{-2.0em}
\end{table}

Fig. \ref{fig:concept} (second row) and Fig. \ref{fig:cityscapes_errorcases} (first row) showcase how applying grid saliency on the same input image but different prediction outputs is useful for isolating failures caused by context bias. Both of them show examples of context explanations for erroneous segmentations of a single object. In Fig. \ref{fig:cityscapes_errorcases}, the arms, head and shoulders of a pedestrian are classified as rider, while the rest of the body is correctly identified as person. The context saliency for the former body parts activates highly on the arms and stroller handles, whereas activations for person does not highlight this support. Thus, a reasonable conclusion is that the misclassification may be attributed to the arm pose and potentially also to the context bias given by the similar appearance of the stroller and bicycle handles.

\vspace{-0.6em}

\section{Conclusions}
\label{sec:conclusions}

\vspace{-0.6em}

We proposed spatial grid saliency, a general framework to produce spatially coherent explanations for (pixel-level) dense prediction networks, which to the best of our knowledge is the first method to extend saliency techniques beyond classification models. We investigated the ability of grid saliency to provide context explanations for semantic segmentation, showing its effectiveness to detect and localize context bias on the synthetic toy dataset specifically designed with an artificially induced bias to benchmark this task. Our results on the real-world data indicated that grid saliency can be successfully employed to produce easily interpretable and faithful context explanations for semantic segmentation, helping to discover spatial and semantic correlations in the data picked up by the network. We hope the proposed grid saliency and the insights of this work can contribute to a better understanding of semantic segmentation networks or other models for dense prediction, elucidating some aspects of the problem that have not been well explored so far.

Besides enabling visual explanations for dense-prediction tasks, we see potential utility of grid saliency for the following applications: 
\textit{1) Architecture comparison:} Context explanations produced by grid saliency
can be used to compare architectures with respect to their capacity
to either learn or to be invariant towards context.
\textit{2) Network generalization via active learning:} 
Existing context biases might impair network generalization.
E.g., cows might mostly appear on grass during training.
A network that was trained and evaluated on this data
and picked up that bias will perform poorly in real-world cases,
where the cow, for example, appears on road and gets misclassified as the horse class, see Fig. \ref{fig:context_cow}.
Actions can be taken, such as targeted extra data collection, to improve network generalization.
\textit{3) Adversarial detection:} Grid saliency can be used to detect and localize adversarial patches outside object boundaries (e.g. \cite{lee2019physical}). Cases for which the salient regions lie largely outside an object would strongly indicate the presence of an adversary or misguided prediction.
We consider the above-mentioned utilities of grid saliency interesting and promising future research directions.

{
\small
\bibliographystyle{unsrtnat} 
\bibliography{tex/bibtex_references}

\begin{thebibliography}{56}
\providecommand{\natexlab}[1]{#1}
\providecommand{\url}[1]{\texttt{#1}}
\expandafter\ifx\csname urlstyle\endcsname\relax
  \providecommand{\doi}[1]{doi: #1}\else
  \providecommand{\doi}{doi: \begingroup \urlstyle{rm}\Url}\fi

\bibitem[Sandler et~al.(2018)Sandler, Howard, Zhu, Zhmoginov, and
  Chen]{Sandler2018MobileNetV2IR}
Mark~B. Sandler, Andrew~G. Howard, Menglong Zhu, Andrey Zhmoginov, and
  Liang-Chieh Chen.
\newblock Mobilenetv2: Inverted residuals and linear bottlenecks.
\newblock In \emph{Proceedings of the IEEE Conference on Computer Vision and
  Pattern Recognition (CVPR)}, pages 4510--4520, 2018.

\bibitem[Chen et~al.(2018{\natexlab{a}})Chen, Zhu, Papandreou, Schroff, and
  Adam]{Chen2018EncoderDecoderWA}
Liang-Chieh Chen, Yukun Zhu, George Papandreou, Florian Schroff, and Hartwig
  Adam.
\newblock Encoder-decoder with atrous separable convolution for semantic image
  segmentation.
\newblock In \emph{Proceedings of the European Conference on Computer Vision
  (ECCV)}, 2018{\natexlab{a}}.

\bibitem[Cordts et~al.(2016)Cordts, Omran, Ramos, Rehfeld, Enzweiler, Benenson,
  Franke, Roth, and Schiele]{Cordts2016Cityscapes}
Marius Cordts, Mohamed Omran, Sebastian Ramos, Timo Rehfeld, Markus Enzweiler,
  Rodrigo Benenson, Uwe Franke, Stefan Roth, and Bernt Schiele.
\newblock The cityscapes dataset for semantic urban scene understanding.
\newblock In \emph{Proceedings of the IEEE Conference on Computer Vision and
  Pattern Recognition (CVPR)}, 2016.

\bibitem[Simonyan et~al.(2013)Simonyan, Vedaldi, and
  Zisserman]{Simonyan2013DeepIC}
Karen Simonyan, Andrea Vedaldi, and Andrew Zisserman.
\newblock Deep inside convolutional networks: Visualising image classification
  models and saliency maps.
\newblock In \emph{International Conference on Learning Representations
  (ICLR)}, 2013.

\bibitem[Selvaraju et~al.(2016)Selvaraju, Das, Vedantam, Cogswell, Parikh, and
  Batra]{Selvaraju2016GradCAMWD}
Ramprasaath~R. Selvaraju, Abhishek Das, Ramakrishna Vedantam, Michael Cogswell,
  Devi Parikh, and Dhruv Batra.
\newblock Grad-cam: Why did you say that?
\newblock \emph{\tt arXiv:1611.07450}, 2016.

\bibitem[Fong and Vedaldi(2017)]{Fong2017InterpretableEO}
Ruth Fong and Andrea Vedaldi.
\newblock Interpretable explanations of black boxes by meaningful perturbation.
\newblock In \emph{Proceedings of the IEEE International Conference on Computer
  Vision (ICCV)}, pages 3449--3457, 2017.

\bibitem[Adebayo et~al.(2018)Adebayo, Gilmer, Muelly, Goodfellow, Hardt, and
  Kim]{Adebayo2018SanityCF}
Julius Adebayo, Justin Gilmer, Michael Muelly, Ian~J. Goodfellow, Moritz Hardt,
  and Been Kim.
\newblock Sanity checks for saliency maps.
\newblock In \emph{Advances in Neural Information Processing Systems
  (NeurIPS)}, 2018.

\bibitem[Chang et~al.(2019)Chang, Creager, Goldenberg, and
  Duvenaud]{Chang2018ExplainingIC}
Chun-Hao Chang, Elliot Creager, Anna Goldenberg, and David Duvenaud.
\newblock Explaining image classifiers by counterfactual generation.
\newblock In \emph{International Conference on Learning Representations
  (ICLR)}, 2019.

\bibitem[Zintgraf et~al.(2017)Zintgraf, Cohen, Adel, and
  Welling]{Zintgraf2017VisualizingDN}
Luisa~M. Zintgraf, Taco Cohen, Tameem Adel, and Max Welling.
\newblock Visualizing deep neural network decisions: Prediction difference
  analysis.
\newblock In \emph{International Conference on Learning Representations
  (ICLR)}, 2017.

\bibitem[Mottaghi et~al.(2014)Mottaghi, Chen, Liu, Cho, Lee, Fidler, Urtasun,
  and Yuille]{Mottaghi2014TheRO}
Roozbeh Mottaghi, Xianjie Chen, Xiaobai Liu, Nam-Gyu Cho, Seong-Whan Lee, Sanja
  Fidler, Raquel Urtasun, and Alan~Loddon Yuille.
\newblock The role of context for object detection and semantic segmentation in
  the wild.
\newblock 2014.

\bibitem[Uijlings et~al.(2012)Uijlings, Smeulders, and
  Scha]{uijlings2012visual}
Jasper~RR Uijlings, Arnold~WM Smeulders, and Remko~JH Scha.
\newblock The visual extent of an object.
\newblock \emph{International journal of computer vision}, 96\penalty0
  (1):\penalty0 46--63, 2012.

\bibitem[Azaza et~al.(2018)Azaza, van~de Weijer, Douik, and
  Masana]{azaza2018context}
Aymen Azaza, Joost van~de Weijer, Ali Douik, and Marc Masana.
\newblock Context proposals for saliency detection.
\newblock \emph{Computer Vision and Image Understanding}, 174:\penalty0 1--11,
  2018.

\bibitem[LeCun et~al.(1998)LeCun, Cortes, and Burges]{MNIST}
Yann LeCun, Corinna Cortes, and Christopher~J.C. Burges.
\newblock The {MNIST} database of handwritten digits.
\newblock Technical report, 1998.

\bibitem[Sundararajan et~al.(2017)Sundararajan, Taly, and
  Yan]{Sundararajan2017AxiomaticAF}
Mukund Sundararajan, Ankur Taly, and Qiqi Yan.
\newblock Axiomatic attribution for deep networks.
\newblock In \emph{International Conference on Machine Learning (ICML)}, 2017.

\bibitem[Smilkov et~al.(2017)Smilkov, Thorat, Kim, Vi{\'e}gas, and
  Wattenberg]{Smilkov2017SmoothGradRN}
Daniel Smilkov, Nikhil Thorat, Been Kim, Fernanda~B. Vi{\'e}gas, and Martin
  Wattenberg.
\newblock Smoothgrad: removing noise by adding noise.
\newblock \emph{\tt arXiv:1706.03825}, 2017.

\bibitem[Karpathy and Fei-Fei(2015)]{Karpathy2015DeepVA}
Andrej Karpathy and Li~Fei-Fei.
\newblock Deep visual-semantic alignments for generating image descriptions.
\newblock In \emph{Proceedings of the IEEE Conference on Computer Vision and
  Pattern Recognition (CVPR)}, 2015.

\bibitem[Jos{\'e}Oramas and Tuytelaars(2016)]{JosOramas2016ModelingVC}
M.~Oramas Jos{\'e}Oramas and Tinne Tuytelaars.
\newblock Modeling visual compatibility through hierarchical mid-level
  elements.
\newblock \emph{\tt arXiv:1604.00036}, 2016.

\bibitem[Gulshad et~al.(2019)Gulshad, Metzen, Smeulders, and
  Akata]{Gulshad2019InterpretingAE}
Sadaf Gulshad, Jan~Hendrik Metzen, Arnold W.~M. Smeulders, and Zeynep Akata.
\newblock Interpreting adversarial examples with attributes.
\newblock \emph{\tt arXiv:1904.08279}, 2019.

\bibitem[Park et~al.(2018)Park, Hendricks, Akata, Rohrbach, Schiele, Darrell,
  and Rohrbach]{Park2018MultimodalEJ}
Dong~Huk Park, Lisa~Anne Hendricks, Zeynep Akata, Anna Rohrbach, Bernt Schiele,
  Trevor Darrell, and Marcus Rohrbach.
\newblock Multimodal explanations: Justifying decisions and pointing to the
  evidence.
\newblock In \emph{Proceedings of the IEEE Conference on Computer Vision and
  Pattern Recognition (CVPR)}, 2018.

\bibitem[Hendricks et~al.(2016)Hendricks, Akata, Rohrbach, Donahue, Schiele,
  and Darrell]{Hendricks2016GeneratingVE}
Lisa~Anne Hendricks, Zeynep Akata, Marcus Rohrbach, Jeff Donahue, Bernt
  Schiele, and Trevor Darrell.
\newblock Generating visual explanations.
\newblock In \emph{Proceedings of the European Conference on Computer Vision
  (ECCV)}, 2016.

\bibitem[Dong et~al.(2017)Dong, Su, Zhu, and Zhang]{Dong2017ImprovingIO}
Yinpeng Dong, Hang Su, Jun Zhu, and Bo~Zhang.
\newblock Improving interpretability of deep neural networks with semantic
  information.
\newblock In \emph{Proceedings of the IEEE Conference on Computer Vision and
  Pattern Recognition (CVPR)}, 2017.

\bibitem[Zhou et~al.(2016)Zhou, Khosla, Lapedriza, Oliva, and
  Torralba]{Zhou2016LearningDF}
Bolei Zhou, Aditya Khosla, {\`A}gata Lapedriza, Aude Oliva, and Antonio
  Torralba.
\newblock Learning deep features for discriminative localization.
\newblock In \emph{Proceedings of the IEEE Conference on Computer Vision and
  Pattern Recognition (CVPR)}, 2016.

\bibitem[Lin et~al.(2014)Lin, Maire, Belongie, Bourdev, Girshick, Hays, Perona,
  Ramanan, Doll{\'a}r, and Zitnick]{Lin2014MicrosoftCC}
Tsung-Yi Lin, Michael Maire, Serge~J. Belongie, Lubomir~D. Bourdev, Ross~B.
  Girshick, James Hays, Pietro Perona, Deva Ramanan, Piotr Doll{\'a}r, and
  C.~Lawrence Zitnick.
\newblock Microsoft coco: Common objects in context.
\newblock In \emph{Proceedings of the European Conference on Computer Vision
  (ECCV)}, 2014.

\bibitem[Zeiler and Fergus(2014)]{Zeiler2014VisualizingAU}
Matthew~D. Zeiler and Rob Fergus.
\newblock Visualizing and understanding convolutional networks.
\newblock In \emph{Proceedings of the European Conference on Computer Vision
  (ECCV)}, 2014.

\bibitem[Springenberg et~al.(2014)Springenberg, Dosovitskiy, Brox, and
  Riedmiller]{Springenberg2014StrivingFS}
Jost~Tobias Springenberg, Alexey Dosovitskiy, Thomas Brox, and Martin~A.
  Riedmiller.
\newblock Striving for simplicity: The all convolutional net.
\newblock In \emph{International Conference on Learning Representations
  (ICLR)}, 2014.

\bibitem[Bach et~al.(2015)Bach, Binder, Montavon, Klauschen, M{\"u}ller, Samek,
  and Suarez]{Bach2015OnPE}
Sebastian Bach, Alexander Binder, Gr{\'e}goire Montavon, Frederick Klauschen,
  Klaus-Robert M{\"u}ller, Wojciech Samek, and Oscar~Deniz Suarez.
\newblock On pixel-wise explanations for non-linear classifier decisions by
  layer-wise relevance propagation.
\newblock In \emph{PloS one}, 2015.

\bibitem[Zhang et~al.(2017)Zhang, Bargal, Lin, Brandt, Shen, and
  Sclaroff]{Zhang2017TopDownNA}
Jianming Zhang, Sarah~Adel Bargal, Zhe~L. Lin, Jonathan Brandt, Xiaohui Shen,
  and Stan Sclaroff.
\newblock Top-down neural attention by excitation backprop.
\newblock \emph{International Journal of Computer Vision (IJCV)}, 2017.

\bibitem[Dabkowski and Gal(2017)]{Dabkowski2017RealTI}
Piotr Dabkowski and Yarin Gal.
\newblock Real time image saliency for black box classifiers.
\newblock In \emph{Advances in Neural Information Processing Systems
  (NeurIPS)}, 2017.

\bibitem[Al-Shedivat et~al.(2018)Al-Shedivat, Dubey, and
  Xing]{AlShedivat2018ContextualEN}
Maruan Al-Shedivat, Kumar~Avinava Dubey, and Eric~P. Xing.
\newblock Contextual explanation networks.
\newblock \emph{\tt arXiv:1705.10301}, 2018.

\bibitem[Zhao et~al.(2017)Zhao, Shi, Qi, Wang, and Jia]{Zhao2017PyramidSP}
Hengshuang Zhao, Jianping Shi, Xiaojuan Qi, Xiaogang Wang, and Jiaya Jia.
\newblock Pyramid scene parsing network.
\newblock In \emph{Proceedings of the IEEE Conference on Computer Vision and
  Pattern Recognition (CVPR)}, pages 6230--6239, 2017.

\bibitem[Pohlen et~al.(2017)Pohlen, Hermans, Mathias, and
  Leibe]{Pohlen2017FullResolutionRN}
Tobias Pohlen, Alexander Hermans, Markus Mathias, and Bastian Leibe.
\newblock Full-resolution residual networks for semantic segmentation in street
  scenes.
\newblock In \emph{Proceedings of the IEEE Conference on Computer Vision and
  Pattern Recognition (CVPR)}, pages 3309--3318, 2017.

\bibitem[Lin et~al.(2017)Lin, Milan, Shen, and Reid]{Lin2017RefineNetMR}
Guosheng Lin, Anton Milan, Chunhua Shen, and Ian~D. Reid.
\newblock Refinenet: Multi-path refinement networks for high-resolution
  semantic segmentation.
\newblock In \emph{Proceedings of the IEEE Conference on Computer Vision and
  Pattern Recognition (CVPR)}, pages 5168--5177, 2017.

\bibitem[Shelhamer et~al.(2015)Shelhamer, Long, and
  Darrell]{Shelhamer2015FullyCN}
Evan Shelhamer, Jonathan Long, and Trevor Darrell.
\newblock Fully convolutional networks for semantic segmentation.
\newblock In \emph{Proceedings of the IEEE Conference on Computer Vision and
  Pattern Recognition (CVPR)}, 2015.

\bibitem[Chen et~al.(2015)Chen, Papandreou, Kokkinos, Murphy, and
  Yuille]{Chen2015SemanticIS}
Liang-Chieh Chen, George Papandreou, Iasonas Kokkinos, Kevin Murphy, and
  Alan~Loddon Yuille.
\newblock Semantic image segmentation with deep convolutional nets and fully
  connected crfs.
\newblock In \emph{International Conference on Learning Representations
  (ICLR)}, 2015.

\bibitem[Yu and Koltun(2016)]{Yu2016MultiScaleCA}
Fisher Yu and Vladlen Koltun.
\newblock Multi-scale context aggregation by dilated convolutions.
\newblock In \emph{International Conference on Learning Representations
  (ICLR)}, 2016.

\bibitem[Lin et~al.(2016)Lin, Shen, Reid, and van~den
  Hengel]{Lin2016EfficientPT}
Guosheng Lin, Chunhua Shen, Ian~D. Reid, and Anton van~den Hengel.
\newblock Efficient piecewise training of deep structured models for semantic
  segmentation.
\newblock In \emph{Proceedings of the IEEE Conference on Computer Vision and
  Pattern Recognition (CVPR)}, 2016.

\bibitem[Zheng et~al.(2015)Zheng, Jayasumana, Romera-Paredes, Vineet, Su, Du,
  Huang, and Torr]{Zheng2015ConditionalRF}
Shuai Zheng, Sadeep Jayasumana, Bernardino Romera-Paredes, Vibhav Vineet,
  Zhizhong Su, Dalong Du, Chang Huang, and Philip H.~S. Torr.
\newblock Conditional random fields as recurrent neural networks.
\newblock In \emph{Proceedings of the IEEE International Conference on Computer
  Vision (ICCV)}, 2015.

\bibitem[Chandra and Kokkinos(2016)]{Chandra2016FastEA}
Siddhartha Chandra and Iasonas Kokkinos.
\newblock Fast, exact and multi-scale inference for semantic image segmentation
  with deep gaussian crfs.
\newblock In \emph{Proceedings of the European Conference on Computer Vision
  (ECCV)}, 2016.

\bibitem[Jampani et~al.(2016)Jampani, Kiefel, and
  Gehler]{Jampani2016LearningSH}
Varun Jampani, Martin Kiefel, and Peter~V. Gehler.
\newblock Learning sparse high dimensional filters: Image filtering, dense crfs
  and bilateral neural networks.
\newblock In \emph{Proceedings of the IEEE Conference on Computer Vision and
  Pattern Recognition (CVPR)}, 2016.

\bibitem[Chen et~al.(2018{\natexlab{b}})Chen, Papandreou, Kokkinos, Murphy, and
  Yuille]{Chen2018DeepLabSI}
Liang-Chieh Chen, George Papandreou, Iasonas Kokkinos, Kevin Murphy, and
  Alan~Loddon Yuille.
\newblock Deeplab: Semantic image segmentation with deep convolutional nets,
  atrous convolution, and fully connected crfs.
\newblock \emph{IEEE Transactions on Pattern Analysis and Machine Intelligence
  (PAMI)}, 40:\penalty0 834--848, 2018{\natexlab{b}}.

\bibitem[Ghiasi and Fowlkes(2016)]{Ghiasi2016LaplacianPR}
Golnaz Ghiasi and Charless~C. Fowlkes.
\newblock Laplacian pyramid reconstruction and refinement for semantic
  segmentation.
\newblock In \emph{Proceedings of the European Conference on Computer Vision
  (ECCV)}, 2016.

\bibitem[Hong et~al.(2015)Hong, Noh, and Han]{Hong2015DecoupledDN}
Seunghoon Hong, Hyeonwoo Noh, and Bohyung Han.
\newblock Decoupled deep neural network for semi-supervised semantic
  segmentation.
\newblock In \emph{Advances in Neural Information Processing Systems
  (NeurIPS)}, 2015.

\bibitem[Noh et~al.(2015)Noh, Hong, and Han]{Noh2015LearningDN}
Hyeonwoo Noh, Seunghoon Hong, and Bohyung Han.
\newblock Learning deconvolution network for semantic segmentation.
\newblock In \emph{Proceedings of the IEEE International Conference on Computer
  Vision (ICCV)}, 2015.

\bibitem[Ronneberger et~al.(2015)Ronneberger, Fischer, and
  Brox]{Ronneberger2015UNetCN}
Olaf Ronneberger, Philipp Fischer, and Thomas Brox.
\newblock U-net: Convolutional networks for biomedical image segmentation.
\newblock In \emph{MICCAI}, 2015.

\bibitem[Badrinarayanan et~al.(2016)Badrinarayanan, Kendall, and
  Cipolla]{Badrinarayanan2016SegNetAD}
Vijay Badrinarayanan, Alex Kendall, and Roberto Cipolla.
\newblock Segnet: A deep convolutional encoder-decoder architecture for image
  segmentation.
\newblock \emph{IEEE Transactions on Pattern Analysis and Machine Intelligence
  (PAMI)}, 39:\penalty0 2481--2495, 2016.

\bibitem[Poudel et~al.(2019)Poudel, Liwicki, and Cipolla]{Poudel2019FastSCNNFS}
Rudra P.~K. Poudel, Stephan Liwicki, and Roberto Cipolla.
\newblock Fast-scnn: Fast semantic segmentation network.
\newblock \emph{\tt arXiv:1902.04502}, 2019.

\bibitem[Paszke et~al.(2017)Paszke, Chaurasia, Kim, and
  Culurciello]{Paszke2017ENetAD}
Adam Paszke, Abhishek Chaurasia, Sangpil Kim, and Eugenio Culurciello.
\newblock Enet: A deep neural network architecture for real-time semantic
  segmentation.
\newblock \emph{\tt arXiv:1606.02147}, 2017.

\bibitem[Mazzini(2018)]{Mazzini2018GuidedUN}
Davide Mazzini.
\newblock Guided upsampling network for real-time semantic segmentation.
\newblock 2018.

\bibitem[Zhao et~al.(2018)Zhao, Qi, Shen, Shi, and Jia]{Zhao2018ICNetFR}
Hengshuang Zhao, Xiaojuan Qi, Xiaoyong Shen, Jianping Shi, and Jiaya Jia.
\newblock Icnet for real-time semantic segmentation on high-resolution images.
\newblock In \emph{Proceedings of the European Conference on Computer Vision
  (ECCV)}, 2018.

\bibitem[He et~al.(2016)He, Zhang, Ren, and Sun]{He16}
Kaiming He, Xiangyu Zhang, Shaoqing Ren, and Jian Sun.
\newblock Deep residual learning for image recognition.
\newblock In \emph{Proceedings of the IEEE Conference on Computer Vision and
  Pattern Recognition (CVPR)}, pages 770--778, 2016.

\bibitem[Simonyan and Zisserman(2015)]{simonyan2014very}
Karen Simonyan and Andrew Zisserman.
\newblock Very deep convolutional networks for large-scale image recognition.
\newblock In \emph{International Conference on Learning Representations
  (ICLR)}, 2015.

\bibitem[SiteOrigin()]{BackgTextGen}
SiteOrigin.
\newblock Background image generator.
\newblock \url{http://bg.siteorigin.com/}.

\bibitem[Bac()]{BackgTextPatterns}
Subtle patterns.
\newblock \url{https://www.toptal.com/designers/subtlepatterns/}.

\bibitem[Lee and Kolter(2019)]{lee2019physical}
Mark Lee and Zico Kolter.
\newblock On physical adversarial patches for object detection.
\newblock \emph{arXiv preprint arXiv:1906.11897}, 2019.

\bibitem[Tieleman and Hinton(2012)]{tieleman2012rmsprop}
Tijmen Tieleman and Geoffrey Hinton.
\newblock Lecture 6.5-rmsprop: Divide the gradient by a running average of its
  recent magnitude.
\newblock \emph{COURSERA: Neural Networks for Machine Learning 4}, 2012.

\bibitem[Chollet(2017)]{Chollet2016XceptionDL}
François Chollet.
\newblock Xception: Deep learning with depthwise separable convolutions.
\newblock \emph{Proceedings of the IEEE Conference on Computer Vision and
  Pattern Recognition (CVPR)}, 2017.

\end{thebibliography}
}

\clearpage

\section{Further Details on Context Bias Detection on Synthetic Data}

\subsection{Synthetic Dataset}

\begin{figure}[h!]
	\centering
	\includegraphics[width=0.78\linewidth]{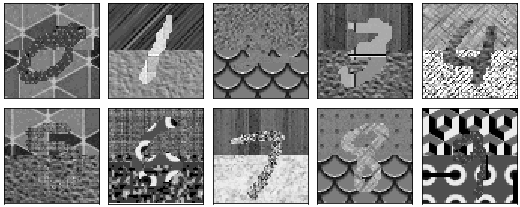}
	\caption{\label{fig:toy_data} Examples from the toy dataset.
	}
\end{figure}

\begin{figure}[h!]
	\centering
	\includegraphics[width=0.8\linewidth]{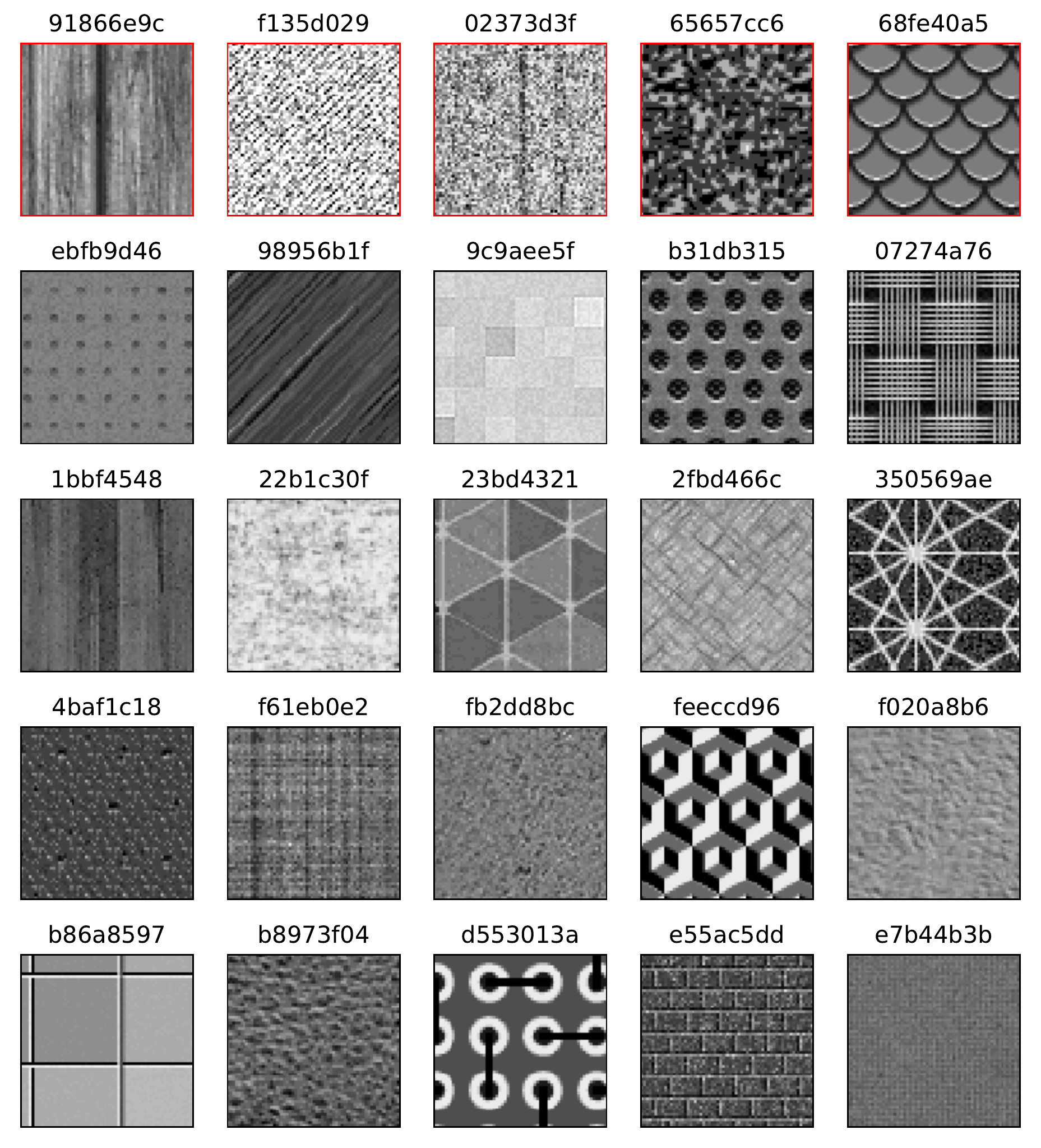}
	\caption{\label{fig:texture_pool} Random crop of each texture in the texture pool. The textures used as bias texture are shown in the first row (red framed).
	}
\end{figure}

\begin{figure}[h!]
	\centering
	\includegraphics[width=0.8\linewidth]{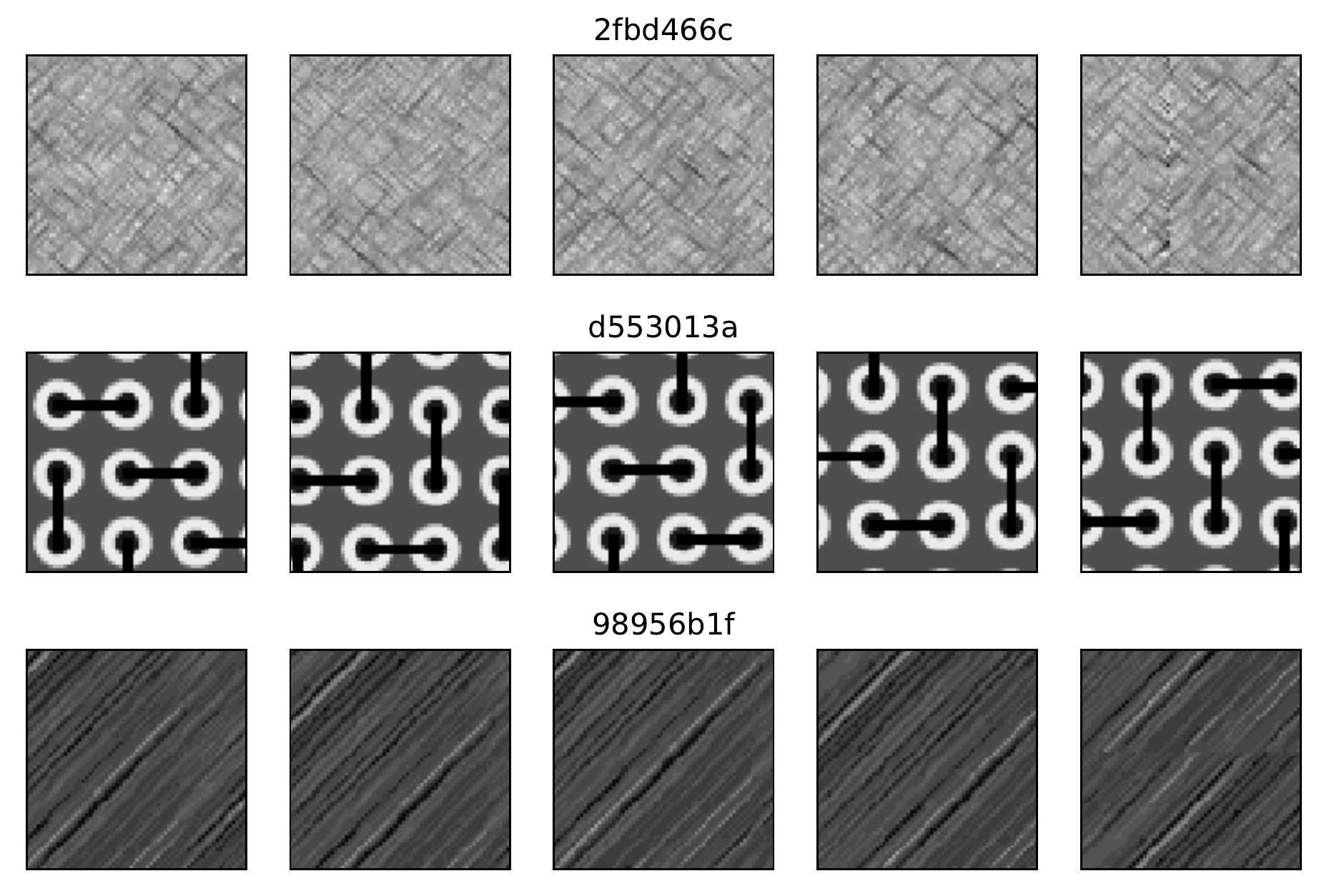}
	\caption{\label{fig:texture_crop_variance} Variance of textures across different random crops.
	}
\end{figure}

The proposed synthetic toy dataset consists of grayscale images of size \(64 \times 64\) pixels,
generated by combining digits from MNIST~\cite{MNIST} with foreground and background textures from  \cite{BackgTextGen,BackgTextPatterns}, as can be seen in Fig. \ref{fig:toy_data}.
Each MNIST image is bilinearly upsampled to \(64 \times 64\) and used as a binary mask for the digit shape (with \(\text{threshold} = 0.5\)), which interior and exterior regions are filled with fore- and background textures, respectively.
In total, a pool of \(25\) textures are considered for fore- and background generation.
To induce the texture variance among instances and make the segmentation task more challenging, we crop the textures randomly from the texture images of at least size \(400 \times 400\) generated via~\cite{BackgTextGen}. One crop for each texture is visualized in Fig. \ref{fig:texture_pool}. The texture variation between different random crops is shown in Fig. \ref{fig:texture_crop_variance}.
For each synthetic image a corresponding segmentation ground truth is generated, where the MNIST digit defines the mask and the semantic class. Overall, \(11\) semantic classes are considered, $10$ digits plus the background class.

For all dataset variants, a training and a test set are generated from the original MNIST training and validation sets respectively, using all \(25\) textures. The training set contains 50k images and the test set consist of 1k images. For the training set, the seed for setting up the random generator can be varied in order to obtain different dataset variants. It specifies the digit order as well as the texture selection. Code to generate the toy dataset is provided under \url{https://github.com/boschresearch/GridSaliency-ToyDatasetGen}. 

\FloatBarrier
\subsection{Experimental Setup}

We utilize the U-Net \cite{Ronneberger2015UNetCN} architecture with VGG16 \cite{simonyan2014very}, Mobilenetv2 \cite{Sandler2018MobileNetV2IR}, and ResNet18 \cite{He16} backbones, with a pixel-wise softmax layer (\(11\) classes) as the final layer.
For training, we use the multi-class cross entropy loss with the RMSPROP \cite{tieleman2012rmsprop} optimizer. The networks are trained for \(50\) epochs, with the batch size \(100\); the learning rate is set to \(10^{-3}\) and \(10^{-4}\) for the first \(40\) and last \(10\) epochs, respectively.

To avoid adversarial artifacts in the estimation of the perturbation-based context saliency, we use the saliency maps of a lower resolution for optimization, i.e. \(4 \times 4\) for \(M_{context}\), and then upsample them to the original size of \(64 \times 64\) using nearest neighbor upsampling for evaluation.
We use a set of different constant perturbation grayscale values \(\{0, 0.25, 0.5, 0.75, 1.0\}\) and for each image choose the one with the lowest loss value for $M = 0$, to avoid hurting the segmentation by having a low contrast between the object and the perturbation image.
In order to avoid optimizing for the border artifacts of the network predictions, for the preservation loss computation, the request mask is eroded with a \(3 \times 3\) kernel size.

The saliency maps are optimized using SGD with a momentum of \(0.5\) and a learning rate of \(0.2\) for \(100\) steps, starting with the initial map \(M = 0.5\) and clipping saliency map values below \(0.2\) to \(0\) at each step in order to avoid artifacts.
A weighting factor of \(\lambda = 0.05\) is used (see Eq. 3).
After the optimization, we check if an empty mask achieves a lower loss value than the optimized one and use the mask with the lowest loss as the final saliency map.

For the IG saliency method, the number of interpolation steps $n$ is set to \(25\) and for the SG saliency, we use \(25\) Gaussian noise samples with \(\mu=0, \sigma = 0.15 (\text{max}(x) - \text{min}(x))\)).

All computations were done on a Nvidia Titan Xp GPU with 12 GB memory.

\FloatBarrier
\subsection{Bias Detection}
\label{subsec:bias_detection}

In this and the following section, we provide a more detailed analysis of the different saliency methods benchmarked on the toy dataset described in the main paper in Sec. 4.  Along with our approach we additionally consider the Vanilla Gradient (VG) \cite{Zeiler2014VisualizingAU}, Integrated Gradient (IG) \cite{Sundararajan2017AxiomaticAF}, and SmoothGrad (SG) \cite{Smilkov2017SmoothGradRN} saliency methods. Instead of aggregating the IoU, context bias detection (CBD) and context bias localization (CBL) metrics over multiple bias textures, we show them separately, to gain some insight how the bias texture may affect the performance of the context saliencies. The results are averaged over 5 different random seeds used for the training data generation.

\begin{figure}[h!]
	\centering
  \input{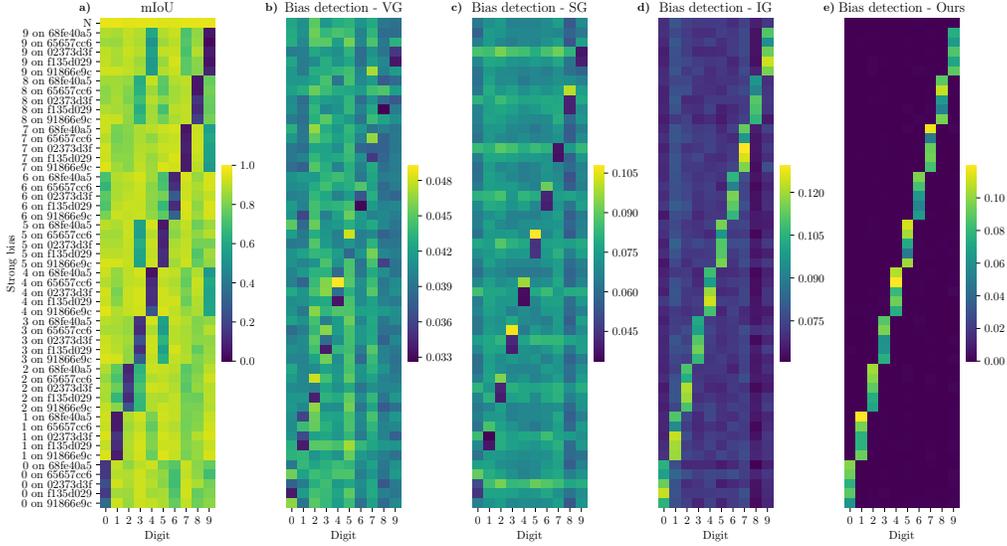}
  \caption{
  Context bias detection comparison of different saliency methods for strongly biased datasets. Instead of averaging over five different bias textures as done in Fig. 3, each bias texture is listed separately to show how the bias texture affects the different saliency methods. The bias detection performance is measured using the CBD metric (see Eq. 6).
  }
  \label{supp:toy_strong_saliencies}
\end{figure}

\begin{figure}[t]
	\centering
  \input{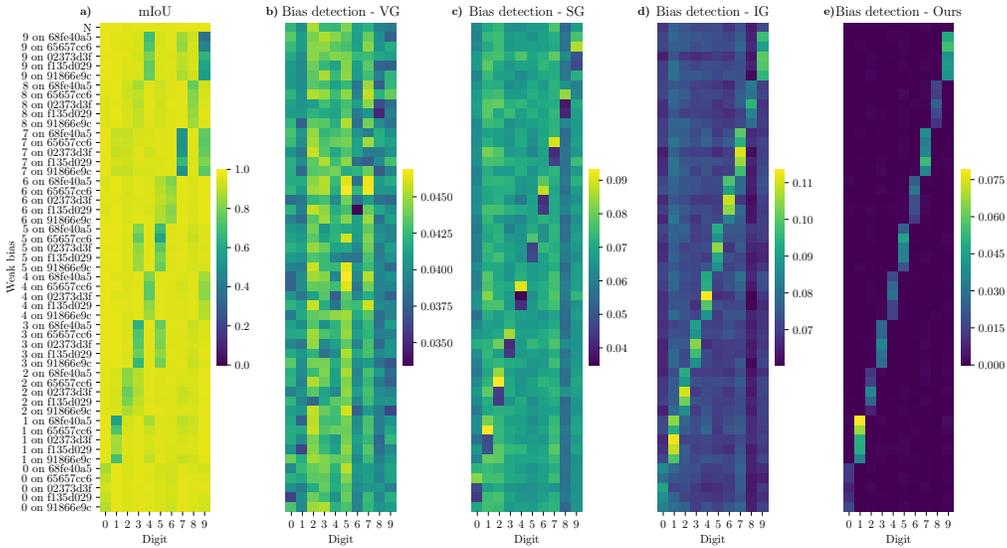}
  \caption{
  Context bias detection comparison of different saliency methods for weakly biased datasets. Instead of averaging over five different bias textures as done in Fig. 3, each bias texture is listed separately to show how the bias texture affects the different saliency methods. The bias detection performance is measured using the CBD metric (see Eq. 6).
  }
  \label{supp:toy_weak_saliencies}
\end{figure}

We evaluate different networks trained on data biased towards a specific digit with a specific bias texture. When testing on the unbiased dataset, there is a drop in the segmentation IoU for the biased digit, which shows that the network has picked up the bias. 
As can be seen in Fig. \ref{supp:toy_strong_saliencies} a) and \ref{supp:toy_weak_saliencies} a) the extent of the drop is stable across different bias textures and is mostly affected by the bias digit.

For the different saliency methods, we have checked if this bias can be detected only using the biased dataset. For VG and SG (see Fig. \ref{supp:toy_strong_saliencies} b/c) and \ref{supp:toy_weak_saliencies} b/c)), the CBD highly deviates between the biased digits (diagonal), with the amount and direction heavily dependent on the bias texture. In practice, these methods are not applicable as there is no control over the bias texture.  By design these gradient-based methods are more sensitive to high frequency patterns and thus lead to unfaithful explanations \cite{Adebayo2018SanityCF}.
For IG and our context saliency (see Fig. \ref{supp:toy_strong_saliencies} d/e) and \ref{supp:toy_weak_saliencies} d/e)), we can observe a significantly smaller dependency on the bias texture allowing a bias detection to be independent from the bias texture.
Moreover, for the weakly biased dataset, our grid saliency also roughly reflects the extent of the bias, which we observed in the segmentation IoU drop on the unbiased dataset (compare Fig. \ref{supp:toy_weak_saliencies} a) and Fig. \ref{supp:toy_weak_saliencies} e)).

Note that the mIoU drop for unbiased digits which look similar to the biased ones (e.g., four and nine) is not reflected in the generated saliency maps. By design, our method only looks for positive evidence (context that is present in the image and supports the classification) without taking into account negative evidence for the prediction. By applying an biased network to an unbiased dataset it is exposed to those negative biases as the bias texture acts as negative evidence for unbiased digits.

\FloatBarrier
\subsection{Bias Localization}

\begin{figure}[h!]
	\centering
  \input{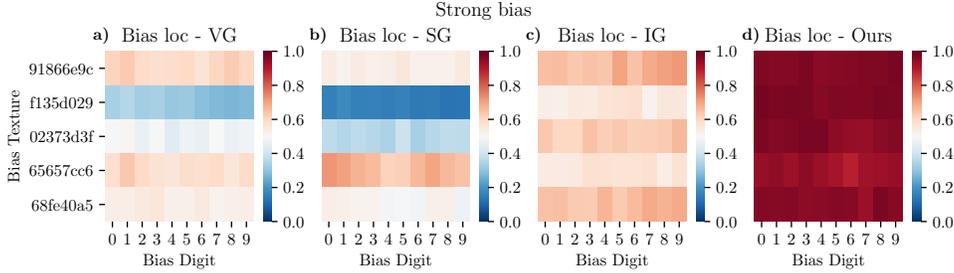}
  \caption{
  Bias localization across different grid saliency methods for strongly biased datasets. The CBL metric (see Eq. 7), averaged over five training dataset generation seeds, is shown with respect to bias texture and bias digit.
  }
  \label{supp:toy_strong_saliencies_loc}
\end{figure}

A similar effect can be seen for bias localization in Fig. \ref{supp:toy_strong_saliencies_loc} and \ref{supp:toy_weak_saliencies_loc}. While VG and SG (column a) and b)) highly depend on the bias texture causing the localization even to focus more on the unbiased half, IG and our method (column c) and d)) depend significantly less on the bias texture. However, IG only achieves a bias localization slightly above random guessing while our method is able to localize both strong and weak biases very well.

\begin{figure}[h!]
	\centering
  \input{figures/suppl_figure_toy_bias_loc_unagg_weak.tikz}
  \caption{
  Bias localization across different grid saliency methods for weakly biased datasets. The CBL metric (see Eq. 7), averaged over five training dataset generation seeds, is shown with respect to bias texture and bias digit.
  }
  \label{supp:toy_weak_saliencies_loc}
\end{figure}

\FloatBarrier
\subsection{Network Comparison}
\begin{figure}[t]
	\centering
	\vspace{-1em}
  \input{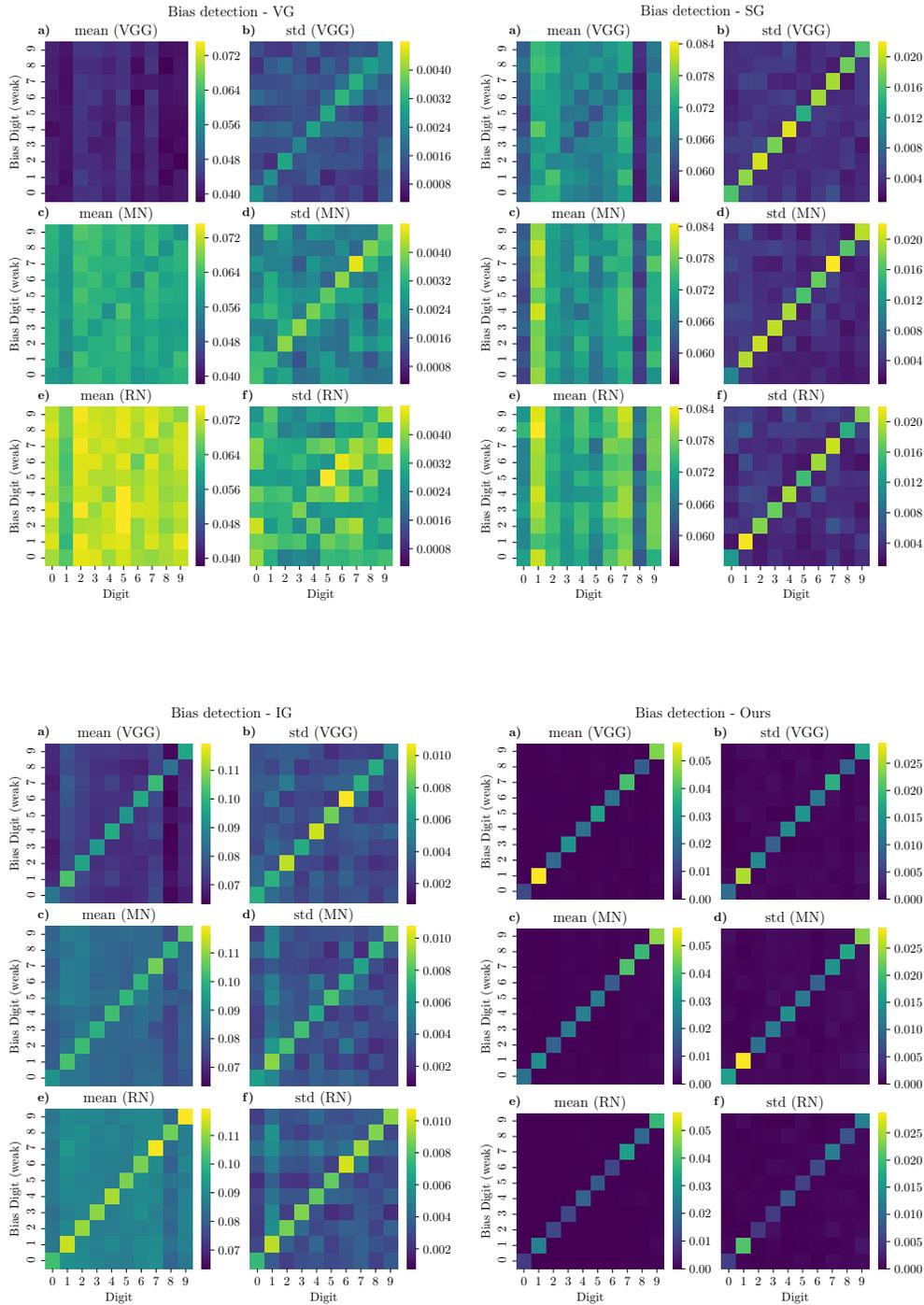}
  \caption{
  Comparison of the different grid saliency methods VG (top-left), SG (top-right), IG (bottom-left), and our perturbation saliency (bottom-right) across different network architectures (from top to bottom: VGG16 (VGG), Mobilenetv2 (MN), and ResNet18 (RN)). For each combination of grid saliency method and architecture, the mean and standard deviation (std) plots for context bias detection (CBD metric) are shown. All values are aggregated over five different bias textures and training set generation seeds.
  }
	\vspace{-0.5em}
  \label{supp:toy_network_comparison}
\end{figure}

\begin{figure}[t]
	\centering
	\vspace{-1em}
	\includegraphics[width=0.7\linewidth]{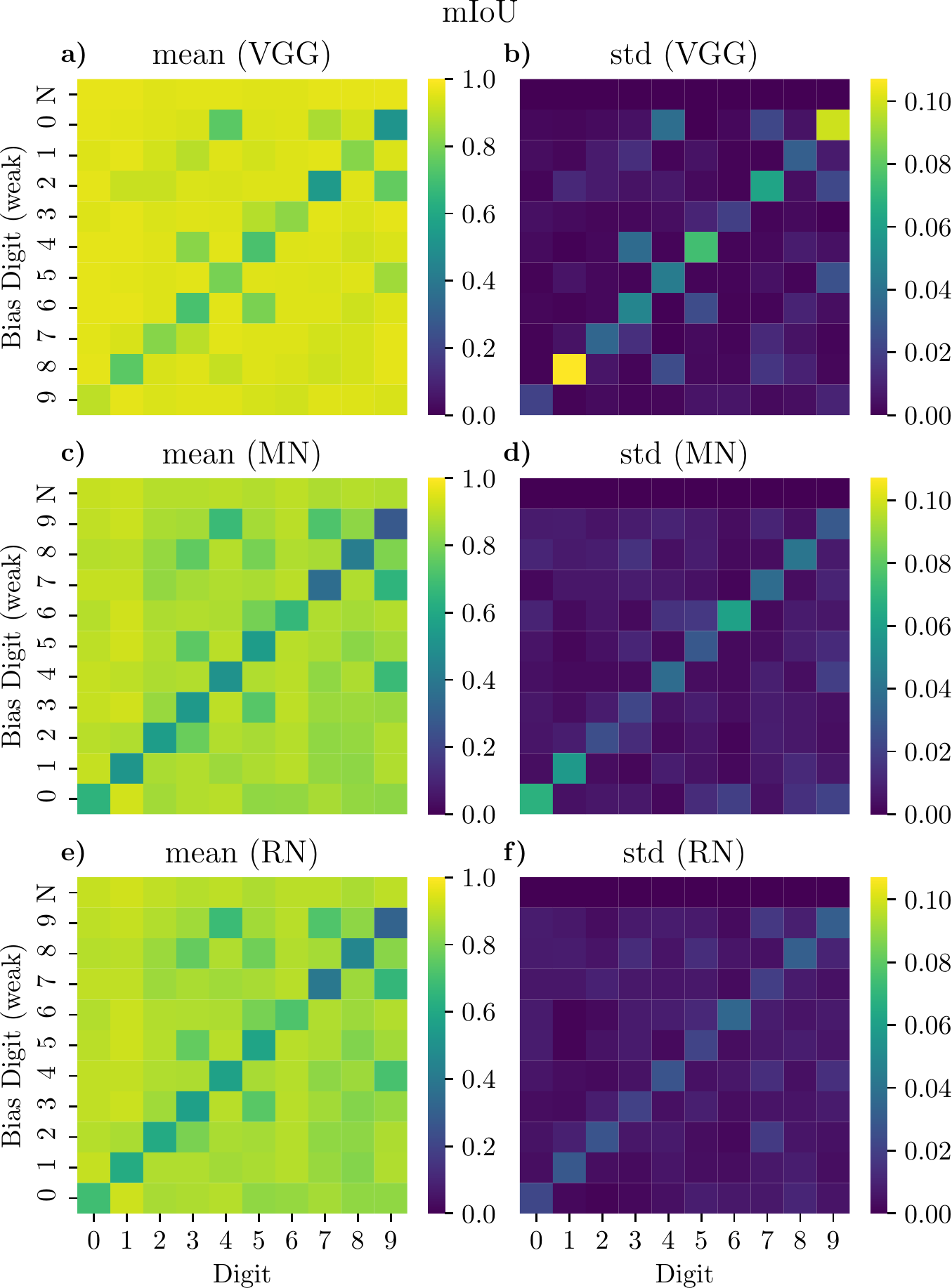}
  \caption{
  Comparison of the segmentation mIoU on the unbiased test set for different network architectures: VGG16 (VGG), Mobilenetv2 (MN), and ResNet18 (RN). All values are aggregated over five different bias textures and training set generation seeds. Both the mean and the standard deviation (std) are visualized. 
  }
	\vspace{-0.5em}
  \label{supp:toy_network_miou}
\end{figure}

\begin{figure}[t]
	\centering
	\vspace{-1em}
  \input{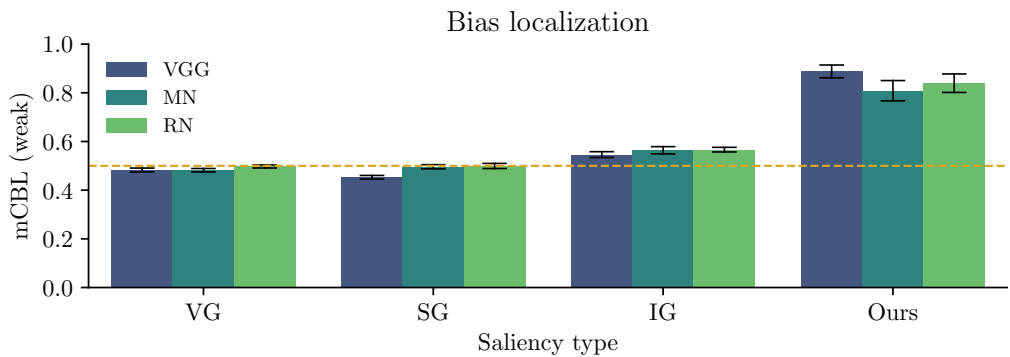}
  \caption{
  Comparison of localization capability for different network architectures (\(x\) axis) using the CBL metric. All values are aggregated over ten bias digits, five different bias textures and five training set generation seeds. The error bars indicate the standard deviation over the bias digits. 
  }
	\vspace{-0.5em}
  \label{supp:toy_network_loc}
\end{figure}

In order to show that our grid saliency can be applied to several network architectures, we have repeated the experiments from Sec. 4.2 of the main paper with different backbones for Unet~\cite{Ronneberger2015UNetCN}. For that purpose, we have chosen VGG16~\cite{simonyan2014very}, ResNet18 (RN) \cite{He16}, and MobileNetv2 (MN) \cite{Sandler2018MobileNetV2IR} due to their different structure. All values are aggregated over five different bias textures and five training set generation seeds showing both the mean and the standard deviation. For the architecture comparison, only the weakly biased dataset variants are used.

\paragraph{Segmentation mIoU on the unbiased test set} 
First, we have checked if the different architectures have picked up the bias from the dataset by applying network instances trained on biased datasets to the unbiased test set and calculated the segmentation mIoU. As can be see in Fig. \ref{supp:toy_network_miou} a), all architectures have a clear drop in mIoU for the biased digits (diagonal) in comparison to the baseline mIoU of the unbiased network (top row N). Therefore, we can conclude that all networks have picked up the bias.
VGG achieves the best base segmentation performance and has also a smaller drop in mIoU for biased digits, meaning that it is less susceptible to a context bias. For that reason, we have chosen this configuration for the main paper.

\paragraph{Bias Detection and Localization}
Next, we have applied the context saliency methods to the different versions of the biased datasets to check if they are able to show the presence of a bias in the biased dataset itself. 

For VG and SG (see Fig. \ref{supp:toy_network_comparison}), the mean over different bias textures is not susceptible to the bias, however, there is a high standard deviation for the biased digits (diagonal) caused by the dependence on the bias texture (see Sec. \ref{subsec:bias_detection}). IG and our method are both able to detect the context bias independent from the chosen network architecture.
For VG, SG, and IG, we also observe different levels of focus on the context for different backbones, independent of the digit and if it is biased or not. 
For the bias localization, our method stably outperforms VG, SG, and IG across different backbones (see Fig. \ref{supp:toy_network_loc}).
For grid saliency we also observe a slight localization improvement for VGG over MN and RN, in contrast to other methods.

\FloatBarrier
\subsection{Effect of Optimization Parameters}

\begin{figure}[H]
	\begin{centering}
		\setlength{\tabcolsep}{0em}
		\renewcommand{\arraystretch}{0}
		\par\end{centering}
	\begin{centering}
		\vspace{0em}
		\hfill{}%
		\begin{tabular}{@{}c@{}c@{}c@{}c@{}}
			
			\includegraphics[width=0.25\linewidth,height=0.16\textheight,height=0.16\textheight]{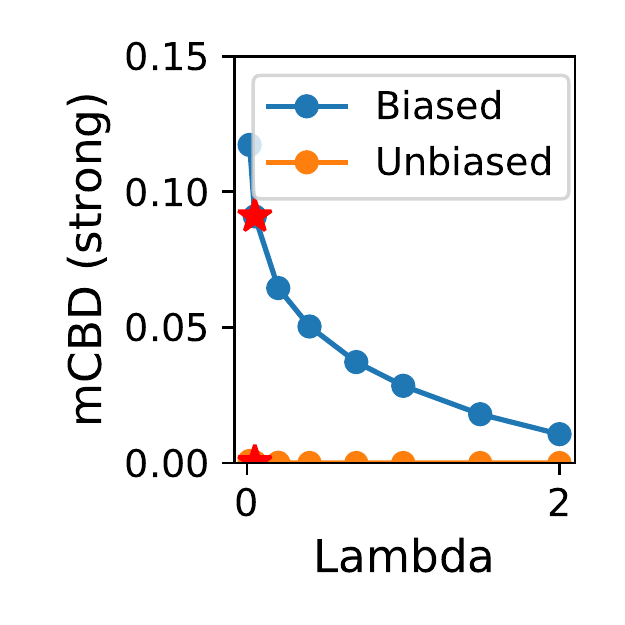} & {\footnotesize{}}
			\includegraphics[width=0.25\linewidth,height=0.16\textheight]{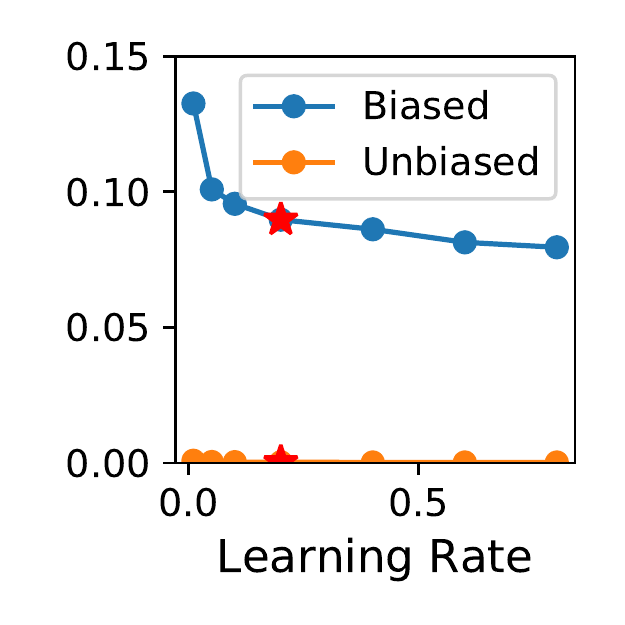} & {\footnotesize{}}
			\includegraphics[width=0.25\linewidth,height=0.16\textheight]{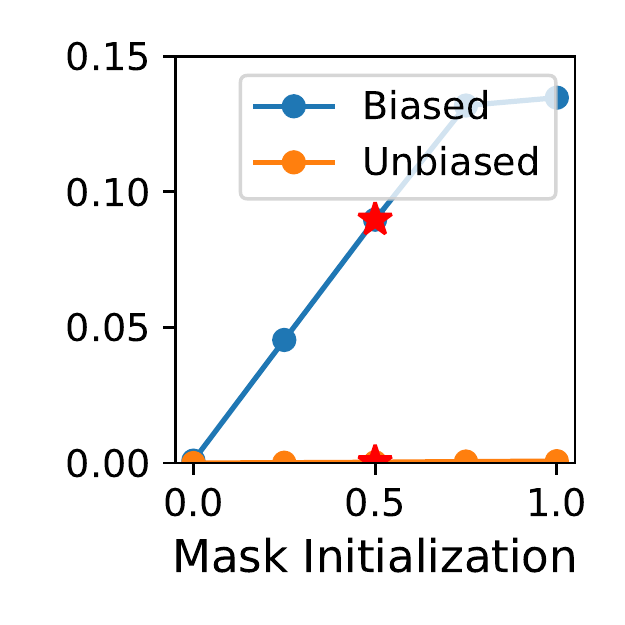}
            \includegraphics[width=0.25\linewidth,height=0.16\textheight]{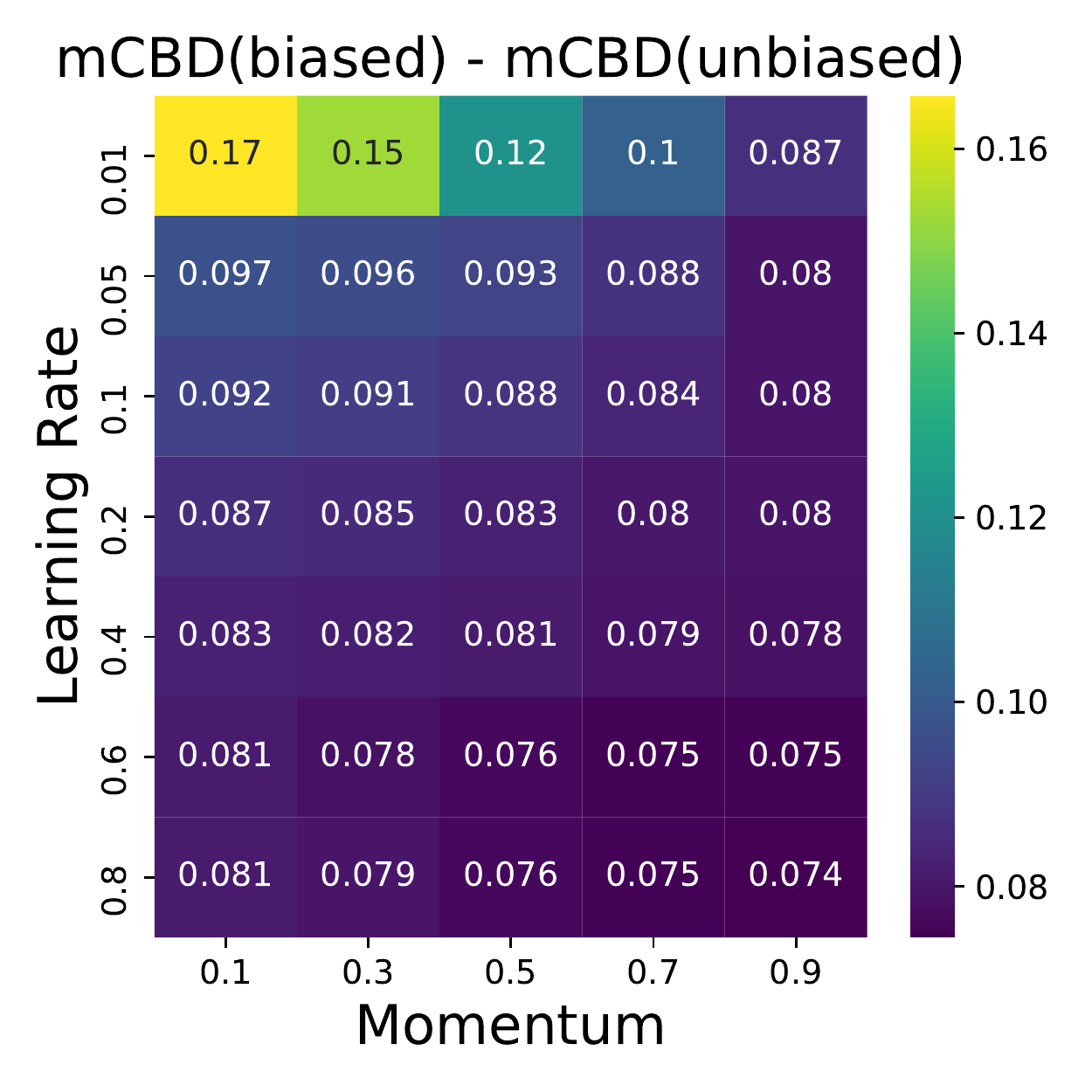}
 			\tabularnewline
 			
			\includegraphics[width=0.25\linewidth,height=0.16\textheight]{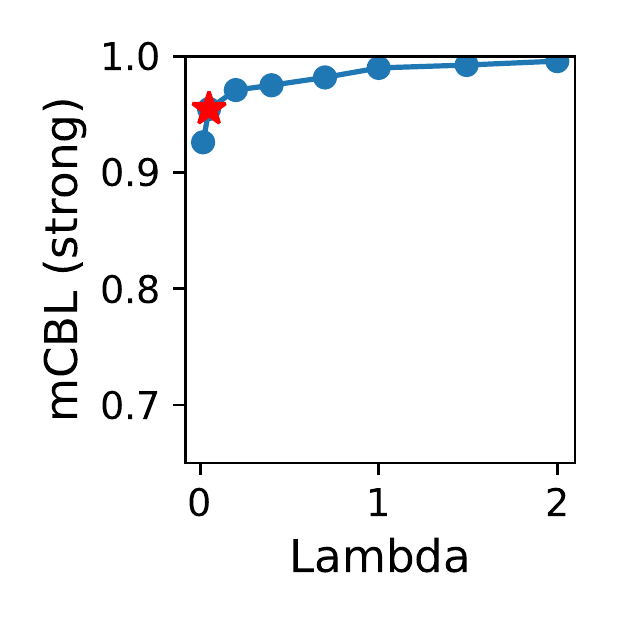} & {\footnotesize{}}
			\includegraphics[width=0.25\linewidth,height=0.16\textheight]{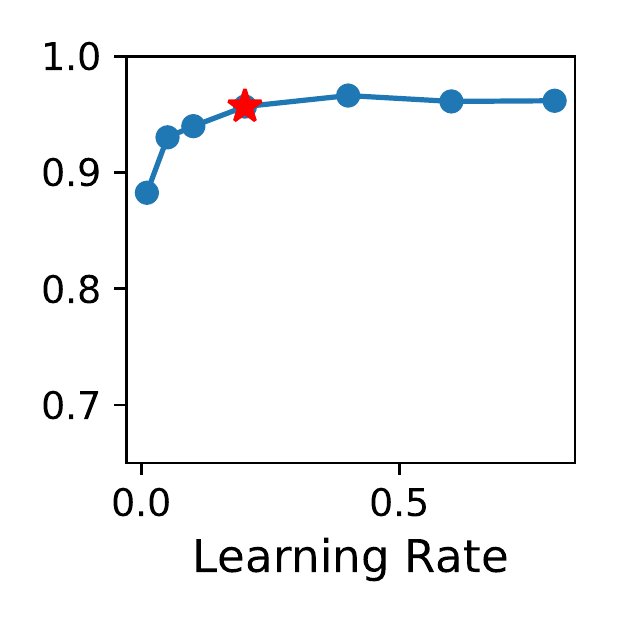} & {\footnotesize{}}
			\includegraphics[width=0.25\linewidth,height=0.16\textheight]{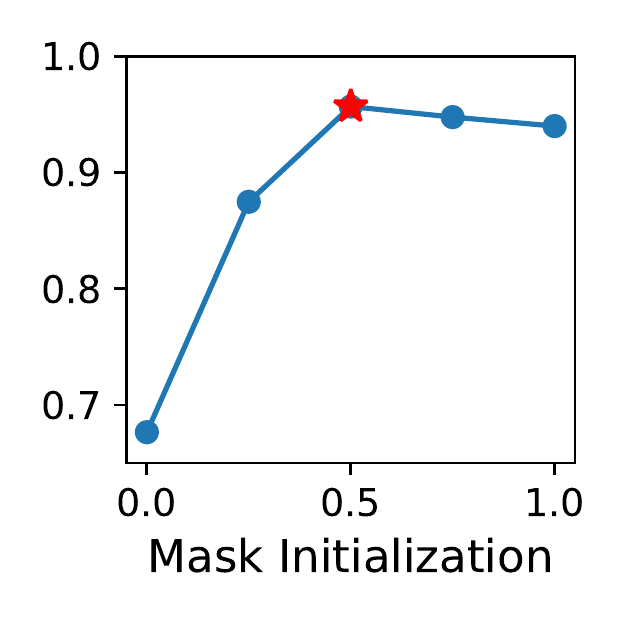}
			\includegraphics[width=0.25\linewidth,height=0.16\textheight]{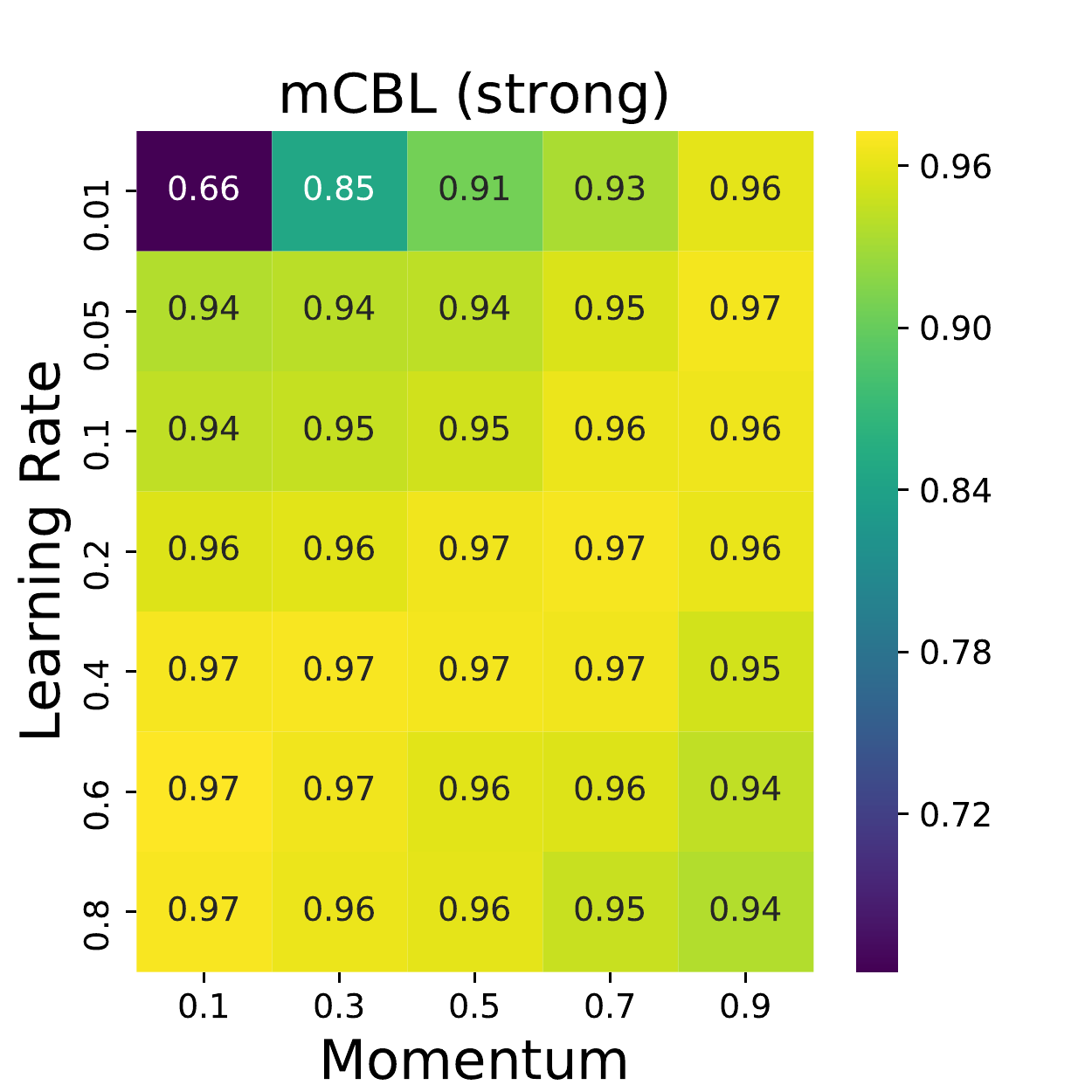}
			\tabularnewline
			
		\end{tabular}\hfill{}
		\par\end{centering}
	\caption{
		Effect of the grid saliency optimization parameters on the quality of context bias detection (mCBD) and localization (mCBL).
		mCBD (top row) and mCBL (bottom row) where computed for different settings of optimization parameters (from left to right): weighting lambda, learning rate, grey value initialization. All values are averaged over five randomly selected bias digits, five bias textures, and three training set generation seeds. The optimization parameters used in our experiments are depicted as red points. One sees, that the chosen parameter setting balances well mCBD and mCBL performances (each high). Note that for mCBD a high difference between biased and unbiased is beneficial.}
	\label{fig:param_sweep}
\end{figure}

In Fig. \ref{fig:param_sweep} we report the effect of optimization parameters for grid saliency on context biased detection and localization performance (CBD, CBL). 
Influence of different optimization parameters on the quality of the obtained context explanations was evaluated on the basis of parameter sweeps around the chosen setting for the loss weighting \(\lambda\), the learning rate and momentum, as well as initialization of the saliency mask (red points in Fig. \ref{fig:param_sweep}). 
In our experiments in the main paper, all optimization parameters were set up by jointly looking at the two loss term values in Eq. 2, Sec. 3.1 and visual inspection of saliencies over a small image subset.

We notice that grid saliency shows comparable performance over a broad space of parameter settings (observing mostly smooth degradation with suboptimal parameter choices), with \(\lambda\) clearly controlling the trade off between bias detection and localization quality (higher \(\lambda\) value leads to a smaller salient region, see Sec. 3.1 of the main paper for method details).
One sees, that the chosen parameter setting (red points in Fig. \ref{fig:param_sweep}) balances well CBD and CBL performances (each high). 

\clearpage
\section{Cityscapes Experiments}

\subsection{Experiment Setup}

\paragraph{Cityscapes Dataset} A central motivation of this study is to better understand the behavior of segmentation networks. Cityscapes images describe rich scenes that make it arguably easy for segmentation models to learn clear context biases. We use the 500 (finely annotated) validation images available for Cityscapes as a starting point. We choose to analyze the validation data instead of training data in order to highlight behaviors that are likely to occur in a real world deployment scenario.

For the semantic segmentation, we have used the state-of-the-art network Deeplabv3+~\cite{Chen2018EncoderDecoderWA} with a Mobilenetv2 backbone~\cite{Sandler2018MobileNetV2IR}. The weights were obtained from the original Deeplab repository\footnote{\url{http://download.tensorflow.org/models/deeplabv3_mnv2_cityscapes_train_2018_02_05.tar.gz}}.

\paragraph{Implementation details}
The computation pipeline used to derive Fig. 6 in the main paper and its detailed version in Fig. \ref{fig:cityscapes_heatmap_pred} requires a selection of hyperparameters and design choices, which we explain here in more detail.

The saliency mask \(M^*_{grid}\) is obtained by optimizing Eq. 2 with SGD, thus there is no guarantee of convergence to a global optimum.
The loss function used in the optimization procedure (see in Eq.2) seeks to find a balance (partially controlled by \(\lambda\)) between penalizing the size of the produced saliency region and the preservation loss, which measures how well the softmax scores inside the request mask $R$ were restored to (at least) their initial values prior to perturbation, i.e. removing the image background.
Typically, this preservation loss can be interpreted as a percentage relating to how much of the original softmax score was restored. So, for a loss of 0.1 or smaller, 90\% or more of the original softmax activation scores must be restored. Samples that do not converge to a preservation loss of 0.1 or smaller are ignored in the computation.
Similarly to the synthetic dataset, bilinear upsampling is used to upsample the optimized mask to the input image size.

\begin{figure}[H]
	\centering
	\includegraphics[width=0.7\textwidth]{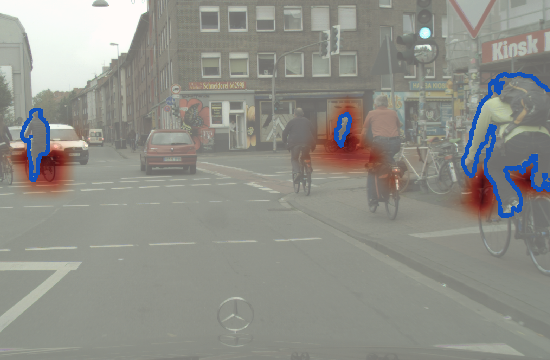}
	\caption{Influence of the size of the request mask $R$ on context explanations provided by grid saliency. Independent of the request mask $R$ size, for all riders salient context always falls on bikes. Note that the context grid saliency explanations were calculated for each request region separately and are only combined together for the visualization.}
	\label{fig:request_size}
\end{figure}

The preservation loss in Eq. 2 is by definition normalized by the size of $R$, thus the size of $R$ does not directly influence the optimization convergence.
In Fig. \ref{fig:request_size} we show the effect of the size of $R$, where context saliencies for each request mask $R$ were obtained with the same optimization parameters. Independent of the request mask $R$ size, for all riders salient context always falls on bikes.

\begin{figure}
 	\begin{centering}
 		\setlength{\tabcolsep}{0.1em}
 		\renewcommand{\arraystretch}{0}
 		\par\end{centering}
 	\begin{centering}
 		\hfill{}%
 		\begin{tabular}{@{}c@{}c@{}c@{}}
 			\includegraphics[width=0.45\textwidth]{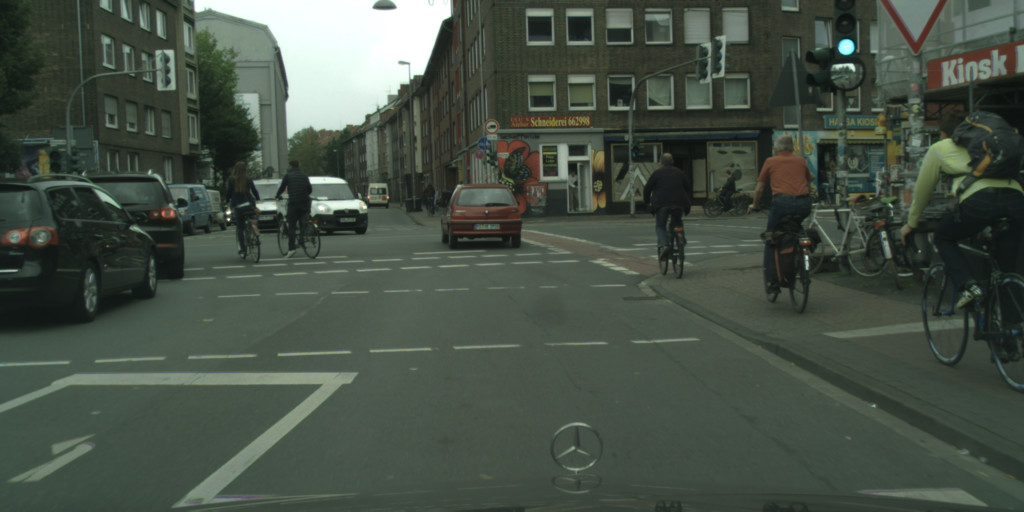} & {\footnotesize{}}
 			\includegraphics[width=0.45\textwidth]{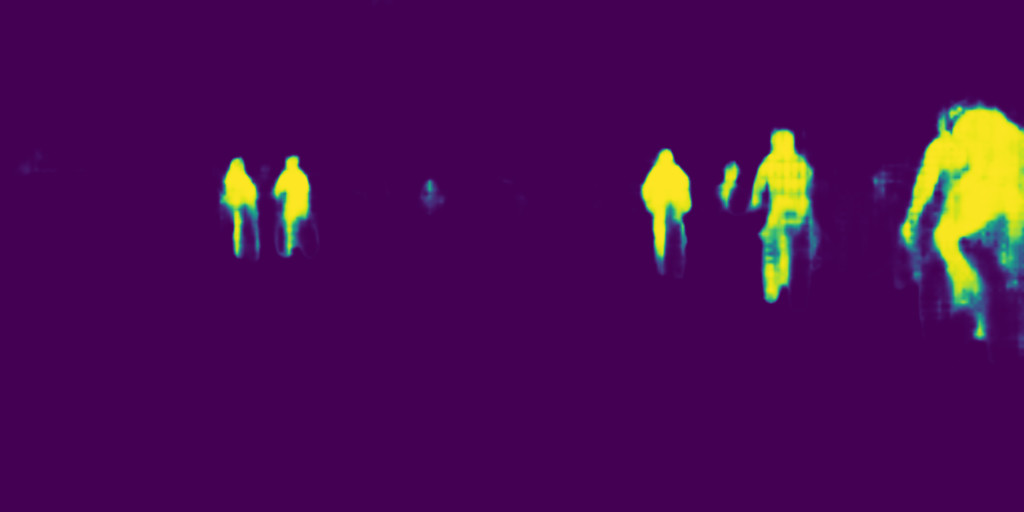}\tabularnewline
			a) Input frame & b) Softmax output on input frame\tabularnewline

 			\includegraphics[width=0.45\textwidth]{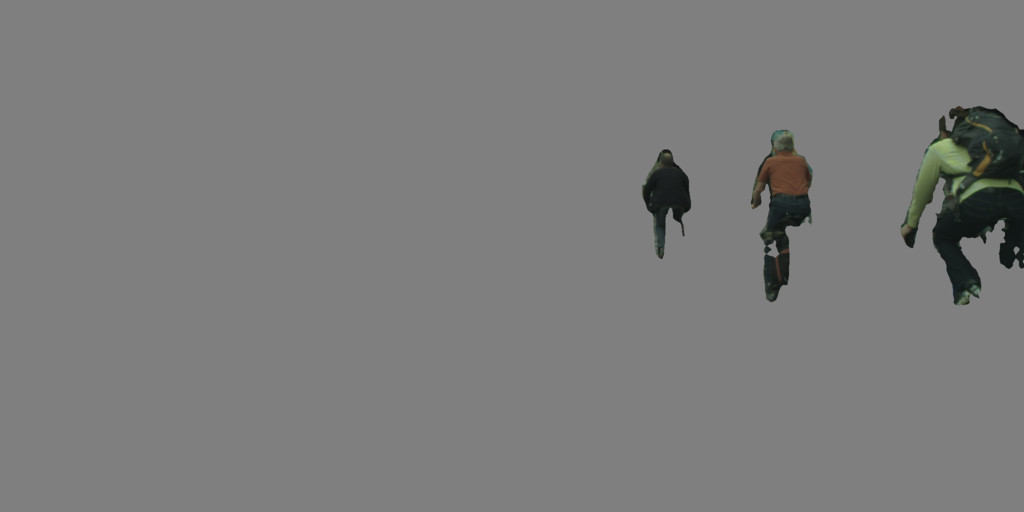}& {\footnotesize{}}
			\includegraphics[width=0.45\textwidth]{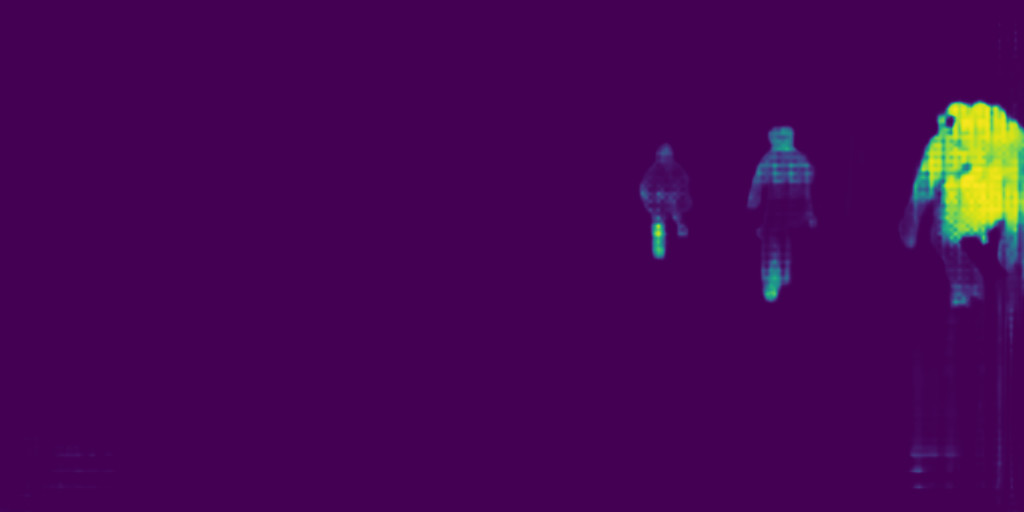}\tabularnewline
			c) Perturbed context & d) Softmax output on peturbed context \tabularnewline

			\includegraphics[width=0.45\textwidth]{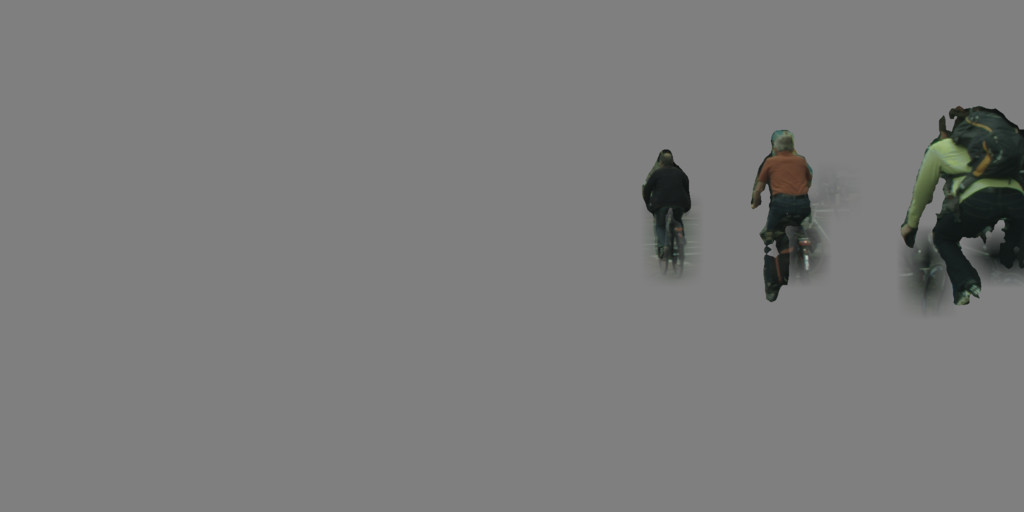} & {\footnotesize{}}
			\includegraphics[width=0.45\textwidth]{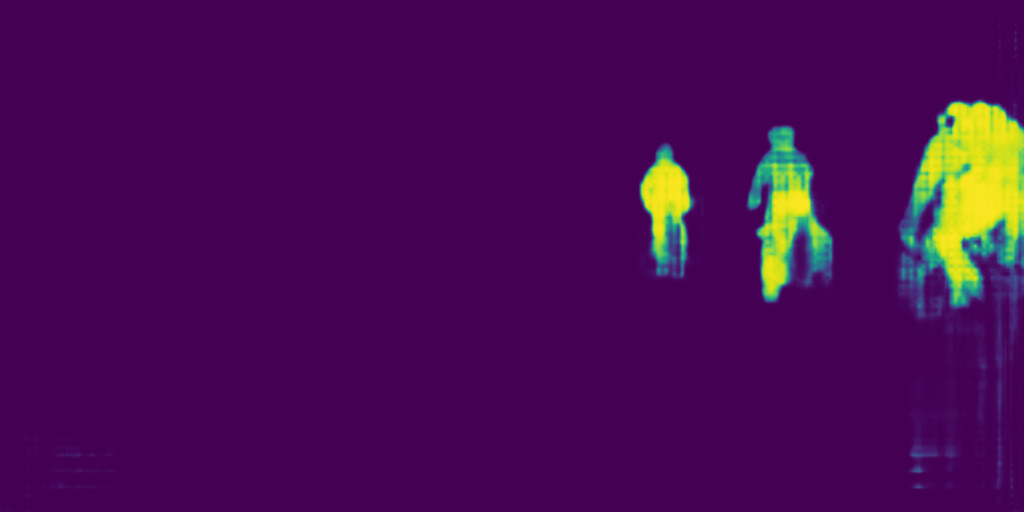}\tabularnewline
			e) Optimized image & f) Softmax output on optimized image \tabularnewline

			\includegraphics[width=0.45\textwidth]{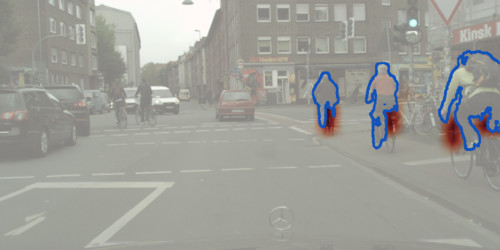} & {\footnotesize{}}
			\includegraphics[width=0.45\textwidth]{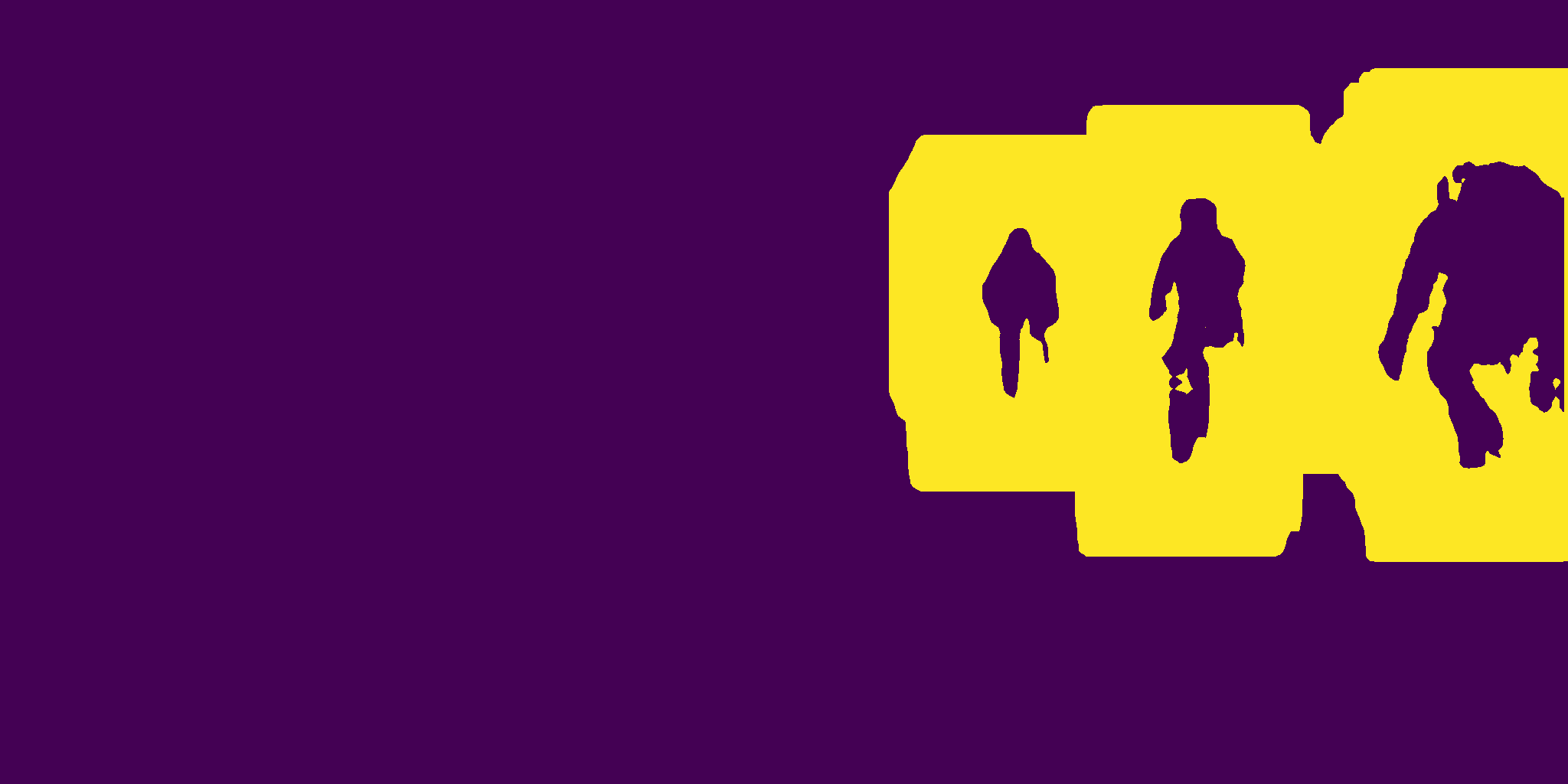}\tabularnewline
			g) Optimized grid saliency map & h) Baseline fixed contour dilation \tabularnewline
 		\end{tabular}\hfill{}
 		\par\end{centering}
	\caption{\label{fig:cityscapes_saliency_intermediate} Intermediate visualizations for the rider class request mask on an input frame.
	}
	\vspace{0em}
\end{figure}

\begin{figure}[h!]
	\centering
%
%

	\includegraphics[width=1.0\textwidth]{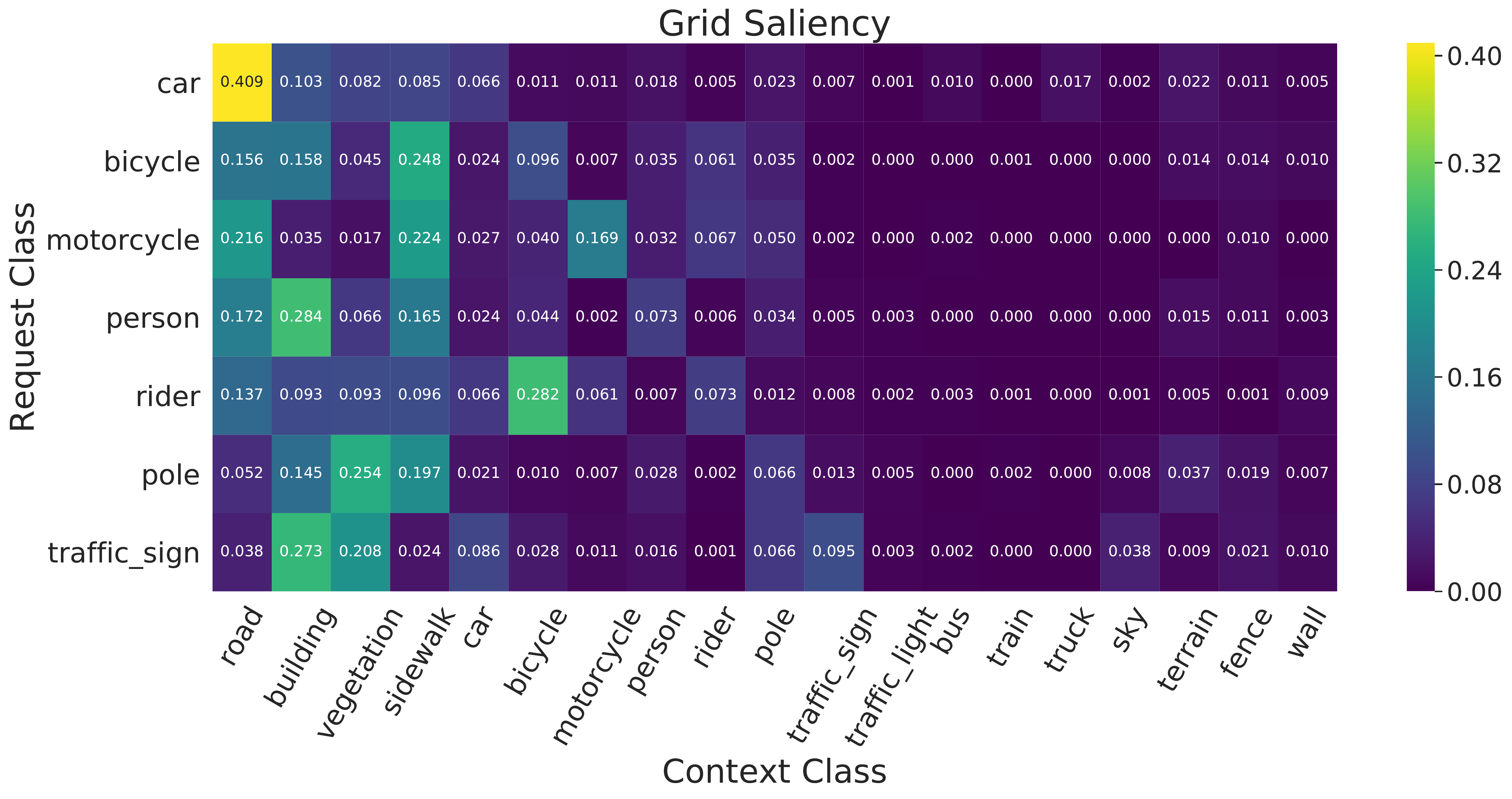}
    \vspace{-0.5em}

   \includegraphics[width=1.0\textwidth]{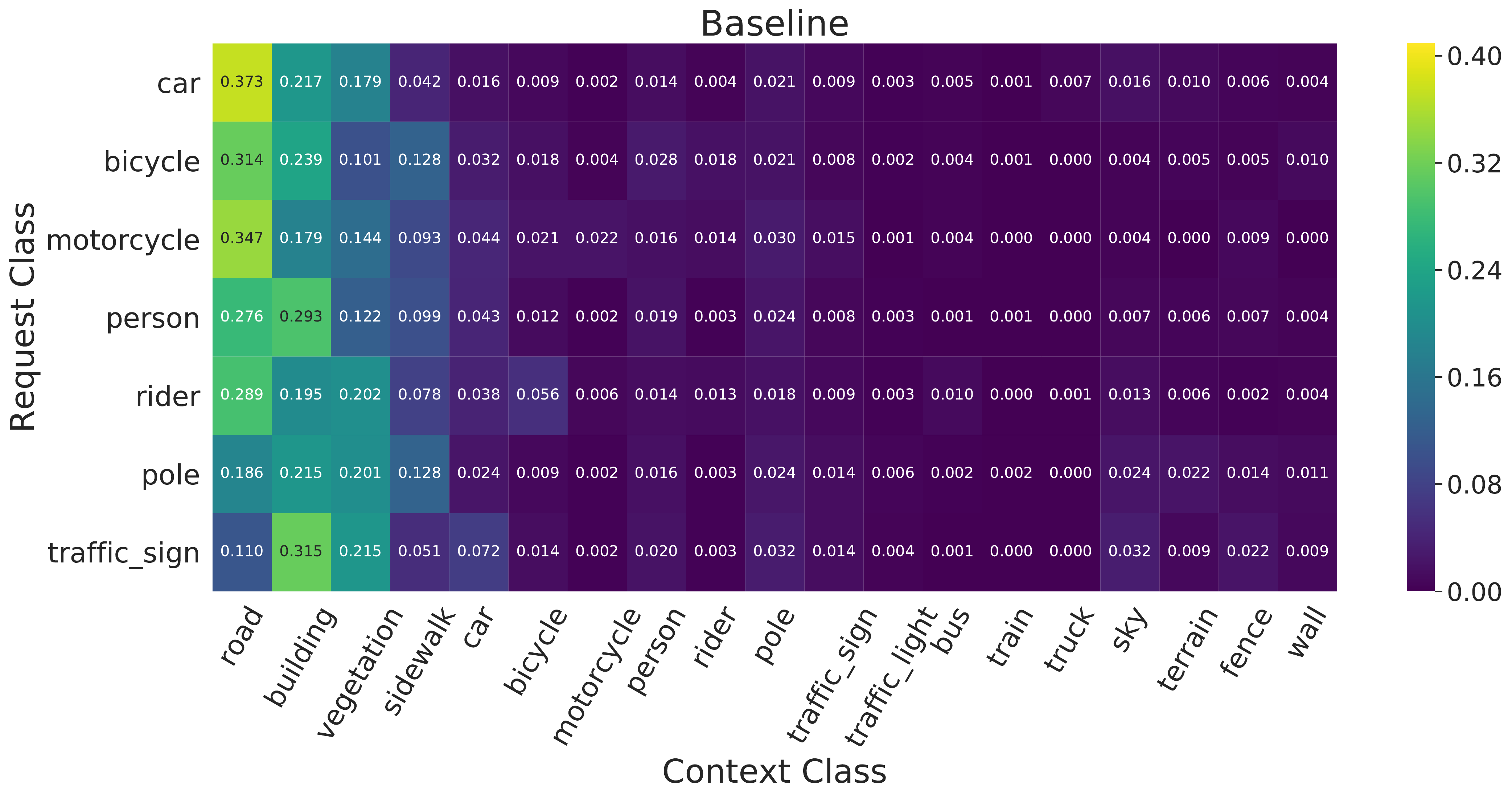}
   \vspace{-0.5em}

   \includegraphics[width=1.0\textwidth]{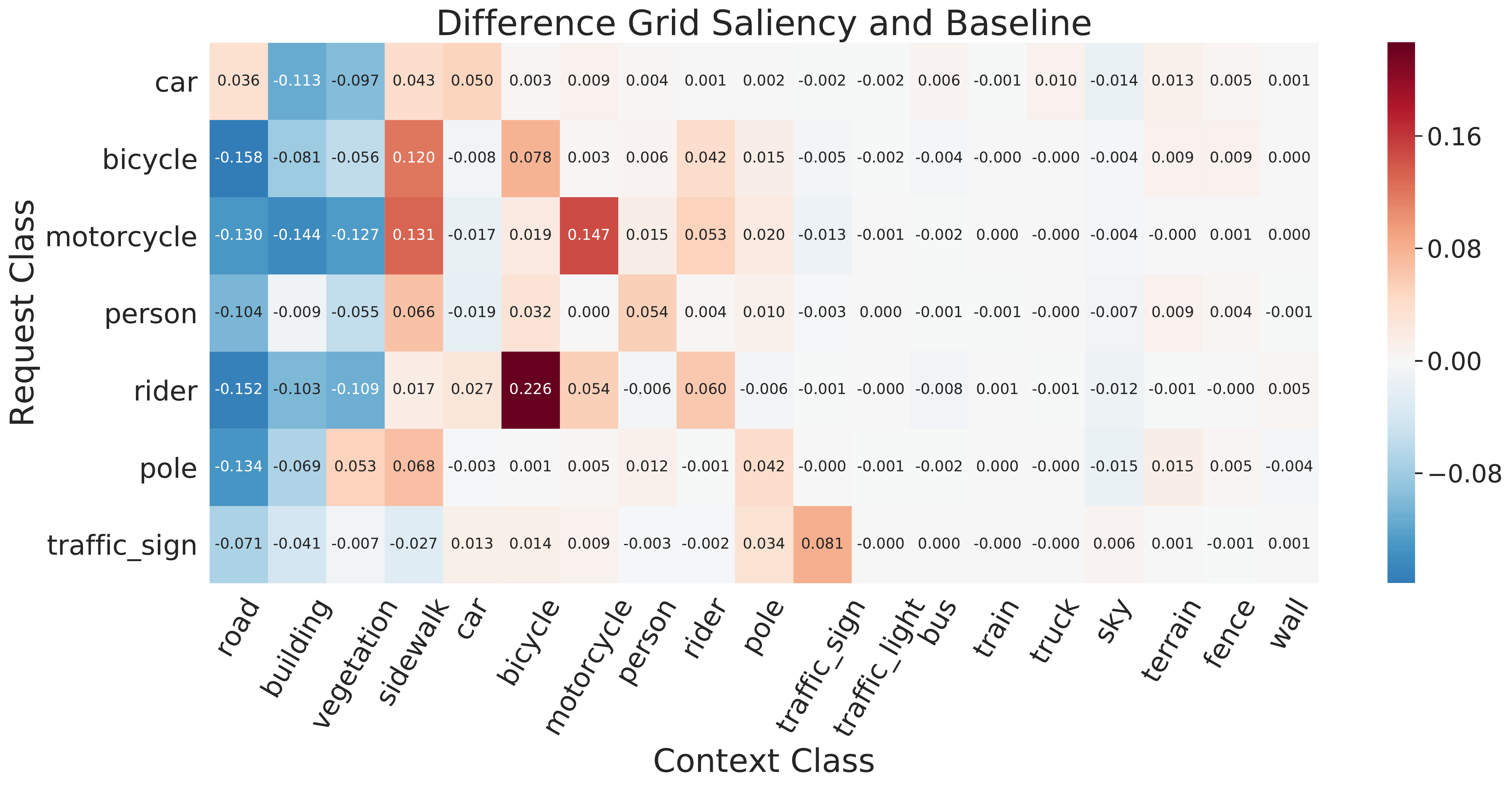}
   \vspace{-0.5em}

	\caption{\label{fig:cityscapes_heatmap_pred} Comparison of saliency distributions for a selected set of classes across the Cityscapes validation data. Note that the rows of the \textit{Grid Saliency} and \textit{Baseline} maps do not sum up to \(1\), because only a subset of classes is considered.
}
	\vspace{0em}
\end{figure}

Fig. \ref{fig:cityscapes_saliency_intermediate} shows some intermediate results for the analysis of a single frame, which illustrates the method and the optimization in a slightly more detailed way.  The components include (a) the input frame, (b) the softmax scores for the "rider" class on the input frame, (c) the image with perturbed context given a request mask for the "rider" class, (d) the softmax output score of (c), (e) the optimized image (request mask plus learned background context), (f) the output softmax scores on (e) as an input, (g) the learned grid saliency, and (h) the dilation around the prediction used for the comparison to baseline.

Generally, penalizing the size of the saliency region is not sufficient to avoid spurious activations. We find that in general, the saliency mask at convergence represents a superset of the important pixels, i.e. it is often possible to slim down the saliency further without severely harming the preservation loss. Rather than adding regularizers directly in the loss function as \cite{Fong2017InterpretableEO}, we simply clip all mask activation values smaller than or equal to 0.2 back to 0.0 for each step of the optimization. We find this leads to considerably less noisy activations and more focused and spatially coherent saliencies. Moreover, we manage potential border artifacts of the network predictions by eroding the request mask of each instance with a \(3 \times 3\) erosion kernel.

Optimizing a low resolution saliency mask makes it hard to deal with small object instances in images. Specifically, a single pixel of the coarse mask already corresponds to a relatively wide spatial context for small objects in the background of a scene. Because we count class labels in the activated saliency, this adds significant noise to the resulting statistics. In order to counteract this, we remove any connected components of the predicted object instances that have a pixel count smaller than $10k$ (or $625$ for context explanations of erroneous predictions) from our request masks.
This filters out a considerable number of images from the heatmap computation.
Table \ref{tab:table_cityscapes_frames_used} and Table \ref{tab:table_cityscapes_frames_used2} count the number of samples available for each class that pass the selection criteria used for the experiments.

The baseline metric dilates the contours of the predicted instances in a given image that pass the minimum size requirement using a total of $400k$ pixels for the dilated region. If multiple instances are present, these dilation pixels are shared amongst them. The number was chosen so as to be able to capture a meaningful portion of an object's spatial context, even if said object is large and in the foreground. An example baseline mask with the fixed dilation can be seen in Fig. \ref{fig:cityscapes_saliency_intermediate} (h).

\begin{table}[h!]
	\begin{center}
		\caption{Number of images in the Cityscapes validation dataset used to compute context grid saliency of the model predictions for each request class. For each request class, an image is included if and only if its computed grid saliency is not empty or invalid.
			This is the case only if the following conditions are met: 1. The request mask for the image contains at least one connected component larger than $10k$ pixels after applying the border erosion kernel. 2. The grid saliency computation does not converge to an empty context saliency mask. 3. The saliency optimization procedure converges with a preservation loss of $0.1$ or less.\\}
		\label{tab:table_cityscapes_frames_used}
		\begin{tabular}{c|c}

		\footnotesize	Request (prediction) class &  \footnotesize Number of samples\\
			\hline
		\footnotesize	rider & \footnotesize 39 \\
		\footnotesize	car & \footnotesize 249 \\
		\footnotesize	person & \footnotesize 45 \\
		\footnotesize	bicycle & \footnotesize 105 \\
		\footnotesize	motorcycle & \footnotesize 16 \\
		\footnotesize	pole & \footnotesize 162 \\
		\footnotesize	traffic sign & \footnotesize 81 \\

		\end{tabular}
	\end{center}
\end{table}

\begin{table}[h!]
	\vspace{-2em}
	\setlength{\tabcolsep}{0.2em}
	\renewcommand{\arraystretch}{1.0}
	\centering
	\caption{Number of images in the Cityscapes validation dataset used to compute context grid saliency of the correct and erroneous model predictions. For each class, an image is included if and only if its computed grid saliency is not empty or invalid. This is the case only if the following conditions are met: 1. The request mask for the image contains at least one connected component larger than $625$ pixels after applying the border erosion kernel. Here the request mask consists of the intersection between the ground truth specified in the first column and the model prediction specified in the second column. As these segments are typically much smaller and more disconnected, the above threshold was reduced to from $10k$ to $625$ pixels.
		2. The grid saliency computation does not converge to an empty context saliency mask. 3. The saliency optimization procedure converges with a preservation loss of $0.1$ or less. \\}
	\label{tab:table_cityscapes_frames_used2}
	\begin{tabular}{cc|c}
		\multirow{2}{*}{\footnotesize GT class }	& \multirow{2}{*}{\footnotesize Prediction class} & \multirow{2}{*}{ \footnotesize Number of samples} \tabularnewline
	&   &  \tabularnewline
		\hline 	\hline

				\multirow{2}{*}{\footnotesize rider} & \footnotesize \textcolor{darkgreen}{rider} & \footnotesize $160$ \tabularnewline

	& \footnotesize \textcolor{red}{person}   & \footnotesize $67$ \tabularnewline

				\hline

					\multirow{3}{*}{\footnotesize person} & \footnotesize \textcolor{darkgreen}{person}  & \footnotesize $217$   \tabularnewline

		 & \footnotesize \textcolor{red}{rider}   & \footnotesize $45$ \tabularnewline

		 & \footnotesize \textcolor{red}{car}   & \footnotesize $29$ \tabularnewline

				\hline

					\multirow{2}{*}{\footnotesize bicycle} & \footnotesize \textcolor{darkgreen}{bicycle}  & \footnotesize $262$ \tabularnewline

		 & \footnotesize \textcolor{red}{motorcycle}   & \footnotesize $12$ \tabularnewline

					\hline

	      	\multirow{3}{*}{\footnotesize car} & \footnotesize \textcolor{darkgreen}{car}  & \footnotesize $323$ \tabularnewline

	  & \footnotesize \textcolor{red}{motorcycle}   & \footnotesize $8$ \tabularnewline

	 & \footnotesize \textcolor{red}{person}   & \footnotesize $33$ \tabularnewline

			\hline

		  \multirow{3}{*}{ 	\footnotesize motorcycle} & \footnotesize \textcolor{darkgreen}{motorcycle}  & \footnotesize $44$ \tabularnewline

	  & \footnotesize \textcolor{red}{bicycle}   & \footnotesize $18$ \tabularnewline

		  & \footnotesize \textcolor{red}{car}   & \footnotesize $9$ \tabularnewline

				\hline

	\multirow{3}{*}{	\footnotesize pole }& \footnotesize \textcolor{darkgreen}{pole}  & \footnotesize $410$ \tabularnewline

	 & \footnotesize \textcolor{red}{building}   & \footnotesize $324$ \tabularnewline

	 & \footnotesize \textcolor{red}{vegetation}   & \footnotesize $145$ \tabularnewline

	\end{tabular}
	\vspace{-1.3em}
\end{table}

\subsection{Further Quantitative Examples}

\begin{figure}[t!]
 	\begin{centering}
 		\setlength{\tabcolsep}{0.1em}
 		\renewcommand{\arraystretch}{0}
 		\par\end{centering}
 	\begin{centering}
 		\hfill{}%
 		\begin{tabular}{@{}c@{}c@{}c@{}}
 			\includegraphics[width=0.31\textwidth]{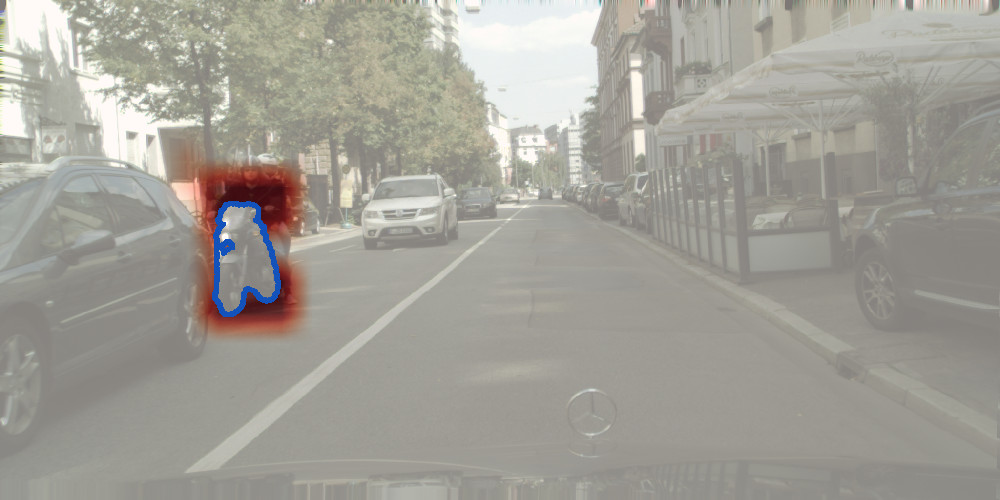} & {\footnotesize{}}
 			\includegraphics[width=0.31\textwidth]{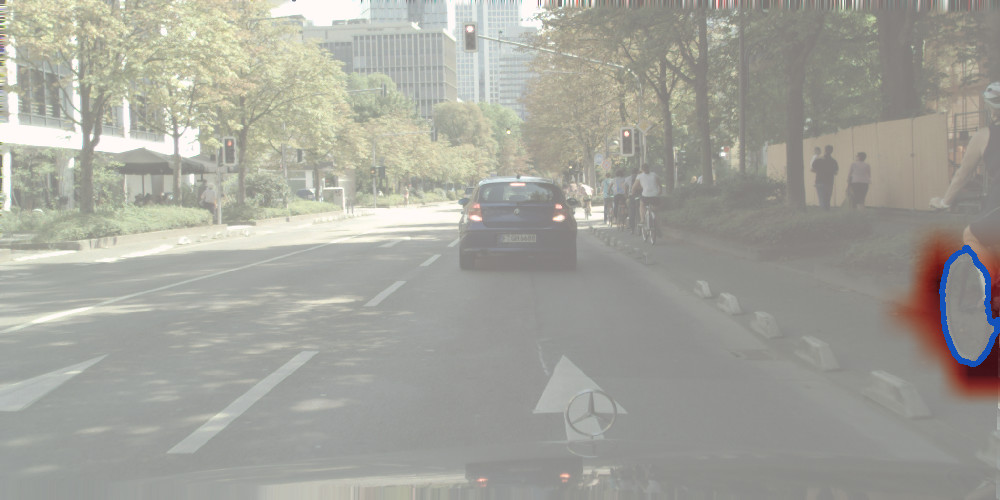}& {\footnotesize{}}
 			\includegraphics[width=0.31\textwidth]{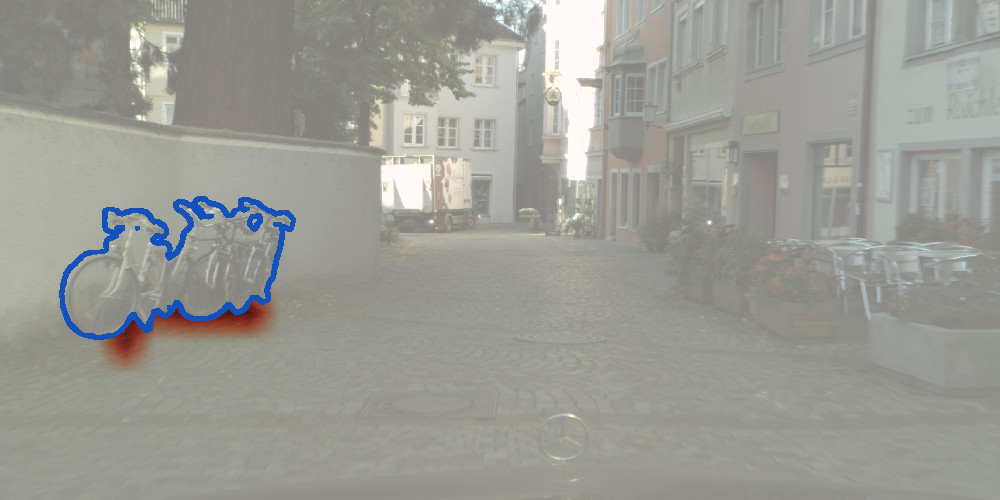}\tabularnewline
			a) Motorbike & b) Bicycle & c) Bicycle Group\tabularnewline

 			\includegraphics[width=0.31\textwidth]{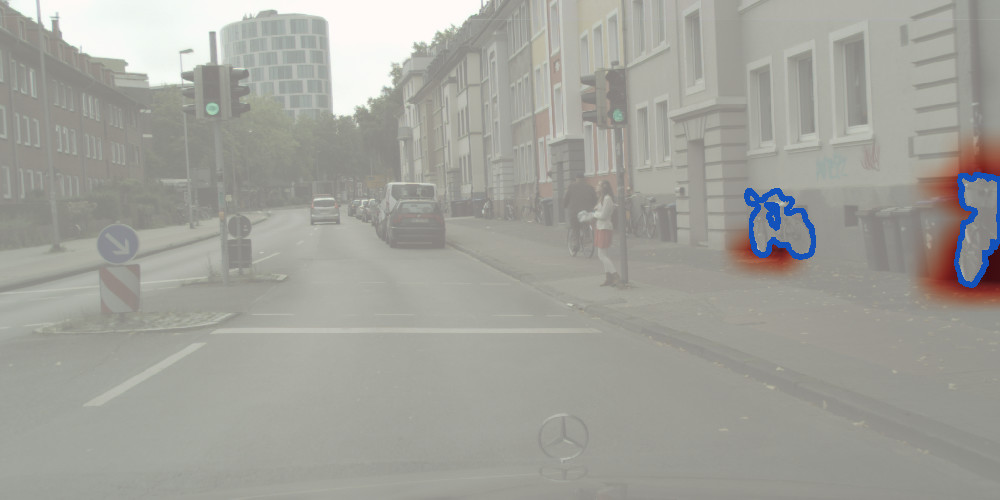}& {\footnotesize{}}
			\includegraphics[width=0.31\textwidth]{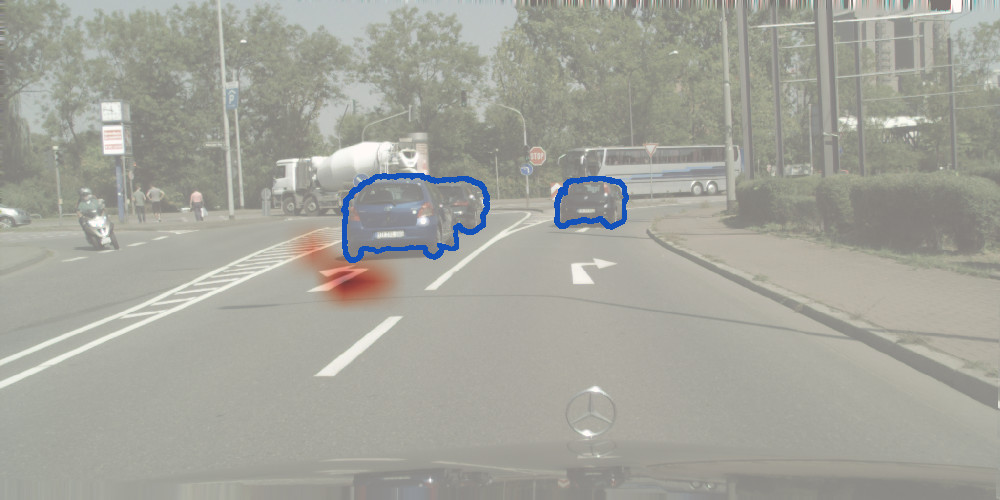} & {\footnotesize{}}
			\includegraphics[width=0.31\textwidth]{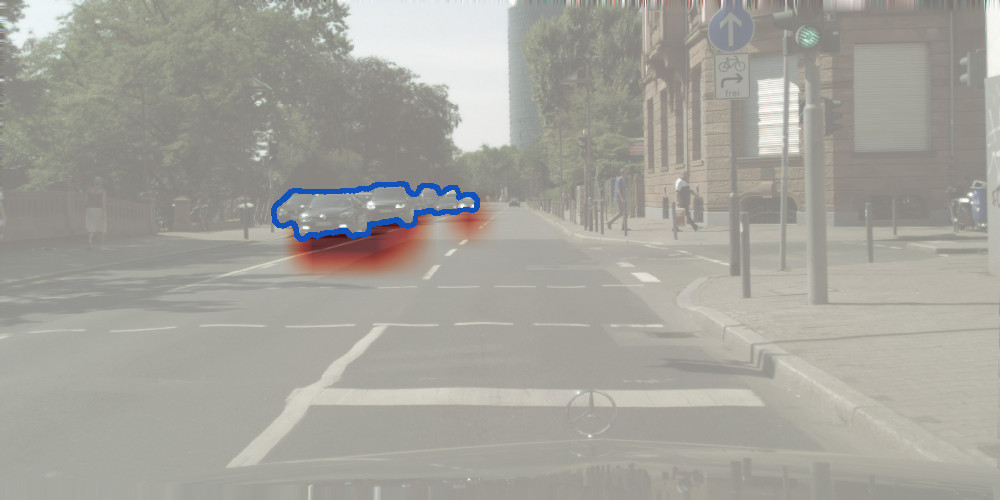}\tabularnewline
			d) Bicycle & e) Car & f) Car Group\tabularnewline

			\includegraphics[width=0.31\textwidth]{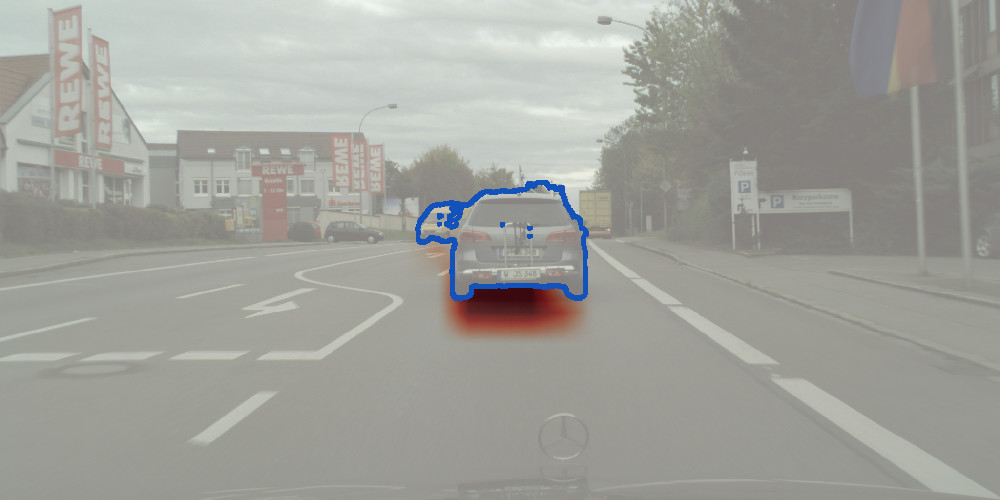} & {\footnotesize{}}
			\includegraphics[width=0.31\textwidth]{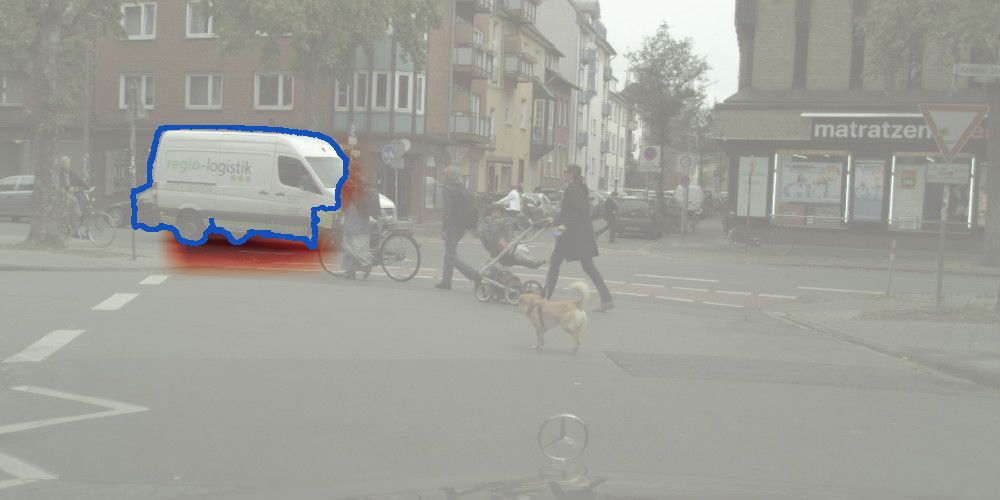} & {\footnotesize{}}
			\includegraphics[width=0.31\textwidth]{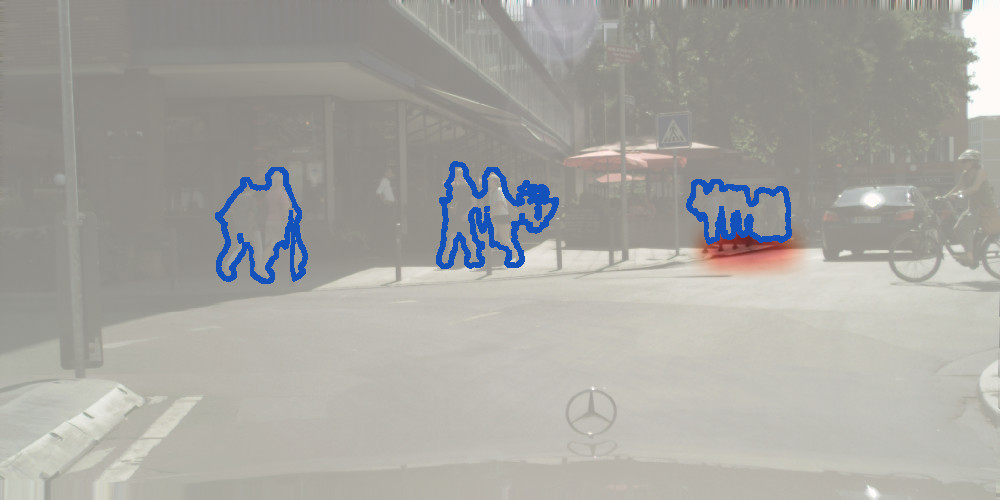}\tabularnewline
			g) Car & h) Large Car & i) Pedestrian Groups\tabularnewline

			\includegraphics[width=0.31\textwidth]{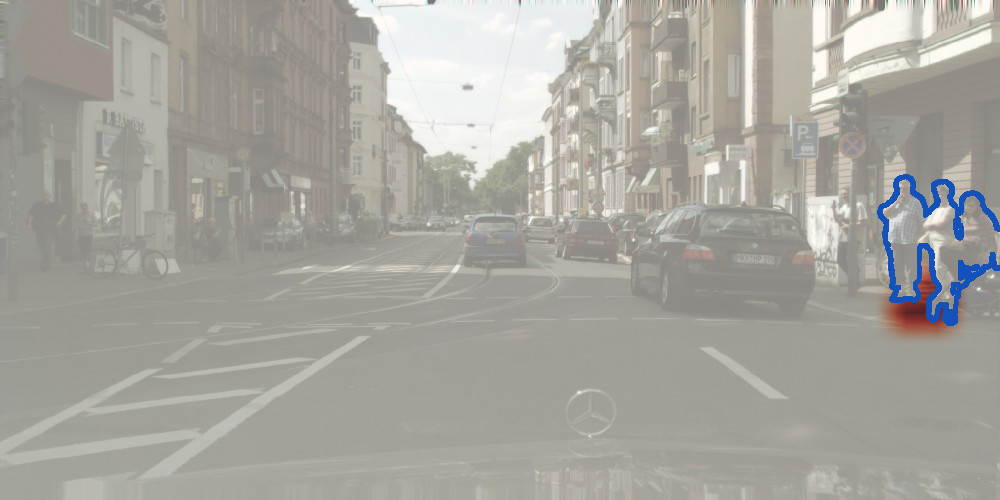}& {\footnotesize{}}
			\includegraphics[width=0.31\textwidth]{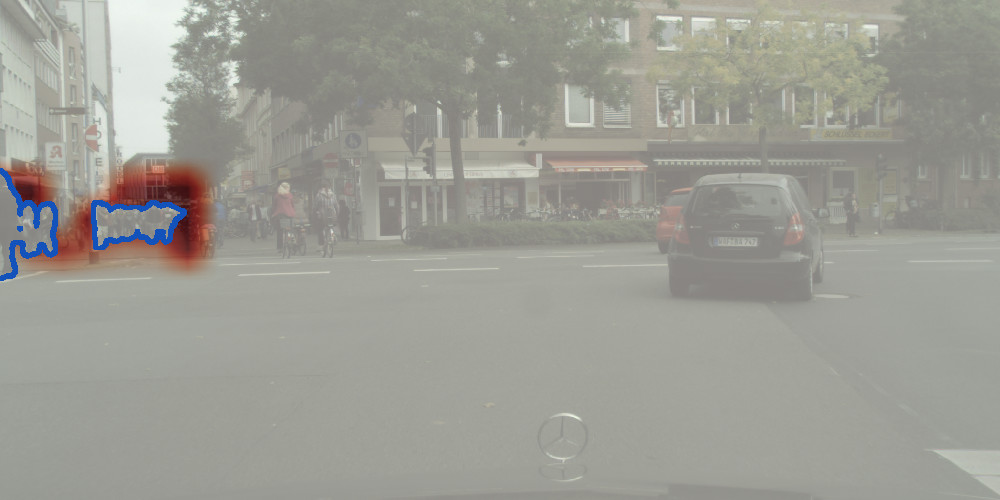}& {\footnotesize{}}
			\includegraphics[width=0.31\textwidth]{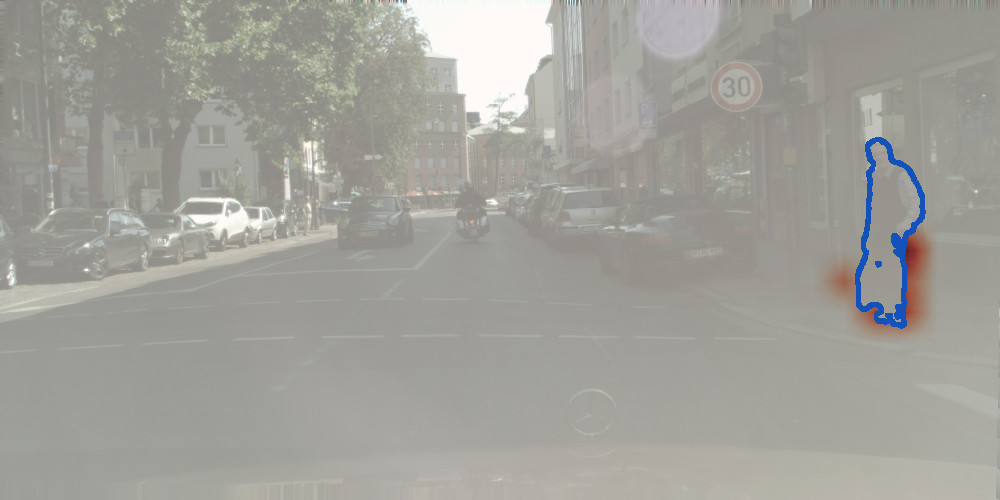} \tabularnewline
			j) Pedestrian Group & k) Pedestrian Group & l) Pedestrian \tabularnewline

			\includegraphics[width=0.31\textwidth]{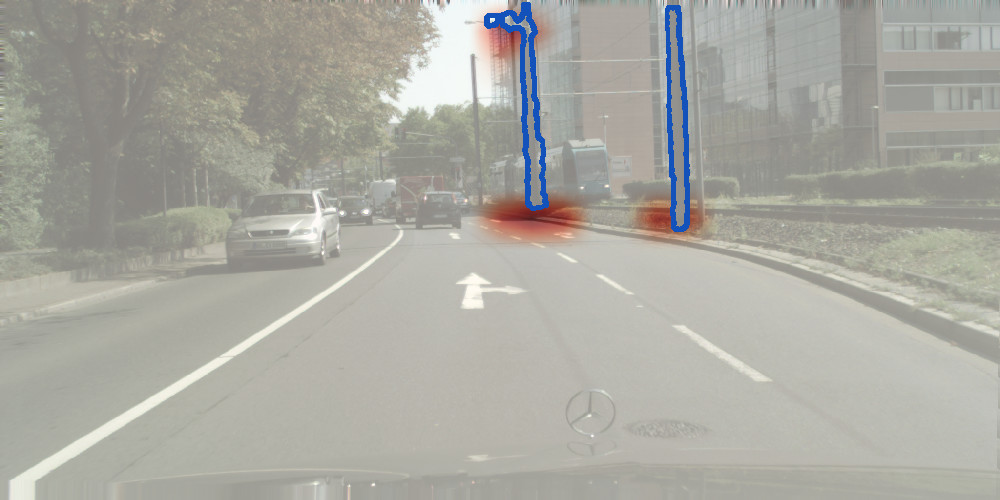}& {\footnotesize{}}
			\includegraphics[width=0.31\textwidth]{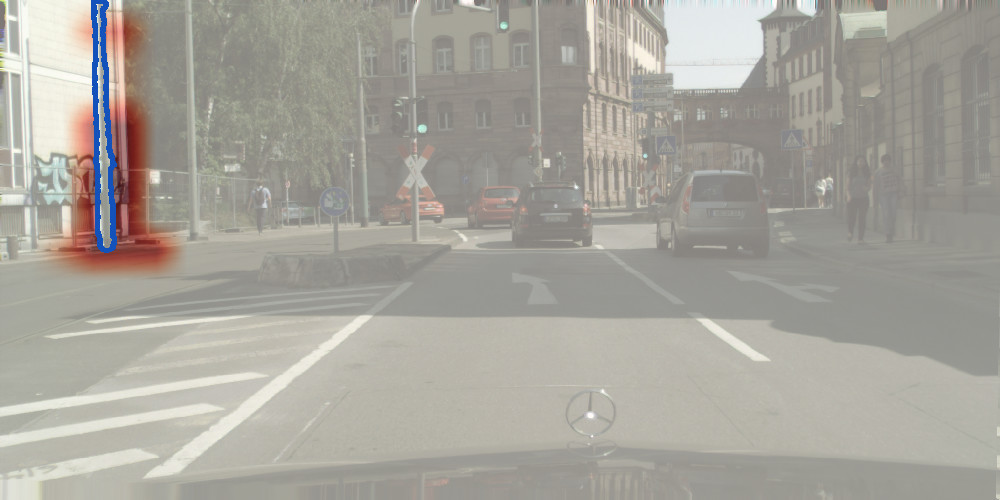}& {\footnotesize{}}
			\includegraphics[width=0.31\textwidth]{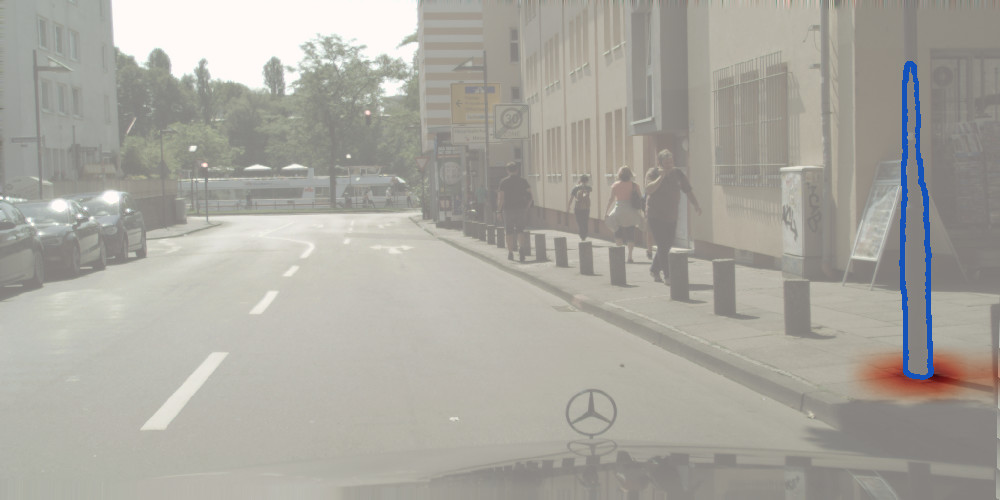}\tabularnewline
			m) Poles & n) Pole & o) Pole \tabularnewline

			\includegraphics[width=0.31\textwidth]{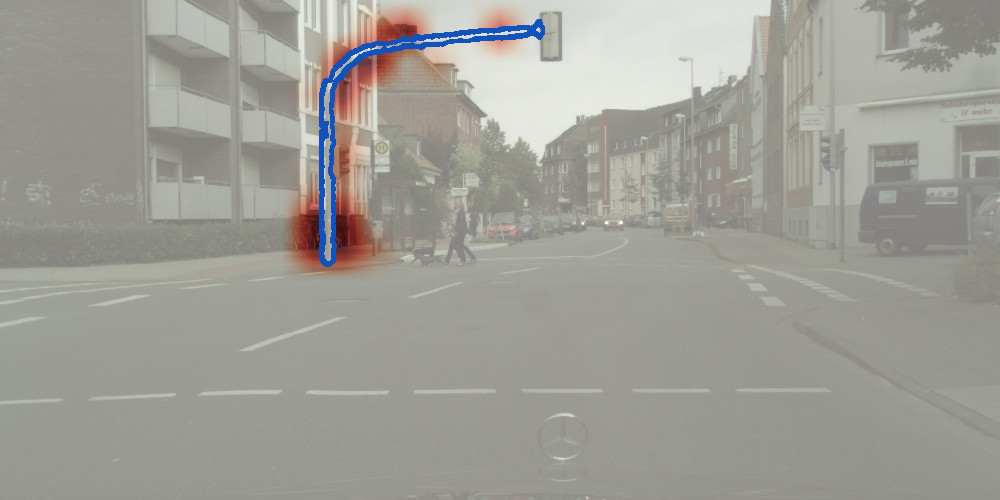} & {\footnotesize{}}
			\includegraphics[width=0.31\textwidth]{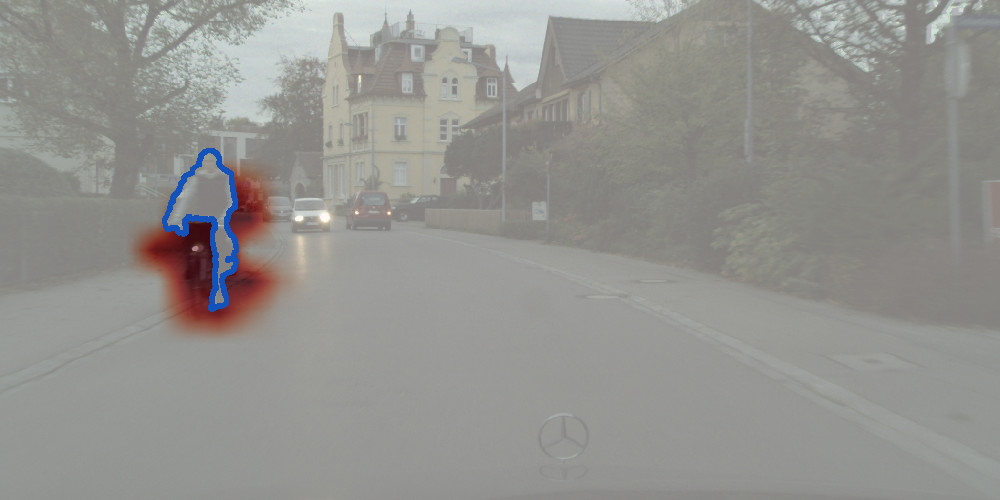}& {\footnotesize{}}
			\includegraphics[width=0.31\textwidth]{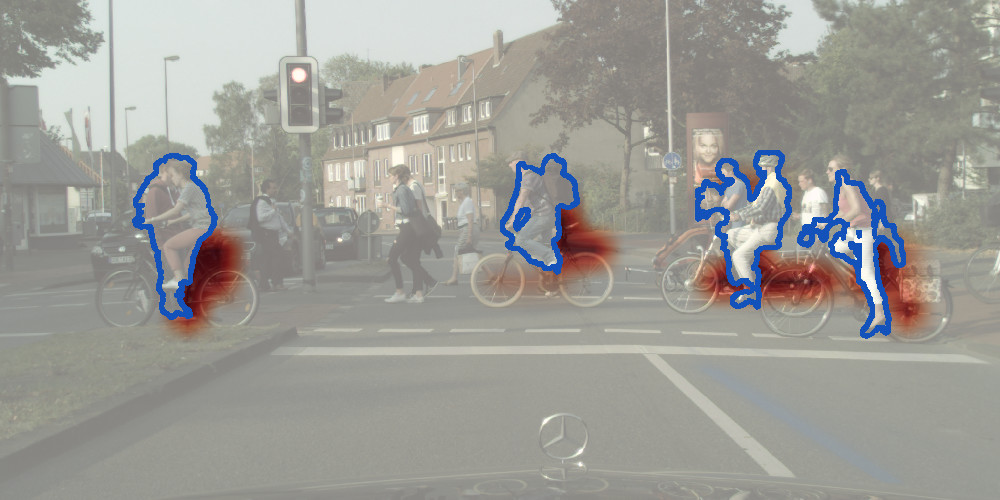}\tabularnewline
			p) Pole & q) Cycler & r) Cyclers\tabularnewline

			\includegraphics[width=0.31\textwidth]{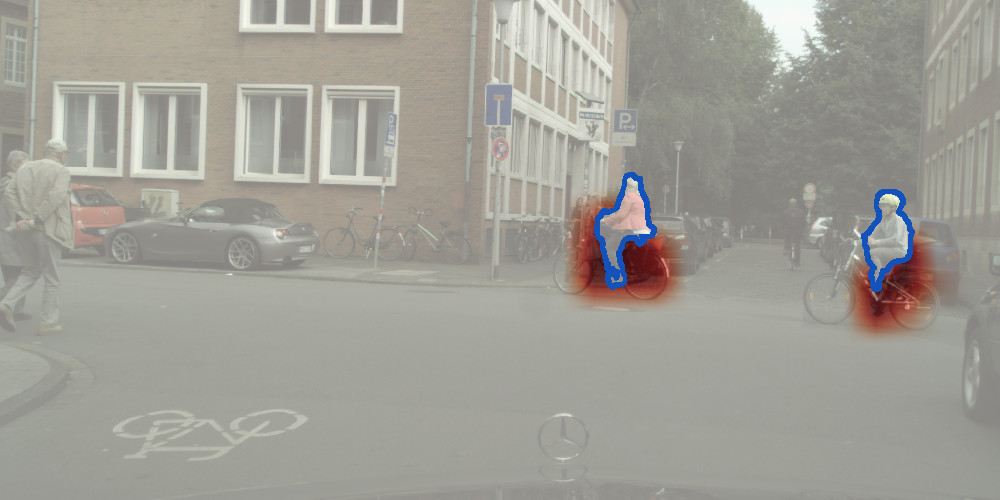}& {\footnotesize{}}
			\includegraphics[width=0.31\textwidth]{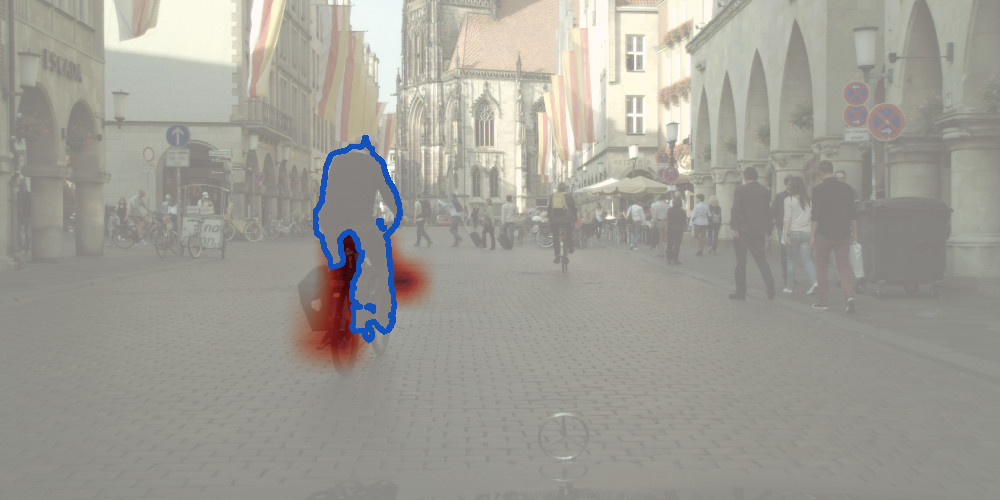}& {\footnotesize{}}
			\includegraphics[width=0.31\textwidth]{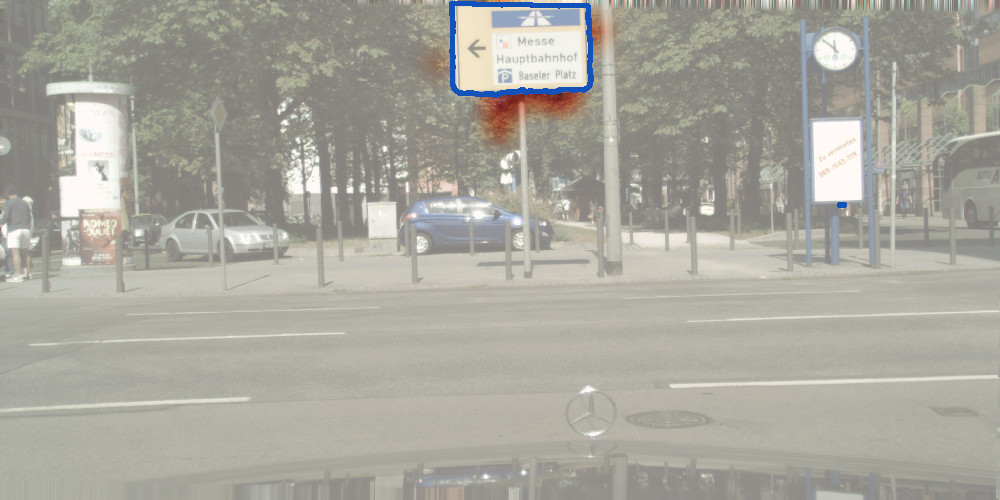}\tabularnewline
			s) Cyclers & t) Cycler & u) Traffic Sign\tabularnewline

			\includegraphics[width=0.31\textwidth]{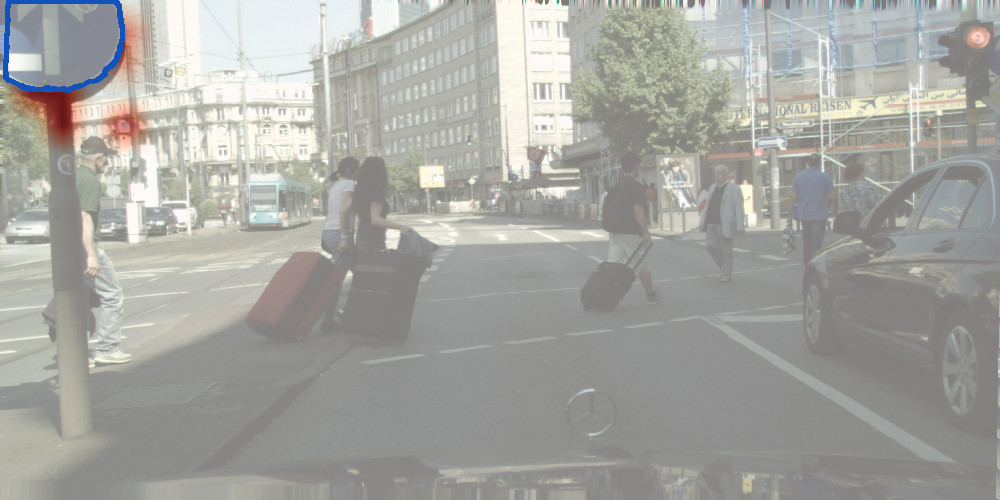}& {\footnotesize{}}
			\includegraphics[width=0.31\textwidth]{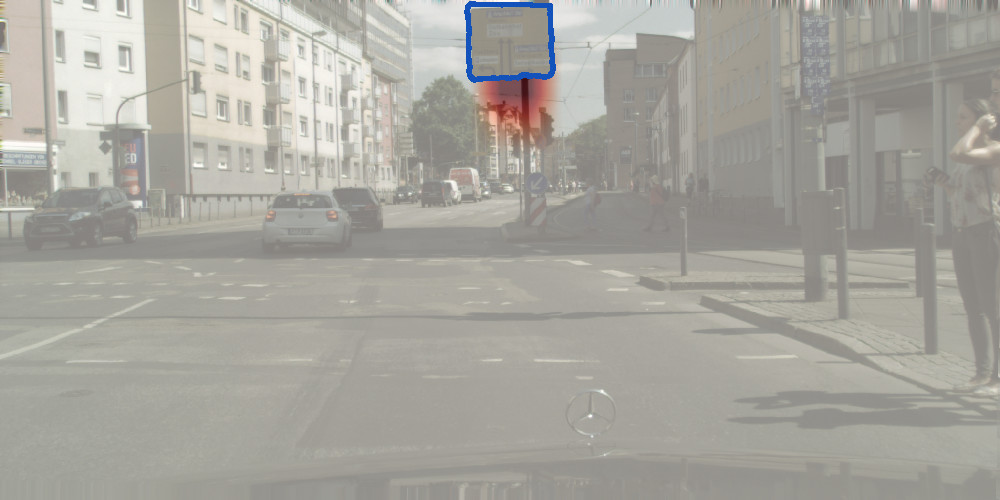}& {\footnotesize{}}
			\includegraphics[width=0.31\textwidth]{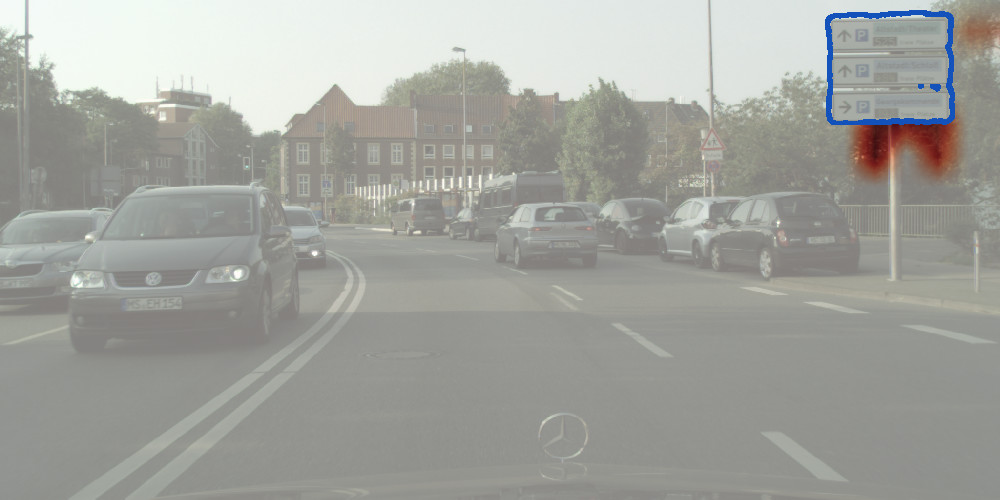} \tabularnewline
			v) Traffic Sign & w) Traffic Sign & x) Traffic Sign \tabularnewline
			
 		\end{tabular}\hfill{}
 		\par\end{centering}
 	
	\caption{\label{fig:cityscapes_saliency_2} Cityscapes examples supporting the class statistics.
	}
	\vspace{-1em}
\end{figure}

Fig. \ref{fig:cityscapes_saliency_2} shows additional representative qualitative examples of our method for a variety of request classes. Some interesting effects include saliency activations on road signs marked on the pavement for the car class (Fig. \ref{fig:cityscapes_saliency_2} (e)), as well as the behavior of the method for occluded objects. As can be seen in subfigures (b) and (d), the grid saliency for occluded objects tends to activate more strongly along the entire contour of the object. In particular, subfigure (d) contains two objects of similar size, pose and background, where only the second is partially occluded.

In certain cases, the prediction of the image with completely perturbed context is already so good so that no gradients are available for the preservation loss, and the optimization then focuses on minimizing the total loss by simply getting rid of the saliency activation altogether.
This typically results in large object instances of clearly identifiable classes such as cars or pedestrians with no saliency activations. Fig. \ref{fig:cityscapes_saliency_2} i) illustrates this for the left and center group of pedestrians. We consider this effect in itself to be a valid solution. An empty context saliency means that the requested object is self-containing and context is not necessary for its segmentation.

\clearpage
\section{Additional Results on MS COCO}

In order to show that our grid saliency can be applied on different types of real data and to various network architectures, we have repeated the experiments from Sec. 4.2 of the main paper on the MS COCO images~\cite{Lin2014MicrosoftCC} with different backbones for the state-of-the-art  Deeplabv3+~\cite{Chen2018EncoderDecoderWA} network. In particular, we  have chosen MobileNetv2~\cite{Sandler2018MobileNetV2IR} and Xception (XC)~\cite{Chollet2016XceptionDL} as backbones due to their different structure. All implementation details and optimization parameter settings are borrowed from the Cityscapes experiments.

\begin{figure}[H]
			\setlength{\tabcolsep}{0.35em}
			\renewcommand{\arraystretch}{1.4}
			\centering
			\begin{tabular}{ccccc}
				 & MobileNetv2 (MN) &  Xception (XC) & MobileNetv2 (MN) & Xception (XC) \tabularnewline
		\begin{turn}{90}
			\hspace{3.5em}{Boat}
		\end{turn}
				&
				\includegraphics[width=0.22\linewidth,height=0.15\textheight]{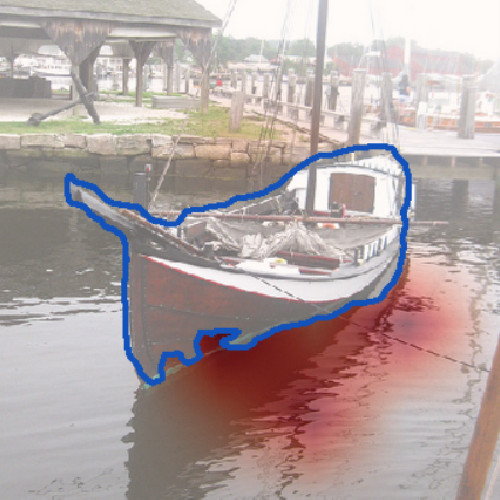}
				&
				\includegraphics[width=0.22\linewidth,height=0.15\textheight]{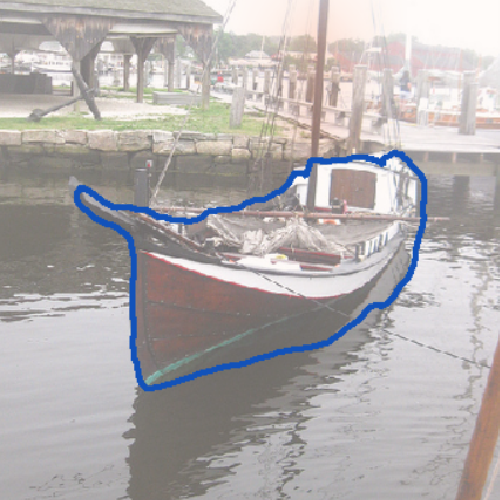}
				&
				\includegraphics[width=0.22\linewidth,height=0.15\textheight]{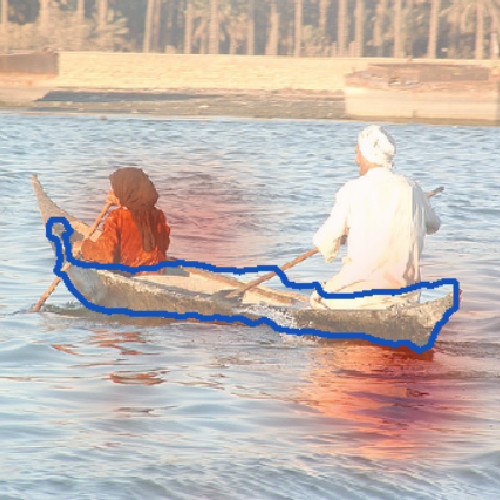}
				&
				\includegraphics[width=0.22\linewidth,height=0.15\textheight]{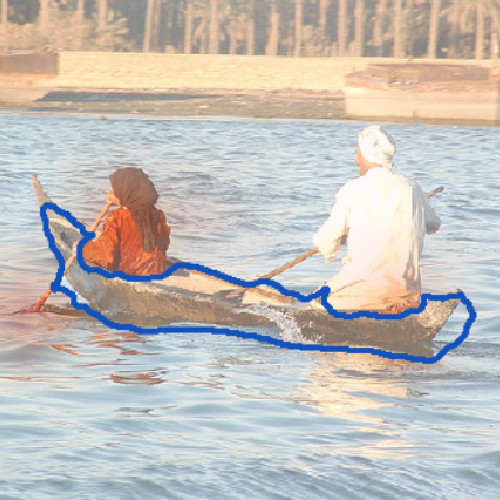}
				\tabularnewline
		
				\begin{turn}{90}
				\hspace{3.5em}{Train}
			\end{turn}
				&
				\includegraphics[width=0.22\linewidth,height=0.15\textheight]{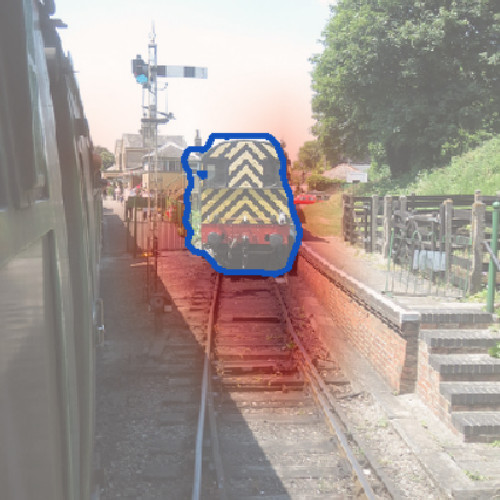}
				&
				\includegraphics[width=0.22\linewidth,height=0.15\textheight]{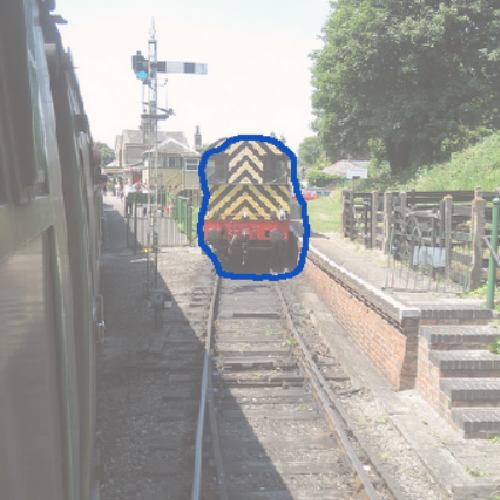}
				&
				\includegraphics[width=0.22\linewidth,height=0.15\textheight]{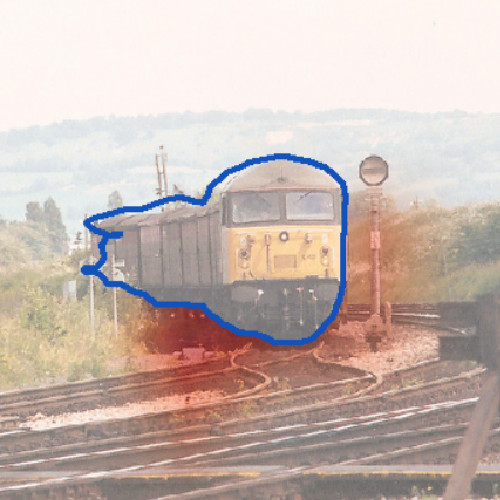}
				&
				\includegraphics[width=0.22\linewidth,height=0.15\textheight]{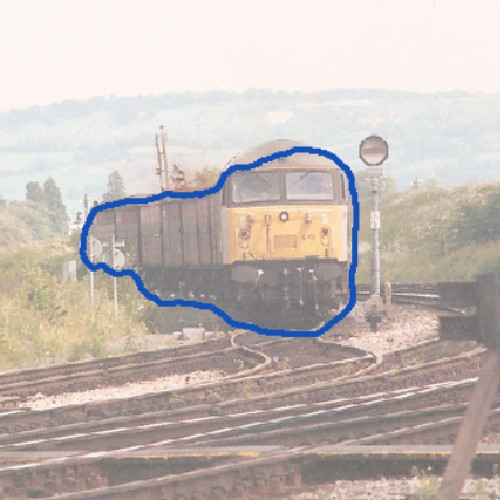}
				\tabularnewline
					\begin{turn}{90}
					\hspace{3.5em}{TV}
				\end{turn}
				&
				\includegraphics[width=0.22\linewidth,height=0.15\textheight]{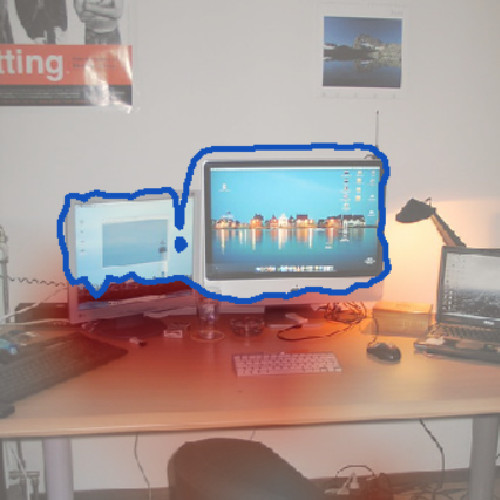}
				&
				\includegraphics[width=0.22\linewidth,height=0.15\textheight]{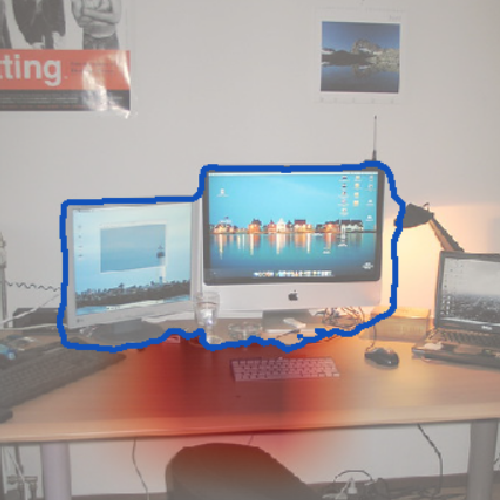}
				&
				\includegraphics[width=0.22\linewidth,height=0.15\textheight]{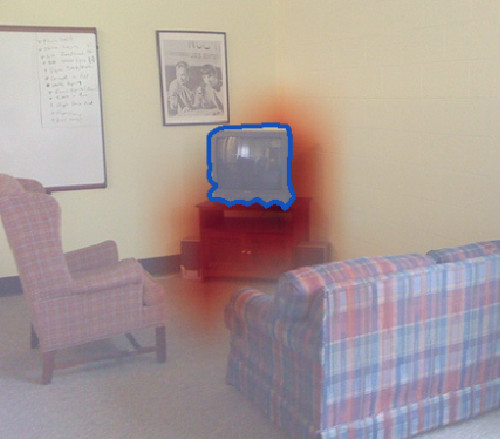}
				&
				\includegraphics[width=0.22\linewidth,height=0.15\textheight]{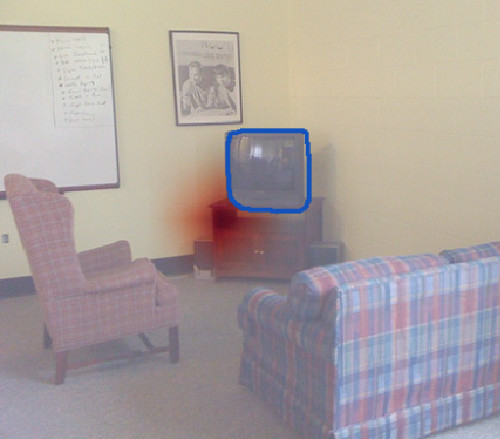}
			\end{tabular}
			\vspace{-0.5em}
			\caption{Context explanations produced by grid saliency on MS COCO~\cite{Lin2014MicrosoftCC} for the Deeplabv3+~\cite{Chen2018EncoderDecoderWA} network with MobileNetv2~\cite{Sandler2018MobileNetV2IR} and Xception (XC)~\cite{Chollet2016XceptionDL} as backbones.}
			\label{fig:coco_saliency}
		\end{figure}

In Fig. \ref{fig:coco_saliency}, we show some examples of context explanations on MS COCO obtained with our grid saliency method.  We observe that grid saliency provides sensible and coherent explanations for network decision making, which reflect semantic dependencies present in the data (e.g. boat appears on the water).
Context explanations produced by grid saliency can be also utilized to compare architectures with respect to their capacity to either learn or to be invariant towards context.
E.g., in Fig.~\ref{fig:coco_saliency} the segmentation network with MobileNetv2 (MN) backbone
learnt to rely more on context in contrast to its variant with a more powerful Xception (XC) backbone, 
which, for example, does not look at rails as context to segment the train.

\end{document}